\documentclass{article}

\usepackage{iclr2027_conference,times}
\usepackage{amsmath,amssymb}
\usepackage{booktabs}
\usepackage{graphicx}
\usepackage{microtype}
\usepackage{multirow}
\usepackage{tabularx}
\usepackage{makecell}
\usepackage{algorithm}
\usepackage{algpseudocode}
\usepackage{placeins}
\usepackage{float}
\usepackage{xcolor}
\usepackage{xurl}
\usepackage{hyperref}
\hypersetup{hidelinks}

\iclrfinalcopy
\usepackage{cleveref}
\makeatletter
\providecommand{\theHALG@line}{}
\renewcommand{\theHALG@line}{\thealgorithm.\arabic{ALG@line}}
\makeatother
\usepackage{tikz}
\usetikzlibrary{arrows.meta,positioning,fit,calc,backgrounds,shapes.geometric}

\definecolor{stateblue}{HTML}{D9EAF7}
\definecolor{raggreen}{HTML}{DDEEDC}
\definecolor{guardgold}{HTML}{F8E9B8}
\definecolor{failred}{HTML}{F7D9D5}
\definecolor{mutedgray}{HTML}{EFEFEF}

\newcommand{\method}{\textsc{PSS}}
\newcommand{\guarded}{\textsc{Guarded PSS}}
\newcommand{\pss}{PSS-320}
\newcommand{\allrecall}{All-Recall@4}

\newcommand{\best}[1]{\textbf{#1}}

\newcommand{\ourscell}[1]{#1}
\newcommand{\ourstag}[1]{#1}
\newcommand{\finalcell}[1]{#1}
\newcommand{\finaltag}[1]{#1}
\newcommand{\diagcell}[1]{#1}
\newcommand{\diagtag}[1]{#1}

\title{Parametric Session States Recover Retrieval Intent without Replaying History}

\author{
  Qingsen Ma \\
  The Chinese University of Hong Kong \\
  \texttt{qings.ma@gmail.com} \\
  \And
  Dianyun Wang \\
  Beijing University of Posts and Telecommunications \\
  \texttt{dianyunwang@bupt.edu.cn} \\
  \AND
  Jiahong Liu \\
  The Chinese University of Hong Kong \\
  \texttt{jiahong.liu21@gmail.com} \\
  \And
  Irwin King \\
  The Chinese University of Hong Kong \\
  \texttt{king@cse.cuhk.edu.hk}
}

\begin{document}

\maketitle

\begin{abstract}
Long-running assistants often interpret follow-up requests whose referents,
constraints, and active goals were established many turns earlier. The central
challenge is therefore not only context length, but retrieval intent: a current
query such as ``use the previous configuration'' may omit the project, entity,
and version terms needed to locate relevant evidence. Long-context models can
recover those terms by replaying the interaction, but repeatedly pay the cost
of processing history. Conventional retrieval-augmented generation (RAG) is
cheaper, yet assumes an information-sufficient query; conversational rewriting
either rereads the history or depends on an explicitly maintained textual
memory. We introduce \emph{Parametric Session State} (\method{}), a reusable
session-memory layer that remembers what to search for while keeping factual
knowledge external. An offline writer compiles the ordered interaction together
with  retrieval supervision into a fixed-shape, session-specific
LoRA state. At
query time, the state briefly produces a compact,  retrieval clue;
the system then retrieves inspectable source text, unloads the state, and lets
the frozen base model answer from that evidence. Across controlled and
sessionized public evaluations, correct PSS states improve target-evidence
retrieval for underspecified follow-ups. On the registered controlled
and public tracks, matched-wrong and shuffled controls attribute the gains to
the corresponding session, and the retrieval benefit transfers across model
families and heterogeneous backends. A three-seed carrier confirmation isolates
ordered-history information beyond supplied retrieval targets and measures a
6.25-point gain over a 256-token text memory on BGE-M3. These results establish \method{} as an
effective retrieval-supervised memory carrier for RAG: it restores missing
session scope, supports strong evidence access across diverse systems, and
preserves an external, auditable knowledge store.
\end{abstract}

\section{Introduction}

\begin{figure}[H]
\centering
\resizebox{\linewidth}{!}{
\includegraphics[width=\linewidth]{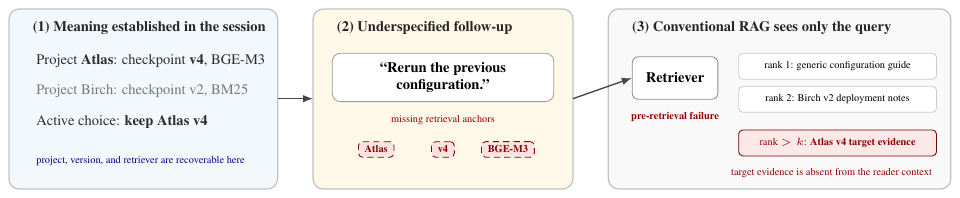}
}
\caption{Failure mode of conventional RAG under an underspecified follow-up.
Session identifiers are absent from the raw query, placing the intended
evidence outside the returned top-$k$.}
\label{fig:rag_failure}
\vspace{-1.6em}
\end{figure}

Long-running assistants accumulate more context than a user will restate in
every request. Consider a research agent that has discussed several projects,
checkpoints, and superseded configurations. The follow-up ``rerun the previous
configuration'' is meaningful inside the session, but nearly useless as a
standalone search query: the terms that identify the active project and
version are absent. In this setting, RAG can fail \emph{before retrieval
begins}: as \Cref{fig:rag_failure} illustrates, the identifiers that resolve
``previous'' remain in history while the retriever sees only the raw query. A
stronger encoder alone does not recover information absent from the query.
We study the following problem: \emph{can a long session leave a
reusable state that restores its retrieval intent without replaying historical
tokens at every query?}


Existing systems expose a difficult trade-off. Long-context models and KV
compression replay or retain token-dependent activations, preserving access to
the history but paying a cost that grows with it
\citep{xiao2024efficient,nawrot2024dynamic,li2024snapkv}. Textual memories,
summaries, and conversational query rewriting reduce that cost by making
the relevant state explicit, and they usually read history again when producing the
rewrite \citep{wu2022conqrr}. Standard RAG stores precise facts externally and
is easy to audit, yet assumes an information-sufficient query
\citep{lewis2020retrieval}. Global-memory RAG and parametric document modules encode
long inputs more deeply \citep{qian2025memorag,caccia2025training,eyuboglu2026cartridges},
but primarily target document understanding or direct answering rather than a
session whose present query is underspecified.

This leaves two linked gaps. The \emph{systems gap} is the absence of a reusable
layer between session history and retrieval: RAG needs a representation of
\emph{what this session is currently about} that can restore missing query
scope while leaving exact facts in the external store. The \emph{evaluation
gap} is the absence of a test that isolates this capability. Ordinary
long-context benchmarks usually present self-contained questions, so success
can come from the current query alone rather than recovered conversational
scope. A memory/no-memory comparison remains ambiguous because a generic
adapter may improve query style without encoding the corresponding session.
Identifying session-memory effects therefore requires underspecified queries,
session-disjoint splits, matched-wrong and shuffled states, null/OOD probes,
and session-level paired statistics.

To bridge the two gaps, we pair a retrieval architecture with an identification
protocol. For the systems gap, we propose \emph{Parametric Session State}
(\method{}): an offline writer compiles ordered history $H_s$ together with
 positive and negative retrieval targets into a fixed-rank LoRA
state $\Delta_s$. The state conditions a short online clue that restores
missing project, entity, relation, or version terms. The clue retrieves
evidence from the external corpus; the state is then unloaded and the frozen
base model answers from the retrieved text. An optional guard can select raw,
text-card, or state-assisted retrieval from observable utility signals. For the
evaluation gap, we make the current query deliberately underspecified, exclude
formal answers from the writer, freeze the corpus and retrieval stack, and
contrast corresponding states with wrong, shuffled, zero, null, and OOD
conditions at the session level. This paired design makes the claim falsifiable:
in the \emph{write--retrieve--unload} architecture, parameters should recover
session scope while precise facts remain external and auditable.

We evaluate whether this mechanism generalizes across public data, model
families, and retrieval backends. We derive  session protocols from
LongBench/InfiniteBench QA tasks
\citep{bai2024longbench,zhang2024infty} and UltraDomain, and freeze the
same state design across multiple language-model families
\citep{hui2024qwen2,jiang2023mistral7b,abdin2024phi3technicalreporthighly}. Correct, matched-wrong,
shuffled, and zero states test whether retrieval changes for the corresponding
session; heterogeneous retrievers test whether the clue transfers beyond its
writer; and same-evidence answer ablations test the value of keeping the state
loaded after retrieval. We then introduce \pss{}, a controlled collection of
technical sessions with ambiguous, compositional, temporal, null, and conflict
probes, to isolate causal effects and robustness properties that ordinary
self-contained QA rarely exposes.

Our contributions and findings are:
\begin{itemize}
    \item \textbf{A retrieval-scoped parametric session memory.} We introduce
    \method{}, a framework that uses a bounded, session-specific LoRA state as
    a carrier for retrieval intent. Its write--retrieve--unload lifecycle
    reconstructs what to search for while leaving precise claims in external,
    inspectable evidence.
    \item \textbf{An evaluation designed for missing retrieval intent.}
    We introduce  public sessionization and \pss{}, combining a
    shared candidate corpus, disjoint splits, causal state controls, null/OOD
    probes, and session-level inference to distinguish recovered session scope
    from generic query improvement. \Cref{app:why_pss320} gives the full
    motivation and design.
    \item \textbf{Generalization and mechanism evidence.} Correct states
    substantially improve retrieval over ambiguous queries and causal controls
    across three model families and heterogeneous lexical, dense, hierarchical,
    graph-compatible, and query-transformation backends. A three-seed
    confirmation shows that ordered history contributes retrieval signal beyond
    supplied targets, while state-fingerprint and functional-alignment analyses
    connect the learned carrier to session-specific outputs.
\end{itemize}

 \section{Method}
\label{sec:method}

\begin{figure}[H]
\centering
\resizebox{0.9\linewidth}{!}{
\includegraphics[width=\linewidth]{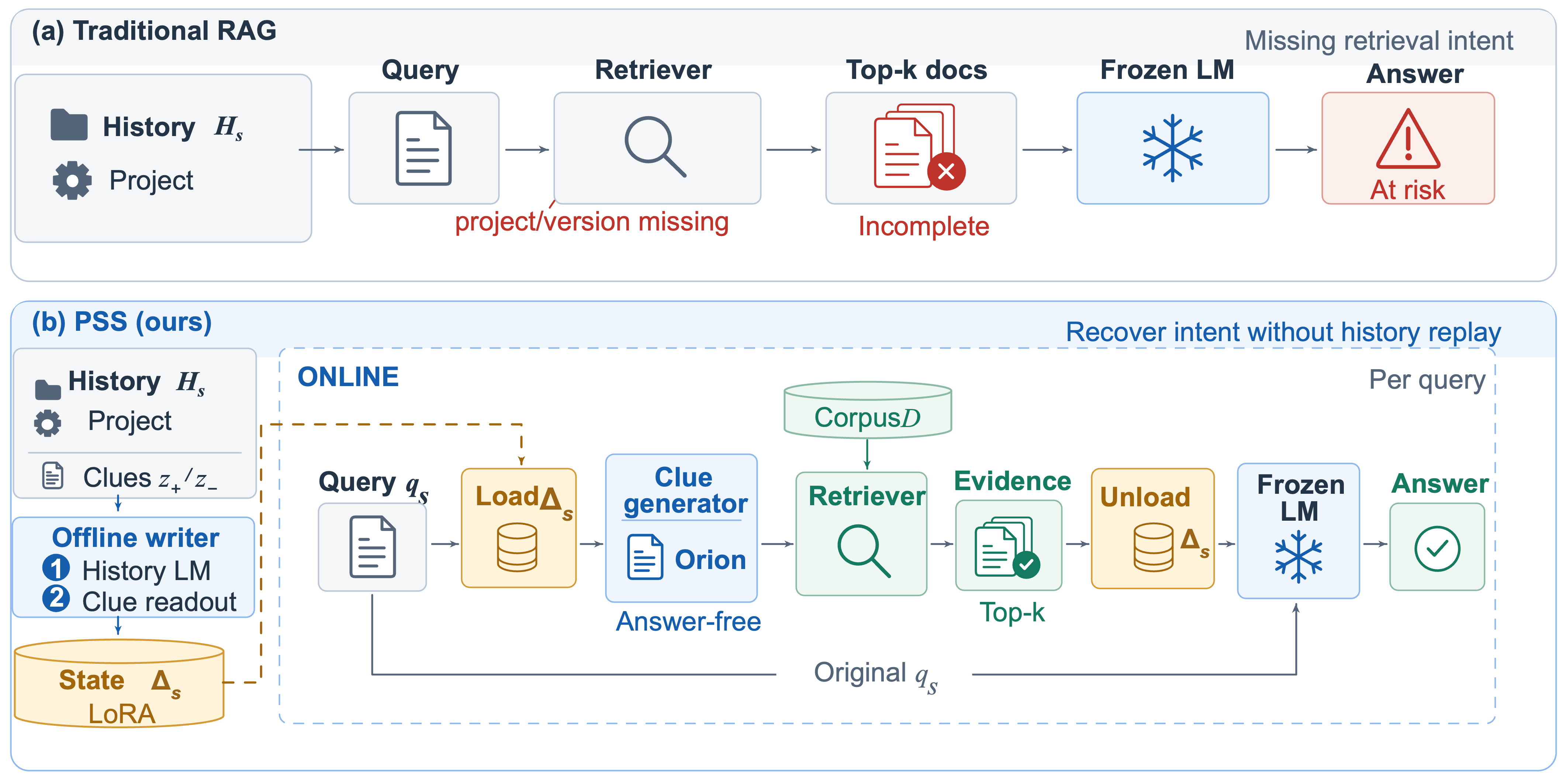}
}
\par\smallskip
\caption{\textbf{Parametric Session State workflow.} An offline writer compiles
the ordered history and positive/negative clue supervision into a
session-specific adapter through consecutive LM and readout phases. At inference, the adapter
generates a retrieval clue and is unloaded before the frozen base
model answers from retrieved evidence.}
    \label{fig:pss_overview}
\label{fig:system}
\end{figure}

\subsection{Problem formulation}
\label{sec:problem_formulation}

\paragraph{Inputs and frozen components.}
For session $s$, let $H_s=(h_1,\ldots,h_T)$ be the ordered interaction history,
$q_s$ the current underspecified query, and $\mathcal{D}$ a shared external
corpus of source chunks. The deployed RAG system contains a frozen retriever
$R_{\phi}$ and a frozen answer model $p_{\theta}$. During evaluation only, let
$E_s^{\star}\subset\mathcal{D}$ denote the target evidence; neither
$E_s^{\star}$ nor the formal answer is an input to the writer or online system.
A full-context reader can resolve references through
$p_{\theta}(y\mid H_s,q_s)$, whereas standard RAG retrieves
$R_{\phi}(q_s,\mathcal{D})$ and fails when the identifying terms occur only in
$H_s$.

\paragraph{Framework contract.}
PSS is a retrieval-scoped parametric memory framework: it changes the
query-interpretation layer while leaving the external index, retriever, and
answer model unchanged. Let $\mathcal{S}_s=(z_s^+,z_s^-)$ denote the offline, writer supervision used in our implementation: a positive search
clue and a same-type archived or conflicting clue. An optimization-based compiler
$\mathcal{A}_{\lambda}$ writes an architecture-compatible, bounded state,
which is invoked only to reconstruct a search clue:
\begin{equation}
\begin{aligned}
    \Delta_s &= \mathcal{A}_{\lambda}
        (H_s,\mathcal{S}_s;\theta,\mathcal{C}_{\theta}),
    &\tilde q_s &= G_{\theta+\Delta_s}(q_s),\\
    E_s &= \operatorname{TopK}_{k}R_{\phi}(\tilde q_s,\mathcal{D}),
    &y_s &\sim p_{\theta}(y\mid q_s,E_s).
\end{aligned}
\label{eq:problem_pipeline}
\end{equation}
The history and writer supervision are offline-only. At inference, $\Delta_s$
is loaded briefly, emits the  clue $\tilde q_s$, and is unloaded
before the frozen base model answers from retrieved evidence $E_s$. Formal
answers and target-document identifiers are excluded from $\mathcal{S}_s$.

\paragraph{Desired improvement.}
Let $U(E,E^{\star})$ be an evidence utility such as target-document recall.
The desired state improves evidence access for an ambiguous follow-up while
remaining bounded and eliminating online history replay:
\begin{equation}
\begin{aligned}
\Delta_s^{\star}\in\arg\max_{\Delta\in\mathcal{C}_{\theta}}\;&
U\!\left(\operatorname{TopK}_{k}R_{\phi}
\bigl(G_{\theta+\Delta}(q_s),\mathcal{D}\bigr),E_s^{\star}\right)\\
\text{s.t. }\;&\operatorname{bytes}(\Delta)\leq B,\quad
H_s\notin\text{online inputs},\quad \theta,\phi\text{ frozen}.
\end{aligned}
\label{eq:problem_objective}
\end{equation}
The gold set $E_s^{\star}$ and its identifiers appear only in this evaluation
objective, not in the compiler or deployed system; the compiler instead uses
the  clue supervision defined above. Because our implementation does
not differentiate through retrieval, \Cref{eq:retloss} provides a
retrieval-aligned surrogate. The intended gain is evidence access under missing
intent.\Cref{app:formal_algorithms} gives the complete offline and online contracts.

\subsection{LoRA as a bounded session-state carrier}

LoRA supplies the bounded state carrier in our implementation; the PSS
contribution is the session-specific state contract and retrieval lifecycle
built around it. For a frozen autoregressive model $\theta$, let
$\mathcal{C}_{\theta}$ specify a set of writable layer--projection pairs.  The
compiler creates one low-rank update for each pair,
\begin{equation}
 W^{(s)}_{\ell,m}=W_{\ell,m}
   + \frac{\alpha}{r}B^{(s)}_{\ell,m}A^{(s)}_{\ell,m},
 \qquad (\ell,m)\in\mathcal{C}_{\theta},
\end{equation}
while the base parameters remain frozen. The compiler optimizes only these
session-local factors and serializes them as $\Delta_s$. Projection names,
fused modules, and writable layers may vary by backbone; the functional
contract remains fixed: one session produces one bounded state, the state
produces a textual retrieval interface, and it is removed before answering.
Unlike conventional task LoRA, the adapter is not a shared task solution and
does not remain active for final generation.

Qwen2.5-7B is the primary concrete instance \citep{hui2024qwen2}; Mistral-7B
and Phi-3-mini instantiate the same contract with their native projection
layouts. Each model writes its own rank-32, single-slot BF16 state, with no
tensor transfer across backbones. Layer placement, optimization steps, and
state-size accounting are reported in \Cref{app:training,app:efficiency}.

\subsection{Retrieval-aligned session compilation}

The selected \texttt{retrieval\_aligned} compiler first optimizes history
language modeling over bounded chunks, then optimizes  clue
readout while retaining the same state. Each phase initializes its own
optimizer at the registered learning rate. Let
$z_s^+$ be a clue assembled from write-time session scope and
evidence-related fields represented in the history, and let $z_s^-$
be a same-type archived or conflicting clue.The readout term is
\begin{equation}
\mathcal{L}_{\mathrm{ret}} =
\ell(z_s^+) +
\max\left(0,\,m+\ell(z_s^+)-\ell(z_s^-)\right),
\label{eq:retloss}
\end{equation}
where $\ell$ is teacher-forced sequence negative log likelihood and $m=0.2$.
The primary readout phase uses a fixed session-clue instruction.
Its supervision is constructed automatically: \pss{} uses controlled
session fields also represented in the history, while the public protocol
uses history-parsed active requests as positive targets and same-task
requests from other sessions as negatives. In these protocols, target
construction requires no per-session manual clue annotation and does not
read future follow-up probes, formal answers, or target-document IDs.
A deployment can supply these write-time inputs through existing session
metadata or a history-to-target extractor. Our controlled evaluation
examines how history and the resulting supervision are compiled into
reusable parameters and accessed for retrieval, with the upstream
session associations explicitly available.
A separate question-conditioned extension with multiple historical records is evaluated in \Cref{app:question_conditioned_records}. Objective, content, rank, topology, and precision ablations are detailed in \Cref{app:training,app:design_interventions}.

\subsection{Intent readout, retrieval, and unloading}

The core path, \emph{PSS clue retrieval}, loads $\Delta_s$, generates
$\tilde q_s$, and exposes that text to a retriever. This
textual boundary makes the memory layer compatible with lexical, dense,
hierarchical, graph, and query-transformation backends. The \emph{Raw + PSS
RRF} ablation unconditionally fuses raw-query and PSS-clue rankings according
to
\begin{equation}
    \mathrm{RRF}(d)=
    \sum_{r\in\{\mathrm{raw},\mathrm{PSS}\}}
    \frac{1}{60+\mathrm{rank}_{r}(d)}
\end{equation}
\citep{cormack2009reciprocal}. \guarded{} additionally routes among raw, structured
card, and PSS-RRF using only observable retrieval features; its cross-fitting
and safety contract appear in \Cref{app:router}. The complete guarded path
includes the card's text input and both clue generators; its measured call
and token counts are reported in \Cref{app:followup_cost}.

After retrieval, the adapter is unloaded and the frozen base model answers
from $(q_s,E_s)$. A same-query, same-evidence ablation varies only the
answer-stage state, isolating decoder effects from retrieval effects
(\Cref{app:answer}). The state is fixed-size per session, although offline
write compute grows with history and total storage grows with the number of
sessions; complete accounting is deferred to \Cref{app:efficiency}.
\section{Experiments}
\label{sec:experiments}

We organize the main evaluation around three questions. \textbf{RQ1: Does PSS
recover retrieval intent across data, models, and retrievers?} \textbf{RQ2:
Do functional outputs align with session-specific intent?} \textbf{RQ3: Which
design choices produce the gain?} We report
the online serving cost alongside retrieval utility; full storage and
amortization profiles appear in \Cref{app:efficiency,app:followup_cost}.

\subsection{Evaluation design}
\label{sec:evaluation_motivation}

We hold the corpus, index, retriever, and reader fixed within each primary
protocol and vary the query source or session state. Public sessionizations
test transfer; \pss{} uses a shared cross-session corpus and state
interventions to identify session memory. \emph{PSS clue retrieval} searches
with the generated clue, \emph{PSS-RRF} fuses raw and clue rankings, and
\emph{Guarded PSS} routes among raw, card, and PSS-RRF. Fixed-question QA
then tests whether better evidence improves answering after state unloading.
Evaluation rationale and full method definitions appear in
\Cref{app:evaluation_design,app:why_pss320}.

\subsection{Experimental setup}
\label{sec:experimental_setup}

We evaluate one frozen PSS recipe in three complementary settings: a sealed
262-session transformation of seven LongBench/InfiniteBench QA tasks, a
112-session UltraDomain transfer set, and \pss{}, whose shared candidate
corpus supports a source-group-disjoint 256-session test. Public tracks report
target-document All-Recall@3; \pss{} reports all-support evidence recall at
$k=4$. Correct, matched-wrong, shuffled, zero, null, and OOD conditions isolate
session-specific memory from generic adapter or rewrite effects.

Qwen2.5-7B is the primary backbone; Mistral-7B and Phi-3-mini test portability,
and lexical, dense, hierarchical, graph, and query-transformation retrievers
test the textual clue interface. Primary state-writing evaluations use writer
seeds 91--93; follow-up protocols specify their seed counts separately. Sessions are
the unit of inference. We use
paired bootstrap intervals, registered McNemar/sign-flip tests, and Holm
correction \citep{tibshirani1993introduction,holm1979simple}; daggers are emitted
only when every registered comparison passes. Dataset construction, frozen
registrations, model contracts, baselines, prompts, and statistical gates are
given in \Cref{app:setup,app:data,app:training,app:arms,app:protocol,app:statistics}.

\subsection{RQ1: Retrieval-intent recovery generalizes}
\label{sec:public_results}

\begin{figure}[t]
\centering
\includegraphics[width=\linewidth]{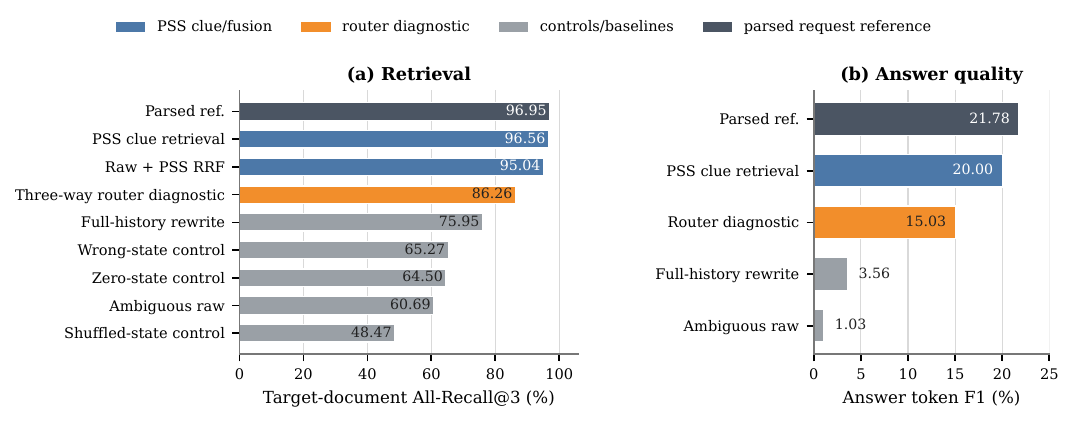}
\caption{Session-weighted target-document retrieval and isolated-reader
answer F1 on 262 sealed public sessions (percent). F1 evaluates the complete
query-and-evidence interface; fixed-question evidence attribution is
reported separately in \Cref{tab:main_controlled_qa}.}
\label{fig:public_pooled}
\vspace{-1.6em}
\end{figure}

\paragraph{LongBench and InfiniteBench.}

On the 262 sealed public sessions, Qwen PSS clue retrieval with the correct
session state improves
All-Recall@3 over raw, wrong-state, and shuffled-state queries by 35.88, 31.30,
and 48.09 points, respectively (all $p_{\mathrm{Holm}}=5.99994\times10^{-5}$).
On the five multi-document tasks, PSS reaches 77.50--100\% versus
30.00--57.50\% for raw retrieval; PSS-RRF reaches 82.50\% on Qasper.
En.QA and NarrativeQA saturate because each constructed session indexes one
shared document (\Cref{tab:public_tasks}). 
The parsed active-request reference directly retrieves with the complete
request designated by the protocol's activity marker, measuring text
retrieval with request selection already resolved. PSS uses this same
request as positive readout supervision and compiles it together with
history into a reusable parameter state. The multi-record extension tests
question-conditioned access to one compiled state
(\Cref{app:question_conditioned_records}). Under the original
query-and-evidence interface, answer F1 changes from 1.03\% to 20.00\%.
Router-transfer and null/OOD results appear in \Cref{app:limitations}.

\paragraph{Retrieved evidence improves answering.}
A registered follow-up fixes each reader's source question on the same
262-session split and varies the retrieved top-3 chunks, with PSS unloaded.
Controlled F1 improves over raw-query
evidence by 10.60, 4.06, and 4.83 points for Qwen, Mistral, and Phi-3,
respectively (all $p_H=0.0003$; \Cref{tab:main_controlled_qa}).
For Qwen, PSS also improves over the cross-domain ConvSearch-R1 checkpoint by
3.48 points (95\% CI $[1.51,5.57]$, $p_H=0.006$).
Retrieval interfaces remain unchanged. Complete contrasts, the UltraDomain
replication, and an independent 16-session evidence control appear in
\Cref{app:cqr_qa,app:followup_qa}.

\begin{table}[t]
\centering
\small
\setlength{\tabcolsep}{5pt}
\caption{Controlled answer F1 (percent) in the 262-session public follow-up.
Each reader receives a method-invariant source question and method-retrieved
top-3 evidence, with history hidden and PSS unloaded. All four query sources
are retained; complete retrieval and paired results appear in \Cref{app:cqr_qa}.}
\label{tab:main_controlled_qa}
\begin{tabular}{lrrr}
\toprule
Query source & Qwen2.5-7B & Mistral-7B & Phi-3-mini \\
\midrule
Ambiguous raw query & 4.62 & 6.54 & 8.61 \\
Parsed ref. & 14.96 & 11.32 & 16.03 \\
ConvSearch-R1 & 11.74 & 10.02 & 11.81 \\
\textbf{PSS clue retrieval} & 15.22 & 10.60 & 13.44 \\
\bottomrule
\end{tabular}
\end{table}

\paragraph{UltraDomain.}
On the  UltraDomain transformation, correct states reach
98.21--99.11\% All-Recall@3 across all three backbones, improving the already
strong raw-query recall by 1.79--4.46 points. The gains are positive for every
backbone. This result extends the method to financial, legal, and mixed-domain documents.

\paragraph{Controlled \pss{} mechanism test.}

The raw underspecified query retrieves all required evidence with probability
0.1734. A compact 32-token structured card and 128-token summary reach
0.3945 and 0.4023. PSS clue retrieval reaches 0.5008 with no online historical
text, improving over the compact card by 10.63 points. The complete
\guarded{} system reaches 0.5063, an 11.18-point gain over that card, while
retaining its card-based routing branch. These results establish the value
of parameterized session scope against compact text-state baselines.
\Cref{app:readout_cost_followup} separates these cards from parsed request
references and supplied-clue caches, and reports complete readout controls.

The paired effects pass H1/H2. \guarded{} corrects 134 sessions missed
by raw retrieval, with no raw-only success; against the card, there are 56
Guarded-PSS-only and seven card-only successes. Within \guarded{}, the reported correct-state performance exceeds
the matched-wrong and parameter-shuffled controls by 0.1750 and
0.1211 in \allrecall{}, respectively. These contrasts support
session-specific state contributions within the guarded retrieval system.

\begin{figure}[t]
\centering
\includegraphics[width=\linewidth]{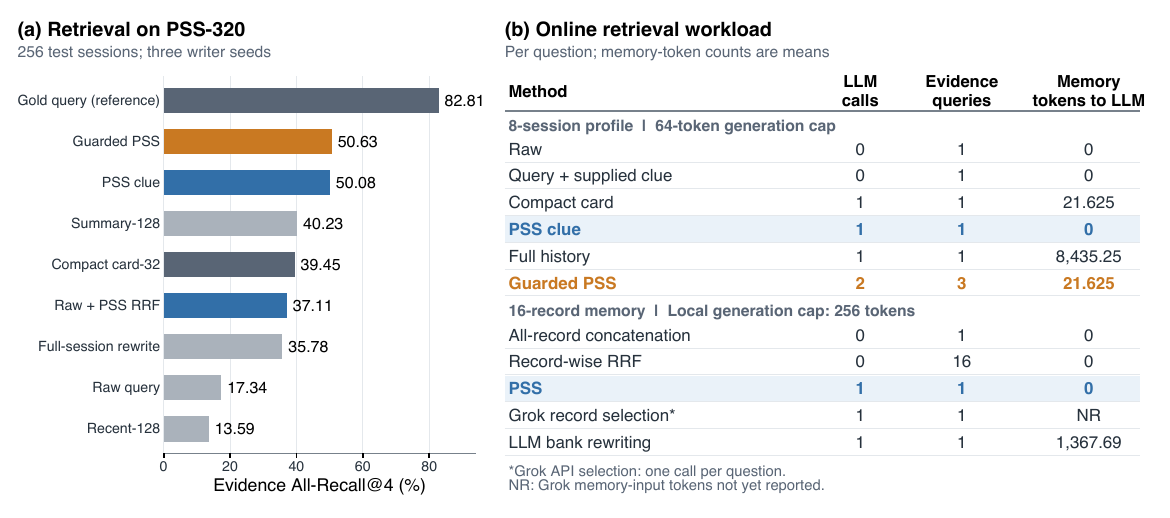}
\caption{\textbf{Retrieval utility and online workload.}
(a) Original 256-session \pss{} test aggregation.
(b) Separate eight-session profile and $K=16$ memory experiment.
Columns count generations, evidence queries, and mean memory tokens supplied
to LLMs; zero memory input does not imply an absent text cache.
Each block retains its generation budget. Workload components exclude final
answering and offline writing; measured timings and full accounting appear
in \Cref{app:question_conditioned_records,app:followup_cost}.}
\label{fig:main_retrieval}
\vspace{-1.6em}
\end{figure}

\paragraph{Online retrieval workload.}
\Cref{fig:main_retrieval}(b) separates generation calls, evidence queries,
and memory-text input. PSS uses one generation and one evidence query,
with no memory-text input to the LLM. At $K=16$, it replaces record-wise
fusion's 16 queries with one and reaches 83.75/95.00\% All-Recall@4
(BGE-M3/BM25). Grok-based record selection accesses the persistent text bank
through an external model API before evidence retrieval.
PSS performs question-conditioned readout from its compiled
parameter state without supplying the text bank to the readout model
(\Cref{app:question_conditioned_records}).

\paragraph{Cross-model and cross-retriever summary.}
\label{sec:cross_model_results}

Across all three backbones, PSS improves public-session retrieval by
33.97--35.88 points and passes the matched-wrong and shuffled controls. On
\pss{}, state clues improve BGE-M3 by 21.48--45.63 points and the
graph-compatible retriever by 32.11--53.13 points; every state row also exceeds
its card baseline and receives a script-computed dagger. Independently written
states therefore preserve the same retrieval-intent function across three
model architectures.

\vspace{-1.6em}
\begin{table}[H]
\caption{Transfer across backbones and retrieval backends (percent). Panel A
reports the formal cross-model evaluation: Public and Ultra use target-document
All-Recall@3; BGE-M3 and GraphRAG use \pss{} evidence \allrecall{}.
Panel B is a disjoint 112-session public strong-backend confirmation.
Sup. denotes positive Holm-corrected effects over raw, wrong, and shuffled
states for the same frozen backend. Panel A's \pss{} card comparisons use
compact structured cards; Panel B uses the parsed active-request reference
(``Parsed ref.'').
$\dagger$ in Panel A marks \pss{} states that beat raw and card.
Asterisks identify compatible HyDE/RQ-RAG pipelines.
Arrows indicate the preferred direction: higher recall and gains,
and lower state storage.}
\label{tab:cross_retrievers}
\label{tab:cross_model_portability}
\centering
\scriptsize

\textit{Panel A: cross-model evaluation.}\\[0.15em]
\resizebox{\textwidth}{!}{%
\begin{tabular}{lrrcrrrrrrcr}
\toprule
& \multicolumn{3}{c}{Public}
& \multicolumn{2}{c}{BGE-M3}
& \multicolumn{2}{c}{GraphRAG}
& \multicolumn{3}{c}{Ultra}
& \\
\cmidrule(lr){2-4}
\cmidrule(lr){5-6}
\cmidrule(lr){7-8}
\cmidrule(lr){9-11}
Model
& Raw $\uparrow$ & PSS $\uparrow$ & Sup.
& Raw $\uparrow$ & PSS $\uparrow$
& Raw $\uparrow$ & PSS $\uparrow$
& Raw $\uparrow$ & PSS $\uparrow$ & Sup.
& State MiB $\downarrow$ \\
\midrule
Qwen2.5-7B
& 60.69 & 96.56 & yes
& 12.42 & 33.91$^\dagger$
& 16.33 & 48.44$^\dagger$
& 96.43 & 98.21 & no
& 44 \\
Mistral-7B
& 61.83 & 95.80 & yes
& 12.03 & 51.09$^\dagger$
& 16.33 & 67.66$^\dagger$
& 94.64 & 98.21 & no
& 40 \\
Phi-3-mini
& 60.69 & 95.80 & yes
& 11.95 & 57.58$^\dagger$
& 16.09 & 69.22$^\dagger$
& 94.64 & 99.11 & no
& 24 \\
\bottomrule
\end{tabular}}

\textit{Panel B: frozen-backend public confirmation
($n=112$ sessions; disjoint from the smoke set).}\\[0.15em]
\resizebox{\textwidth}{!}{%
\begin{tabular}{lrrrrrrrrrc}
\toprule
Retriever
& Raw $\uparrow$
& Full hist. $\uparrow$
& Parsed ref. $\uparrow$
& PSS clue $\uparrow$
& $\Delta$ vs. raw $\uparrow$
& Wrong $\uparrow$
& Shuffled $\uparrow$
& PSS-RRF $\uparrow$
& Gold $\uparrow$
& Sup. \\
\midrule
BM25
& 66.07 & 69.64 & 83.93 & 84.82
& +18.75 & 71.43 & 59.82 & 84.82 & 83.93 & no \\
Contriever
& 69.64 & 78.57 & 98.21 & 100.00
& +30.36 & 67.86 & 52.68 & 99.11 & 98.21 & yes \\
BGE-M3
& 63.39 & 77.68 & 98.21 & 99.11
& +35.71 & 64.29 & 51.79 & 97.32 & 98.21 & yes \\
Qwen3-Embedding-4B
& 69.64 & 80.36 & 99.11 & 98.21
& +28.57 & 68.75 & 49.11 & 100.00 & 99.11 & yes \\
HyDE-compatible$^{*}$
& 62.50 & 75.89 & 97.32 & 98.21
& +35.71 & 64.29 & 51.79 & 98.21 & 97.32 & yes \\
RQ-RAG-compatible$^{*}$
& 62.50 & 74.11 & 99.11 & 98.21
& +35.71 & 68.75 & 47.32 & 100.00 & 99.11 & yes \\
RAPTOR
& 5.36 & 11.61 & 16.96 & 16.07
& +10.71 & 7.14 & 7.14 & -- & 16.96 & no \\
\bottomrule
\end{tabular}}
\end{table}
\vspace{-1.6em}

\begin{figure}[t]
\centering
\includegraphics[width=\linewidth]{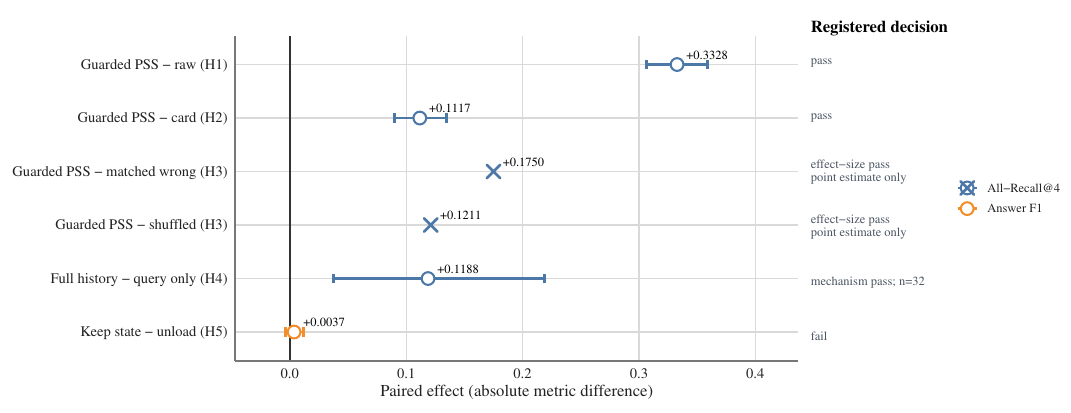}
\caption{Session-level paired effects for the registered controlled
interventions. The vertical reference marks zero effect.}
\label{fig:main_effects}
\vspace{-1.6em}
\end{figure}

Panel B provides a preregistered 112-session confirmation independent of the
seven-session implementation smoke. Across the seven evaluated backends, PSS
improves over the ambiguous raw query by 10.71--35.71 points. BGE-M3,
Contriever, Qwen3-Embedding-4B, and compatible HyDE/RQ-RAG pipelines satisfy the
joint Holm-corrected rule against raw, matched-wrong, and shuffled states; their
mean All-Recall@3 rises from 65.54\% to 98.75\%. BM25 and RAPTOR add gains of
18.75 and 10.71 points, respectively. PSS also matches the parsed request
reference within $-0.89$ to $+1.79$ points without online history replay,
establishing it as a portable intent layer for heterogeneous frozen retrieval
systems. Exact backend
contracts and statistical gates are documented in
\Cref{app:retrievers,app:limitations}
\citep{izacard2021unsupervised,gao2023precise,chan2024rq,
zhang2025qwen3,sarthi2024raptor}.


\subsection{RQ2: Do functional outputs align with session-specific intent?}

Correct-state clues move ambiguous queries toward their gold retrieval-intent
rewrites in the original 1,024-dimensional BGE-M3 space. Relative to raw,
matched-wrong, and shuffled controls, session-level target alignment improves by
0.1136--0.1779; every confidence interval is bounded away from zero and every
Holm-adjusted $p$-value is 0.000300 (\Cref{tab:rq2_relational}).

\begin{table}[H]
    \vspace{-1.6em}
\caption{Session-level functional-output alignment effects aggregated over
writer seeds 91--93 in the original 1,024-dimensional BGE-M3 space, with 95\%
stratified-bootstrap CIs and Holm-adjusted one-sided sign-flip tests.}
\label{tab:rq2_relational}
\centering
\small
\begin{tabular}{llrrr}
\toprule
Candidate & Reference & Effect & 95\% CI & $p_{\mathrm{Holm}}$ \\
\midrule
PSS clue retrieval & Ambiguous raw query & +0.1136 & [+0.1008, +0.1266] & 0.000300 \\
PSS clue retrieval & Matched-wrong state & +0.1227 & [+0.1119, +0.1336] & 0.000300 \\
PSS clue retrieval & Shuffled state & +0.1779 & [+0.1646, +0.1907] & 0.000300 \\
\bottomrule
\end{tabular}
\vspace{-1.6em}
\end{table}

Factorization-invariant effective-update fingerprints are not organized by the
eight coarse semantic categories: cosine silhouette is $-0.0090$, and
same-category nearest-neighbor frequency is 1.17\% versus a 12.16\% chance rate.
Thus the functional output is session-specific even though this coarse taxonomy
does not explain state proximity. Because target alignment is encouraged by the
retrieval-clue objective, we interpret it as validation of the intended state
function. Full
fingerprint construction, UMAPs, energy heatmaps, and analysis scope appear in
\Cref{app:rq2_details,app:limitations}.

\subsection{RQ3: Design interventions and ablations}

The registered interventions confirm that full ordered history, session
correspondence, and learned parameter structure each contribute to retrieval;
PSS clue retrieval supplies the largest mechanism gain, while guarded use
gives the best non-oracle score (\Cref{fig:main_effects}).

\begin{table}[H]
\caption{Selected cross-backbone RQ3 results. Scores and effects are All-Recall percentage points. The canonical cell is $r{=}32$, $\alpha{=}64$, last-8, single-slot. Confirmation uses a disjoint 16-session split and reports a 95\% CI.}
\label{tab:rq3_main_selection}
\centering
\scriptsize
\setlength{\tabcolsep}{3.2pt}
\begin{tabular}{lrrrrr}
\toprule
Model & Canonical & Screen selection & $\Delta_{\mathrm{screen}}$ & $\Delta_{\mathrm{confirm}}$ [95\% CI] & $p_{\mathrm{Holm}}$ \\
\midrule
Qwen2.5-7B & 55.00 & Last 16 layers & +17.50 & +7.50 [-6.25, +21.25] & 0.410 \\
Mistral-7B & 79.38 & Canonical retained & +0.00 & \textemdash & \textemdash \\
Phi-3-mini & 71.88 & Last 16 layers & +6.25 & +7.50 [+0.00, +17.50] & 0.258 \\
\bottomrule
\end{tabular}
\vspace{-1.6em}
\end{table}

Across Qwen, Mistral, and Phi-3, the frozen rank-32, $\alpha{=}64$, last-8,
single-slot configuration remains the most reliable shared operating point.
Backbone-specific screens reveal sensitivity to writer depth, rank, and slot
count, while disjoint confirmation supports retaining the common configuration.
The complete intervention table, all screen cells, and the sensitivity figure
appear in \Cref{app:design_interventions,app:hparam_sensitivity};
frozen-retriever and retrieval-to-QA extensions appear in
\Cref{app:retrievers,app:cqr_qa}.

\paragraph{History contribution and carrier choice.}
A preregistered three-seed confirmation uses 320 independently generated
PSS-Carrier-320 sessions (40 per category), disjoint from the original
benchmark, development, and smoke-test sessions.
Adding ordered history to a
retrieval-supervised PSS improves direct \allrecall{} over a target-only PSS by
5.54 points with BGE-M3 and 10.18 points with BM25; both effects survive Holm
correction and remain positive for every writer seed. Correct states also exceed
matched-wrong and shuffled states by 24.05--39.11 points. Under dense semantic
retrieval, PSS further exceeds a 256-token persistent text memory with matched
writer information and the reported wall-clock write budget by 6.25 points
on this cohort. A separate multi-record experiment evaluates whether one compiled state
can access different historical records in response to future questions.
At $K=16$, PSS reaches 83.75\% and 95.00\% All-Recall@4 on BGE-M3
and BM25, compared with 18.75\% and 35.00\% for all-record
concatenation and 16.25\% and 40.00\% for record-wise RRF.
PSS attains the highest BGE-M3 mean among the evaluated non-reference
methods; on BM25, Grok-based selection reaches 97.50\%.
PSS performs this access through one local parameter-conditioned
generation and one evidence query, without supplying the text bank
to the readout model.
Complete text-access comparisons appear in
\Cref{app:question_conditioned_records};

\section{Related Work}

\paragraph{Compact and parametric memory.}
Long-context models retain historical tokens; StreamingLLM, Dynamic Memory
Compression, SnapKV, and LongLLMLingua reduce their attention or prompt cost
\citep{li2024snapkv,jiang2024longllmlingua},
while AutoCompressor, Gisting, ICAE, and Activation Beacon learn compact token
or activation interfaces
\citep{chevalier2023adapting,mu2023learning,ge2023context,zhang2025long}.
LoRA provides a parameter-efficient carrier \citep{hu2021lora}; Deep Context
Distillation, Cartridges, and Doc-to-LoRA internalize document or corpus content
in parameter/KV modules
\citep{charakorn2026doc}. PSS instead uses
one bounded LoRA state per session to recover retrieval scope after historical
tokens disappear. Exact facts remain external, and the state is unloaded before
answering.

\paragraph{RAG organization and query recovery.}
RAPTOR and HippoRAG organize evidence with hierarchical summaries or graphs
\citep{sarthi2024raptor,gutierrez2024hipporag}. MemoRAG is closest in functional
decomposition because memory generates clues before an expressive reader
\citep{qian2025memorag}; PSS instead stores a per-session parameter delta and
tests it under ambiguous follow-ups and wrong-session interventions.
Conversational rewriters also bridge queries and evidence: CONQRR learns from
retrieval rewards, IterCQR uses retrieval-guided iteration, CHIQ enhances the
history, AdaRewriter applies outcome-supervised selection, and ConvSearch-R1
uses retrieval-aligned reinforcement learning
\citep{wu2022conqrr,mo2024chiq,lai2025adarewriter,zhu2025convsearch}.
They normally reread textual dialogue online; PSS exposes a persistent textual
clue interface without replaying that history. Reproduction and checkpoint
constraints for these baselines are reported in
\Cref{app:cqr_qa,app:retrievers}.

\paragraph{Long-term memory evaluation.}
Conversational rewriting and long-term memory studies examine contextual
query interpretation, indirectly expressed information, and cross-session
reasoning \citep{wu2024longmemeval,maharana2024evaluating}.
Our controlled protocols provide explicit session associations to the
parameter writer and information-matched text controls, allowing us to
study compilation into parameters and subsequent retrieval use.
The public extension preserves LongBench and InfiniteBench source documents
\citep{bai2024longbench} while varying whether the query reveals its active
request; \pss{} supports causal state interventions.

\section{Conclusion}
\label{sec:discussion}

Underspecified follow-ups expose a missing layer in conventional RAG: the
retriever cannot search for session scope that the current query never states.
PSS fills this gap by compiling retrieval supervision and ordered history into a
reusable parametric state that reconstructs what to search for without replaying
historical tokens. Across controlled sessions, public corpora, three
language-model families, and heterogeneous retrieval backends, correct states
produce large, session-attributable retrieval gains. The controlled benchmark
shows a 10.63-point gain over a compact card, and three-seed carrier
controls isolate ordered-history information beyond supplied targets.
PSS therefore provides a retrieval-scoped
memory layer for repeated-use assistants: compile session intent once, reuse it
across RAG systems, and keep exact, versioned knowledge in an external,
auditable evidence store.
 
\subsection*{AI use statement}

Generative AI tools were used to assist with code development, text editing, and translation. AI also contributed to drafting sections of the appendix and assisted in verifying certain datasets and results.

\subsection*{Ethics statement}

\pss{} uses synthetic project names and controlled technical facts, with no human-subject or private user data.

\subsection*{Reproducibility statement}

\Cref{app:setup,app:rq2_details,app:public,app:cross_model,app:carrier_confirmation,app:data,app:training,app:router,app:statistics,app:reproducibility} document the experimental setup, data construction, controls, optimization, statistical analysis, and reproducibility details.

\bibliography{iclr2027_conference}

\begin{thebibliography}{35}
\providecommand{\natexlab}[1]{#1}
\providecommand{\url}[1]{\texttt{#1}}
\expandafter\ifx\csname urlstyle\endcsname\relax
  \providecommand{\doi}[1]{doi: #1}\else
  \providecommand{\doi}{doi: \begingroup \urlstyle{rm}\Url}\fi

\bibitem[Abdin et~al.(2024)Abdin, Aneja, Awadalla, Awadallah, Awan, Bach, Bahree, Bakhtiari, Bao, Behl, Benhaim, Bilenko, Bjorck, Bubeck, Cai, Cai, Chaudhary, Chen, Chen, Chen, Chen, Chen, Cheng, Chopra, Dai, Dixon, Eldan, Fragoso, Gao, Gao, Gao, Garg, Giorno, Goswami, Gunasekar, Haider, Hao, Hewett, Hu, Huynh, Iter, Jacobs, Javaheripi, Jin, Karampatziakis, Kauffmann, Khademi, Kim, Kim, Kurilenko, Lee, Lee, Li, Li, Liang, Liden, Lin, Lin, Liu, Liu, Liu, Liu, Liu, Luo, Madan, Mahmoudzadeh, Majercak, Mazzola, Mendes, Mitra, Modi, Nguyen, Norick, Patra, Perez-Becker, Portet, Pryzant, Qin, Radmilac, Ren, de~Rosa, Rosset, Roy, Ruwase, Saarikivi, Saied, Salim, Santacroce, Shah, Shang, Sharma, Shen, Shukla, Song, Tanaka, Tupini, Vaddamanu, Wang, Wang, Wang, Wang, Wang, Wang, Ward, Wen, Witte, Wu, Wu, Wyatt, Xiao, Xu, Xu, Xu, Xue, Yadav, Yang, Yang, Yang, Yang, Yu, Yuan, Zhang, Zhang, Zhang, Zhang, Zhang, Zhang, Zhang, and Zhou]{abdin2024phi3technicalreporthighly}
Marah Abdin, Jyoti Aneja, Hany Awadalla, Ahmed Awadallah, Ammar~Ahmad Awan, Nguyen Bach, Amit Bahree, Arash Bakhtiari, Jianmin Bao, Harkirat Behl, Alon Benhaim, Misha Bilenko, Johan Bjorck, Sébastien Bubeck, Martin Cai, Qin Cai, Vishrav Chaudhary, Dong Chen, Dongdong Chen, Weizhu Chen, Yen-Chun Chen, Yi-Ling Chen, Hao Cheng, Parul Chopra, Xiyang Dai, Matthew Dixon, Ronen Eldan, Victor Fragoso, Jianfeng Gao, Mei Gao, Min Gao, Amit Garg, Allie~Del Giorno, Abhishek Goswami, Suriya Gunasekar, Emman Haider, Junheng Hao, Russell~J. Hewett, Wenxiang Hu, Jamie Huynh, Dan Iter, Sam~Ade Jacobs, Mojan Javaheripi, Xin Jin, Nikos Karampatziakis, Piero Kauffmann, Mahoud Khademi, Dongwoo Kim, Young~Jin Kim, Lev Kurilenko, James~R. Lee, Yin~Tat Lee, Yuanzhi Li, Yunsheng Li, Chen Liang, Lars Liden, Xihui Lin, Zeqi Lin, Ce~Liu, Liyuan Liu, Mengchen Liu, Weishung Liu, Xiaodong Liu, Chong Luo, Piyush Madan, Ali Mahmoudzadeh, David Majercak, Matt Mazzola, Caio César~Teodoro Mendes, Arindam Mitra, Hardik Modi, Anh Nguyen, Brandon Norick, Barun Patra, Daniel Perez-Becker, Thomas Portet, Reid Pryzant, Heyang Qin, Marko Radmilac, Liliang Ren, Gustavo de~Rosa, Corby Rosset, Sambudha Roy, Olatunji Ruwase, Olli Saarikivi, Amin Saied, Adil Salim, Michael Santacroce, Shital Shah, Ning Shang, Hiteshi Sharma, Yelong Shen, Swadheen Shukla, Xia Song, Masahiro Tanaka, Andrea Tupini, Praneetha Vaddamanu, Chunyu Wang, Guanhua Wang, Lijuan Wang, Shuohang Wang, Xin Wang, Yu~Wang, Rachel Ward, Wen Wen, Philipp Witte, Haiping Wu, Xiaoxia Wu, Michael Wyatt, Bin Xiao, Can Xu, Jiahang Xu, Weijian Xu, Jilong Xue, Sonali Yadav, Fan Yang, Jianwei Yang, Yifan Yang, Ziyi Yang, Donghan Yu, Lu~Yuan, Chenruidong Zhang, Cyril Zhang, Jianwen Zhang, Li~Lyna Zhang, Yi~Zhang, Yue Zhang, Yunan Zhang, and Xiren Zhou.
\newblock Phi-3 technical report: A highly capable language model locally on your phone, 2024.
\newblock URL \url{https://arxiv.org/abs/2404.14219}.

\bibitem[Bai et~al.(2024)Bai, Lv, Zhang, Lyu, Tang, Huang, Du, Liu, Zeng, Hou, et~al.]{bai2024longbench}
Yushi Bai, Xin Lv, Jiajie Zhang, Hongchang Lyu, Jiankai Tang, Zhidian Huang, Zhengxiao Du, Xiao Liu, Aohan Zeng, Lei Hou, et~al.
\newblock Longbench: A bilingual, multitask benchmark for long context understanding.
\newblock In \emph{Proceedings of the 62nd annual meeting of the association for computational linguistics (volume 1: Long papers)}, pp.\  3119--3137, 2024.

\bibitem[Caccia et~al.(2025)Caccia, Ansell, Ponti, Vuli{\'c}, and Sordoni]{caccia2025training}
Lucas Caccia, Alan Ansell, Edoardo Ponti, Ivan Vuli{\'c}, and Alessandro Sordoni.
\newblock Training plug-n-play knowledge modules with deep context distillation.
\newblock \emph{arXiv preprint arXiv:2503.08727}, 2025.

\bibitem[Chan et~al.(2024)Chan, Xu, Yuan, Luo, Xue, Guo, and Fu]{chan2024rq}
Chi-Min Chan, Chunpu Xu, Ruibin Yuan, Hongyin Luo, Wei Xue, Yike Guo, and Jie Fu.
\newblock Rq-rag: Learning to refine queries for retrieval augmented generation.
\newblock \emph{arXiv preprint arXiv:2404.00610}, 2024.

\bibitem[Charakorn et~al.(2026)Charakorn, Cetin, Uesaka, and Lange]{charakorn2026doc}
Rujikorn Charakorn, Edoardo Cetin, Shinnosuke Uesaka, and Robert~Tjarko Lange.
\newblock Doc-to-lora: Learning to instantly internalize contexts.
\newblock \emph{arXiv preprint arXiv:2602.15902}, 2026.

\bibitem[Chevalier et~al.(2023)Chevalier, Wettig, Ajith, and Chen]{chevalier2023adapting}
Alexis Chevalier, Alexander Wettig, Anirudh Ajith, and Danqi Chen.
\newblock Adapting language models to compress contexts.
\newblock In \emph{Proceedings of the 2023 Conference on Empirical Methods in Natural Language Processing}, pp.\  3829--3846, 2023.

\bibitem[Cormack et~al.(2009)Cormack, Clarke, and Buettcher]{cormack2009reciprocal}
Gordon~V Cormack, Charles~LA Clarke, and Stefan Buettcher.
\newblock Reciprocal rank fusion outperforms condorcet and individual rank learning methods.
\newblock In \emph{Proceedings of the 32nd international ACM SIGIR conference on Research and development in information retrieval}, pp.\  758--759, 2009.

\bibitem[Eyuboglu et~al.(2026)Eyuboglu, Ehrlich, Arora, Guha, Zinsley, Liu, Rudra, Zou, Mirhoseini, and Re]{eyuboglu2026cartridges}
Sabri Eyuboglu, Ryan Ehrlich, Simran Arora, Neel Guha, Dylan Zinsley, Emily Liu, Atri Rudra, James~Y Zou, Azalia Mirhoseini, and Christopher Re.
\newblock Cartridges: Lightweight and general-purpose long context representations via self-study.
\newblock In \emph{International Conference on Learning Representations}, volume 2026, pp.\  42642--42687, 2026.

\bibitem[Gao et~al.(2023)Gao, Ma, Lin, and Callan]{gao2023precise}
Luyu Gao, Xueguang Ma, Jimmy Lin, and Jamie Callan.
\newblock Precise zero-shot dense retrieval without relevance labels.
\newblock In \emph{Proceedings of the 61st Annual Meeting of the Association for Computational Linguistics (Volume 1: Long Papers)}, pp.\  1762--1777, 2023.

\bibitem[Ge et~al.(2023)Ge, Hu, Wang, Wang, Chen, and Wei]{ge2023context}
Tao Ge, Jing Hu, Lei Wang, Xun Wang, Si-Qing Chen, and Furu Wei.
\newblock In-context autoencoder for context compression in a large language model.
\newblock \emph{arXiv preprint arXiv:2307.06945}, 2023.

\bibitem[Guti{\'e}rrez et~al.(2024)Guti{\'e}rrez, Shu, Gu, Yasunaga, and Su]{gutierrez2024hipporag}
Bernal~J Guti{\'e}rrez, Yiheng Shu, Yu~Gu, Michihiro Yasunaga, and Yu~Su.
\newblock Hipporag: Neurobiologically inspired long-term memory for large language models.
\newblock \emph{Advances in neural information processing systems}, 37:\penalty0 59532--59569, 2024.

\bibitem[Holm(1979)]{holm1979simple}
Sture Holm.
\newblock A simple sequentially rejective multiple test procedure.
\newblock \emph{Scandinavian journal of statistics}, pp.\  65--70, 1979.

\bibitem[Hu et~al.(2021)Hu, Shen, Wallis, Allen-Zhu, Li, Wang, Wang, and Chen]{hu2021lora}
Edward~J Hu, Yelong Shen, Phillip Wallis, Zeyuan Allen-Zhu, Yuanzhi Li, Shean Wang, Lu~Wang, and Weizhu Chen.
\newblock Lora: Low-rank adaptation of large language models.
\newblock \emph{arXiv preprint arXiv:2106.09685}, 2021.

\bibitem[Hui et~al.(2024)Hui, Yang, Cui, Yang, Liu, Zhang, Liu, Zhang, Yu, Lu, et~al.]{hui2024qwen2}
Binyuan Hui, Jian Yang, Zeyu Cui, Jiaxi Yang, Dayiheng Liu, Lei Zhang, Tianyu Liu, Jiajun Zhang, Bowen Yu, Keming Lu, et~al.
\newblock Qwen2. 5-coder technical report.
\newblock \emph{arXiv preprint arXiv:2409.12186}, 2024.

\bibitem[Izacard et~al.(2021)Izacard, Caron, Hosseini, Riedel, Bojanowski, Joulin, and Grave]{izacard2021unsupervised}
Gautier Izacard, Mathilde Caron, Lucas Hosseini, Sebastian Riedel, Piotr Bojanowski, Armand Joulin, and Edouard Grave.
\newblock Unsupervised dense information retrieval with contrastive learning.
\newblock \emph{arXiv preprint arXiv:2112.09118}, 2021.

\bibitem[Jiang et~al.(2023)Jiang, Sablayrolles, Mensch, Bamford, Chaplot, de~las Casas, Bressand, Lengyel, Lample, Saulnier, Lavaud, Lachaux, Stock, Scao, Lavril, Wang, Lacroix, and Sayed]{jiang2023mistral7b}
Albert~Q. Jiang, Alexandre Sablayrolles, Arthur Mensch, Chris Bamford, Devendra~Singh Chaplot, Diego de~las Casas, Florian Bressand, Gianna Lengyel, Guillaume Lample, Lucile Saulnier, Lélio~Renard Lavaud, Marie-Anne Lachaux, Pierre Stock, Teven~Le Scao, Thibaut Lavril, Thomas Wang, Timothée Lacroix, and William~El Sayed.
\newblock Mistral 7b, 2023.
\newblock URL \url{https://arxiv.org/abs/2310.06825}.

\bibitem[Jiang et~al.(2024)Jiang, Wu, Luo, Li, Lin, Yang, and Qiu]{jiang2024longllmlingua}
Huiqiang Jiang, Qianhui Wu, Xufang Luo, Dongsheng Li, Chin-Yew Lin, Yuqing Yang, and Lili Qiu.
\newblock Longllmlingua: Accelerating and enhancing llms in long context scenarios via prompt compression.
\newblock In \emph{Proceedings of the 62nd Annual Meeting of the Association for Computational Linguistics (Volume 1: Long Papers)}, pp.\  1658--1677, 2024.

\bibitem[Lai et~al.(2025)Lai, Wu, Wang, and Zhou]{lai2025adarewriter}
Yilong Lai, Jialong Wu, Zhenglin Wang, and Deyu Zhou.
\newblock Adarewriter: unleashing the power of prompting-based conversational query reformulation via test-time adaptation.
\newblock In \emph{Proceedings of the 2025 Conference on Empirical Methods in Natural Language Processing}, pp.\  3889--3905, 2025.

\bibitem[Lewis et~al.(2020)Lewis, Perez, Piktus, Petroni, Karpukhin, Goyal, K{\"u}ttler, Lewis, Yih, Rockt{\"a}schel, et~al.]{lewis2020retrieval}
Patrick Lewis, Ethan Perez, Aleksandra Piktus, Fabio Petroni, Vladimir Karpukhin, Naman Goyal, Heinrich K{\"u}ttler, Mike Lewis, Wen-tau Yih, Tim Rockt{\"a}schel, et~al.
\newblock Retrieval-augmented generation for knowledge-intensive nlp tasks.
\newblock \emph{Advances in neural information processing systems}, 33:\penalty0 9459--9474, 2020.

\bibitem[Li et~al.(2024)Li, Huang, Yang, Venkitesh, Locatelli, Ye, Cai, Lewis, and Chen]{li2024snapkv}
Yuhong Li, Yingbing Huang, Bowen Yang, Bharat Venkitesh, Acyr Locatelli, Hanchen Ye, Tianle Cai, Patrick Lewis, and Deming Chen.
\newblock Snapkv: Llm knows what you are looking for before generation.
\newblock \emph{Advances in Neural Information Processing Systems}, 37:\penalty0 22947--22970, 2024.

\bibitem[Maharana et~al.(2024)Maharana, Lee, Tulyakov, Bansal, Barbieri, and Fang]{maharana2024evaluating}
Adyasha Maharana, Dong-Ho Lee, Sergey Tulyakov, Mohit Bansal, Francesco Barbieri, and Yuwei Fang.
\newblock Evaluating very long-term conversational memory of llm agents.
\newblock In \emph{Proceedings of the 62nd Annual Meeting of the Association for Computational Linguistics (Volume 1: Long Papers)}, pp.\  13851--13870, 2024.

\bibitem[Mo et~al.(2024)Mo, Ghaddar, Mao, Rezagholizadeh, Chen, Liu, and Nie]{mo2024chiq}
Fengran Mo, Abbas Ghaddar, Kelong Mao, Mehdi Rezagholizadeh, Boxing Chen, Qun Liu, and Jian-Yun Nie.
\newblock Chiq: Contextual history enhancement for improving query rewriting in conversational search.
\newblock In \emph{Proceedings of the 2024 Conference on Empirical Methods in Natural Language Processing}, pp.\  2253--2268, 2024.

\bibitem[Mu et~al.(2023)Mu, Li, and Goodman]{mu2023learning}
Jesse Mu, Xiang Li, and Noah Goodman.
\newblock Learning to compress prompts with gist tokens.
\newblock \emph{Advances in Neural Information Processing Systems}, 36:\penalty0 19327--19352, 2023.

\bibitem[Multi-Granularity(2024)]{multi2024m3}
Multi-Linguality Multi-Functionality Multi-Granularity.
\newblock M3-embedding: Multi-linguality, multi-functionality, multi-granularity text embeddings through self-knowledge distillation.
\newblock \emph{arXiv preprint arXiv:2402.03216}, 2024.

\bibitem[Nawrot et~al.(2024)Nawrot, {\L}a{\'n}cucki, Chochowski, Tarjan, and Ponti]{nawrot2024dynamic}
Piotr Nawrot, Adrian {\L}a{\'n}cucki, Marcin Chochowski, David Tarjan, and Edoardo~M Ponti.
\newblock Dynamic memory compression: Retrofitting llms for accelerated inference.
\newblock \emph{arXiv preprint arXiv:2403.09636}, 2024.

\bibitem[Qian et~al.(2025)Qian, Liu, Zhang, Mao, Lian, Dou, and Huang]{qian2025memorag}
Hongjin Qian, Zheng Liu, Peitian Zhang, Kelong Mao, Defu Lian, Zhicheng Dou, and Tiejun Huang.
\newblock Memorag: Boosting long context processing with global memory-enhanced retrieval augmentation.
\newblock In \emph{Proceedings of the ACM on Web Conference 2025}, pp.\  2366--2377, 2025.

\bibitem[Sarthi et~al.(2024)Sarthi, Abdullah, Tuli, Khanna, Goldie, and Manning]{sarthi2024raptor}
Parth Sarthi, Salman Abdullah, Aditi Tuli, Shubh Khanna, Anna Goldie, and Christopher Manning.
\newblock Raptor: Recursive abstractive processing for tree-organized retrieval.
\newblock In \emph{International Conference on Learning Representations}, volume 2024, pp.\  32628--32649, 2024.

\bibitem[Tibshirani \& Efron(1993)Tibshirani and Efron]{tibshirani1993introduction}
Robert~J Tibshirani and Bradley Efron.
\newblock An introduction to the bootstrap.
\newblock \emph{Monographs on statistics and applied probability}, 57\penalty0 (1):\penalty0 1--436, 1993.

\bibitem[Wu et~al.(2024)Wu, Wang, Yu, Zhang, Chang, and Yu]{wu2024longmemeval}
Di~Wu, Hongwei Wang, Wenhao Yu, Yuwei Zhang, Kai-Wei Chang, and Dong Yu.
\newblock Longmemeval: Benchmarking chat assistants on long-term interactive memory.
\newblock \emph{arXiv preprint arXiv:2410.10813}, 2024.

\bibitem[Wu et~al.(2022)Wu, Luan, Rashkin, Reitter, Hajishirzi, Ostendorf, and Tomar]{wu2022conqrr}
Zeqiu Wu, Yi~Luan, Hannah Rashkin, David Reitter, Hannaneh Hajishirzi, Mari Ostendorf, and Gaurav~Singh Tomar.
\newblock Conqrr: Conversational query rewriting for retrieval with reinforcement learning.
\newblock In \emph{Proceedings of the 2022 Conference on Empirical Methods in Natural Language Processing}, pp.\  10000--10014, 2022.

\bibitem[Xiao et~al.(2024)Xiao, Tian, Chen, Han, and Lewis]{xiao2024efficient}
Guangxuan Xiao, Yuandong Tian, Beidi Chen, Song Han, and Mike Lewis.
\newblock Efficient streaming language models with attention sinks.
\newblock In \emph{International Conference on Learning Representations}, volume 2024, pp.\  21875--21895, 2024.

\bibitem[Zhang et~al.(2025{\natexlab{a}})Zhang, Liu, Xiao, Shao, Ye, and Dou]{zhang2025long}
Peitian Zhang, Zheng Liu, Shitao Xiao, Ninglu Shao, Qiwei Ye, and Zhicheng Dou.
\newblock Long context compression with activation beacon.
\newblock In \emph{International Conference on Learning Representations}, volume 2025, pp.\  101932--101948, 2025{\natexlab{a}}.

\bibitem[Zhang et~al.(2024)Zhang, Chen, Hu, Xu, Chen, Hao, Han, Thai, Wang, Liu, et~al.]{zhang2024infty}
Xinrong Zhang, Yingfa Chen, Shengding Hu, Zihang Xu, Junhao Chen, Moo~Khai Hao, Xu~Han, Zhen~Leng Thai, Shuo Wang, Zhiyuan Liu, et~al.
\newblock \(\infty\) bench: Extending long context evaluation beyond 100k tokens.
\newblock \emph{arXiv preprint arXiv:2402.13718}, 2024.

\bibitem[Zhang et~al.(2025{\natexlab{b}})Zhang, Li, Long, Zhang, Lin, Yang, Xie, Yang, Liu, Lin, et~al.]{zhang2025qwen3}
Yanzhao Zhang, Mingxin Li, Dingkun Long, Xin Zhang, Huan Lin, Baosong Yang, Pengjun Xie, An~Yang, Dayiheng Liu, Junyang Lin, et~al.
\newblock Qwen3 embedding: Advancing text embedding and reranking through foundation models.
\newblock \emph{arXiv preprint arXiv:2506.05176}, 2025{\natexlab{b}}.

\bibitem[Zhu et~al.(2025)Zhu, Wang, Feng, Song, and Qiu]{zhu2025convsearch}
Changtai Zhu, Siyin Wang, Ruijun Feng, Kai Song, and Xipeng Qiu.
\newblock Convsearch-r1: Enhancing query reformulation for conversational search with reasoning via reinforcement learning.
\newblock In \emph{Proceedings of the 2025 Conference on Empirical Methods in Natural Language Processing}, pp.\  26558--26575, 2025.

\end{thebibliography}


\begin{thebibliography}{29}
\providecommand{\natexlab}[1]{#1}
\providecommand{\url}[1]{\texttt{#1}}
\expandafter\ifx\csname urlstyle\endcsname\relax
  \providecommand{\doi}[1]{doi: #1}\else
  \providecommand{\doi}{doi: \begingroup \urlstyle{rm}\Url}\fi

\bibitem[Abdin et~al.(2024)Abdin, Aneja, Awadalla, Awadallah, Awan, Bach, Bahree, Bakhtiari, Bao, Behl, et~al.]{abdin2024phi3}
Marah Abdin, Jyoti Aneja, Hany Awadalla, Ahmed Awadallah, Ammar~Ahmad Awan, Nguyen Bach, Amit Bahree, Arash Bakhtiari, Jianmin Bao, Harkirat Behl, et~al.
\newblock {Phi-3} technical report: A highly capable language model locally on your phone.
\newblock \emph{arXiv preprint arXiv:2404.14219}, 2024.

\bibitem[Bai et~al.(2024)Bai, Lv, Zhang, Lyu, Tang, Huang, Du, Liu, Zeng, Hou, Dong, Tang, and Li]{bai2024longbench}
Yushi Bai, Xin Lv, Jiajie Zhang, Hongchang Lyu, Jiankai Tang, Zhidian Huang, Zhengxiao Du, Xiao Liu, Aohan Zeng, Lei Hou, Yuxiao Dong, Jie Tang, and Juanzi Li.
\newblock {LongBench}: A bilingual, multitask benchmark for long context understanding.
\newblock In \emph{Annual Meeting of the Association for Computational Linguistics}, pp.\  3119--3137, 2024.

\bibitem[Caccia et~al.(2025)Caccia, Ansell, Ponti, Vuli{\'c}, and Sordoni]{caccia2025dcd}
Lucas Caccia, Alan Ansell, Edoardo Ponti, Ivan Vuli{\'c}, and Alessandro Sordoni.
\newblock Training plug-n-play knowledge modules with deep context distillation.
\newblock In \emph{Conference on Language Modeling}, 2025.

\bibitem[Charakorn et~al.(2026)Charakorn, Cetin, Uesaka, and Lange]{charakorn2026d2l}
Rujikorn Charakorn, Edoardo Cetin, Shinnosuke Uesaka, and Robert~Tjarko Lange.
\newblock Doc-to-{LoRA}: Learning to instantly internalize contexts.
\newblock \emph{arXiv preprint arXiv:2602.15902}, 2026.

\bibitem[Chevalier et~al.(2023)Chevalier, Wettig, Ajith, and Chen]{chevalier2023autocompressor}
Alexis Chevalier, Alexander Wettig, Anirudh Ajith, and Danqi Chen.
\newblock Adapting language models to compress contexts.
\newblock In \emph{Conference on Empirical Methods in Natural Language Processing}, 2023.

\bibitem[Cormack et~al.(2009)Cormack, Clarke, and Buettcher]{cormack2009rrf}
Gordon~V. Cormack, Charles L.~A. Clarke, and Stefan Buettcher.
\newblock Reciprocal rank fusion outperforms condorcet and individual rank learning methods.
\newblock In \emph{International ACM SIGIR Conference on Research and Development in Information Retrieval}, pp.\  758--759, 2009.

\bibitem[Efron \& Tibshirani(1994)Efron and Tibshirani]{efron1994bootstrap}
Bradley Efron and Robert~J. Tibshirani.
\newblock \emph{An Introduction to the Bootstrap}.
\newblock Chapman and Hall/CRC, 1994.

\bibitem[Eyuboglu et~al.(2025)Eyuboglu, Ehrlich, Arora, Guha, Zinsley, Liu, Tennien, Rudra, Zou, Mirhoseini, and R{\'e}]{eyuboglu2025cartridges}
Sabri Eyuboglu, Ryan Ehrlich, Simran Arora, Neel Guha, Dylan Zinsley, Emily Liu, Will Tennien, Atri Rudra, James Zou, Azalia Mirhoseini, and Christopher R{\'e}.
\newblock Cartridges: Lightweight and general-purpose long context representations via self-study.
\newblock \emph{arXiv preprint arXiv:2506.06266}, 2025.

\bibitem[Ge et~al.(2024)Ge, Hu, Wang, Wang, Chen, and Wei]{ge2024icae}
Tao Ge, Jing Hu, Lei Wang, Xun Wang, Si-Qing Chen, and Furu Wei.
\newblock In-context autoencoder for context compression in a large language model.
\newblock In \emph{International Conference on Learning Representations}, 2024.

\bibitem[Guti{\'e}rrez et~al.(2024)Guti{\'e}rrez, Shu, Gu, Yasunaga, and Su]{gutierrez2024hipporag}
Bernal~Jim{\'e}nez Guti{\'e}rrez, Yiheng Shu, Yu~Gu, Michihiro Yasunaga, and Yu~Su.
\newblock {HippoRAG}: Neurobiologically inspired long-term memory for large language models.
\newblock In \emph{Advances in Neural Information Processing Systems}, 2024.

\bibitem[Holm(1979)]{holm1979}
Sture Holm.
\newblock A simple sequentially rejective multiple test procedure.
\newblock \emph{Scandinavian Journal of Statistics}, 6\penalty0 (2):\penalty0 65--70, 1979.

\bibitem[Hu et~al.(2022)Hu, Shen, Wallis, Allen-Zhu, Li, Wang, Wang, and Chen]{hu2022lora}
Edward~J. Hu, Yelong Shen, Phillip Wallis, Zeyuan Allen-Zhu, Yuanzhi Li, Shean Wang, Lu~Wang, and Weizhu Chen.
\newblock {LoRA}: Low-rank adaptation of large language models.
\newblock In \emph{International Conference on Learning Representations}, 2022.

\bibitem[Jiang et~al.(2023)Jiang, Sablayrolles, Mensch, Bamford, Chaplot, de~las Casas, Bressand, Lengyel, Lample, Saulnier, Lavaud, Lachaux, Stock, Le~Scao, Lavril, Wang, Lacroix, and El~Sayed]{jiang2023mistral}
Albert~Q. Jiang, Alexandre Sablayrolles, Arthur Mensch, Chris Bamford, Devendra~Singh Chaplot, Diego de~las Casas, Florian Bressand, Gianna Lengyel, Guillaume Lample, Lucile Saulnier, L{\'e}lio~Renard Lavaud, Marie-Anne Lachaux, Pierre Stock, Teven Le~Scao, Thibaut Lavril, Thomas Wang, Timoth{\'e}e Lacroix, and William El~Sayed.
\newblock Mistral 7b.
\newblock \emph{arXiv preprint arXiv:2310.06825}, 2023.

\bibitem[Jiang et~al.(2024)Jiang, Wu, Luo, Li, Lin, Yang, and Qiu]{jiang2024longllmlingua}
Huiqiang Jiang, Qianhui Wu, Xufang Luo, Dongsheng Li, Chin-Yew Lin, Yuqing Yang, and Lili Qiu.
\newblock {LongLLMLingua}: Accelerating and enhancing {LLM}s in long context scenarios via prompt compression.
\newblock In \emph{Annual Meeting of the Association for Computational Linguistics}, 2024.

\bibitem[Lewis et~al.(2020)Lewis, Perez, Piktus, Petroni, Karpukhin, Goyal, K{\"u}ttler, Lewis, Yih, Rockt{\"a}schel, Riedel, and Kiela]{lewis2020rag}
Patrick Lewis, Ethan Perez, Aleksandra Piktus, Fabio Petroni, Vladimir Karpukhin, Naman Goyal, Heinrich K{\"u}ttler, Mike Lewis, Wen-tau Yih, Tim Rockt{\"a}schel, Sebastian Riedel, and Douwe Kiela.
\newblock Retrieval-augmented generation for knowledge-intensive {NLP} tasks.
\newblock In \emph{Advances in Neural Information Processing Systems}, volume~33, 2020.

\bibitem[Li et~al.(2024)Li, Huang, Yang, Venkitesh, Locatelli, Ye, Cai, Lewis, and Chen]{li2024snapkv}
Yuhong Li, Yingbing Huang, Bowen Yang, Bharat Venkitesh, Acyr Locatelli, Hanchen Ye, Tianle Cai, Patrick Lewis, and Deming Chen.
\newblock {SnapKV}: {LLM} knows what you are looking for before generation.
\newblock \emph{arXiv preprint arXiv:2404.14469}, 2024.

\bibitem[Maharana et~al.(2024)Maharana, Lee, Tulyakov, Bansal, Barbieri, and Fang]{maharana2024locomo}
Adyasha Maharana, Dong-Ho Lee, Sergey Tulyakov, Mohit Bansal, Francesco Barbieri, and Yuwei Fang.
\newblock Evaluating very long-term conversational memory of {LLM} agents.
\newblock \emph{arXiv preprint arXiv:2402.17753}, 2024.

\bibitem[McNemar(1947)]{mcnemar1947}
Quinn McNemar.
\newblock Note on the sampling error of the difference between correlated proportions or percentages.
\newblock \emph{Psychometrika}, 12\penalty0 (2):\penalty0 153--157, 1947.

\bibitem[Mu et~al.(2023)Mu, Li, and Goodman]{mu2023gisting}
Jesse Mu, Xiang~Lisa Li, and Noah Goodman.
\newblock Learning to compress prompts with gist tokens.
\newblock In \emph{Advances in Neural Information Processing Systems}, volume~36, 2023.

\bibitem[Nawrot et~al.(2024)Nawrot, {\L}a{\'n}cucki, Chochowski, Tarjan, and Ponti]{nawrot2024dmc}
Piotr Nawrot, Adrian {\L}a{\'n}cucki, Marcin Chochowski, David Tarjan, and Edoardo~M. Ponti.
\newblock Dynamic memory compression: Retrofitting {LLM}s for accelerated inference.
\newblock In \emph{International Conference on Machine Learning}, pp.\  37396--37412, 2024.

\bibitem[Qian et~al.(2025)Qian, Liu, Zhang, Mao, Lian, Dou, and Huang]{qian2025memorag}
Hongjin Qian, Zheng Liu, Peitian Zhang, Kelong Mao, Defu Lian, Zhicheng Dou, and Tiejun Huang.
\newblock {MemoRAG}: Boosting long context processing with global memory-enhanced retrieval augmentation.
\newblock In \emph{The Web Conference}, 2025.

\bibitem[{Qwen Team}(2025)]{qwen2025report}
{Qwen Team}.
\newblock {Qwen2.5} technical report.
\newblock \emph{arXiv preprint arXiv:2412.15115}, 2025.

\bibitem[Robertson \& Zaragoza(2009)Robertson and Zaragoza]{robertson2009bm25}
Stephen Robertson and Hugo Zaragoza.
\newblock The probabilistic relevance framework: {BM25} and beyond.
\newblock \emph{Foundations and Trends in Information Retrieval}, 3\penalty0 (4):\penalty0 333--389, 2009.

\bibitem[Sarthi et~al.(2024)Sarthi, Abdullah, Tuli, Khanna, Goldie, and Manning]{sarthi2024raptor}
Parth Sarthi, Salman Abdullah, Aditi Tuli, Shubh Khanna, Anna Goldie, and Christopher~D. Manning.
\newblock {RAPTOR}: Recursive abstractive processing for tree-organized retrieval.
\newblock In \emph{International Conference on Learning Representations}, 2024.

\bibitem[Wu et~al.(2025)Wu, Wang, Yu, Zhang, Chang, and Yu]{wu2025longmemeval}
Di~Wu, Hongwei Wang, Wenhao Yu, Yuwei Zhang, Kai-Wei Chang, and Dong Yu.
\newblock {LongMemEval}: Benchmarking chat assistants on long-term interactive memory.
\newblock In \emph{International Conference on Learning Representations}, 2025.

\bibitem[Wu et~al.(2022)Wu, Luan, Rashkin, Reitter, Hajishirzi, Ostendorf, and Tomar]{wu2022conqrr}
Zeqiu Wu, Yi~Luan, Hannah Rashkin, David Reitter, Hannaneh Hajishirzi, Mari Ostendorf, and Gaurav~Singh Tomar.
\newblock {CONQRR}: Conversational query rewriting for retrieval with reinforcement learning.
\newblock In \emph{Conference on Empirical Methods in Natural Language Processing}, 2022.

\bibitem[Xiao et~al.(2024)Xiao, Tian, Chen, Han, and Lewis]{xiao2024streamingllm}
Guangxuan Xiao, Yuandong Tian, Beidi Chen, Song Han, and Mike Lewis.
\newblock Efficient streaming language models with attention sinks.
\newblock In \emph{International Conference on Learning Representations}, 2024.

\bibitem[Zhang et~al.(2024{\natexlab{a}})Zhang, Liu, Xiao, Shao, Ye, and Dou]{zhang2024activation}
Peitian Zhang, Zheng Liu, Shitao Xiao, Ninglu Shao, Qiwei Ye, and Zhicheng Dou.
\newblock Long context compression with activation beacon.
\newblock \emph{arXiv preprint arXiv:2401.03462}, 2024{\natexlab{a}}.

\bibitem[Zhang et~al.(2024{\natexlab{b}})Zhang, Chen, Hu, Xu, Chen, Hao, Han, Thai, Wang, Liu, and Sun]{zhang2024infinitebench}
Xinrong Zhang, Yingfa Chen, Shengding Hu, Zihang Xu, Junhao Chen, Moo~Khai Hao, Xu~Han, Zhen~Leng Thai, Shuo Wang, Zhiyuan Liu, and Maosong Sun.
\newblock {$\infty$Bench}: Extending long context evaluation beyond 100k tokens.
\newblock In \emph{Annual Meeting of the Association for Computational Linguistics}, pp.\  15262--15277, 2024{\natexlab{b}}.

\end{thebibliography}
\bibliographystyle{iclr2027_conference}

\appendix
\newpage
\section{Formal Task Contract and Algorithms}
\label{app:formal_algorithms}

\Cref{eq:problem_pipeline} separates the system into four stages so that the
intervention and its outputs are unambiguous. The writer changes only a
session-local adapter; online clue generation temporarily uses that adapter;
retrieval uses an unchanged external index; and answering uses the unloaded,
frozen base model. \Cref{tab:formal_contract} records the information boundary
at each stage.
The main paired comparison holds $\mathcal{D}$, $R_{\phi}$, $k$, and the
reader fixed while replacing only the query source: raw $q_s$, a textual card,
the correct PSS clue, a matched-wrong clue, or a shuffled-state clue. Hence an
evidence-retrieval gain is attributable to recovered session intent rather
than a different index, retriever, or answer model. The retrieved chunk IDs
and their order are additionally fixed in the answer-stage ablation, isolating
whether a loaded state contributes anything beyond retrieval.

\begin{table}[H]
\centering
\footnotesize
\setlength{\tabcolsep}{3pt}
\begin{tabularx}{\linewidth}{@{}l>{\raggedright\arraybackslash}X>{\raggedright\arraybackslash}X>{\raggedright\arraybackslash}X>{\raggedright\arraybackslash}X@{}}
\toprule
Stage & Available inputs & Mutable object & Output & Explicitly unavailable \\
\midrule
Offline write & Ordered $H_s$, frozen $\theta$, adapter contract $\mathcal{C}_{\theta}$;  clue supervision during training & Session factors $\Delta_s$ only & Serialized bounded state $\Delta_s$ & Formal answers and target-document IDs \\
Online clue & Current $q_s$, $\Delta_s$, frozen $\theta$ & None &  clue $\tilde q_s$ & Historical tokens and gold evidence \\
Retrieval & $\tilde q_s$, shared $\mathcal{D}$, frozen $R_{\phi}$ & None & Ranked evidence $E_s$ & Gold target labels \\
Answer & $q_s$, retrieved $E_s$, frozen $p_{\theta}$ & None & Final response $y_s$ & $H_s$ and loaded $\Delta_s$ \\
\bottomrule
\end{tabularx}
\caption{Stage-wise PSS information and mutation contract.}
\label{tab:formal_contract}
\end{table}

\section{Experimental Setup and Sessionization Details}
\label{app:setup}

\subsection{Evaluation rationale and method vocabulary}
\label{app:evaluation_design}

Our primary evaluation isolates the capability PSS is designed to provide:
recovering retrieval-critical session scope that is absent from the current
query. We remove those terms without placing answers in the writer input,
hold the corpus, index, retriever, and reader fixed, and vary only the query
source or session-state condition. Success therefore measures whether a
reusable state enables an otherwise unchanged RAG pipeline to retrieve the
intended evidence.

We build a three-layer evidence chain.  sessionizations of public
tasks test transfer beyond the controlled domain. \pss{} places many sessions
in a shared candidate corpus and uses matched-wrong, shuffled, zero, null, and
OOD states to identify session-specific effects. Frozen-retriever,
cross-backbone, and same-evidence comparisons then test portability and localize
the gain to retrieval. Together, these experiments distinguish recovered
session intent from generic query improvement or answer-stage memorization.

\paragraph{Stable method vocabulary.}
We use three names throughout the primary evaluations. \emph{PSS clue
retrieval} loads the correct session state, emits an  clue, and
retrieves using that clue alone; it is the core mechanism. \emph{Raw + PSS RRF}
is an unconditional fusion ablation of raw and PSS-clue rankings.
\emph{Guarded PSS} is the complete three-way system whose router chooses raw,
card, or PSS-RRF. Correct, wrong, shuffled, and zero identify causal state
conditions rather than additional methods. On the public transformation,
the three-way arm includes the parsed active-request reference among its candidate
actions. Its transfer and safety analysis is consolidated in
\Cref{app:limitations}. The distinct-record extension in
\Cref{app:question_conditioned_records} separately uses the original question
concatenated with generated or selected retrieval text.

\begin{algorithm}[H]
\caption{Offline compilation of one session state}
\label{alg:pss_write}
\begin{algorithmic}
\Require Frozen base model $\theta$; adapter contract $\mathcal{C}_{\theta}$;
ordered history $H_s$;  positive and negative clue targets
$(z_s^{+},z_s^{-})$; state budget $B$; registered phase lengths
$T_{\mathrm{LM}},T_{\mathrm{read}}$ and learning rates
\Ensure Serialized session state $\Delta_s$
\State Initialize LoRA factors $\Delta_s$ permitted by $\mathcal{C}_{\theta}$;
freeze all parameters in $\theta$
\State Split $H_s$ into bounded chunks while preserving their original order
\State Initialize the LM-phase optimizer on $\Delta_s$
\For{$t=1,\ldots,T_{\mathrm{LM}}$}
    \State $x_t\gets$ the next ordered history chunk
    \State Update $\Delta_s$ using
    $\nabla_{\Delta_s}\mathcal{L}_{\mathrm{LM}}(x_t;\theta+\Delta_s)$
\EndFor
\State Retain $\Delta_s$; initialize a new readout-phase optimizer
\State $c_0\gets$ the fixed session-clue instruction
\For{$t=1,\ldots,T_{\mathrm{read}}$}
    \State $\ell_+\gets-\log p_{\theta+\Delta_s}(z_s^+\mid c_0)/|z_s^+|$
    \State $\ell_-\gets-\log p_{\theta+\Delta_s}(z_s^-\mid c_0)/|z_s^-|$
    \State $\mathcal{L}_{\mathrm{read}}\gets\ell_++\max(0,m+\ell_+-\ell_-)$
    \State Update $\Delta_s$ only using $\nabla_{\Delta_s}\mathcal{L}_{\mathrm{read}}$
\EndFor
\State Verify adapter topology, dtype, finite tensors, and
$\operatorname{bytes}(\Delta_s)\leq B$
\State Serialize $\Delta_s$ with its model, writer, seed, and configuration metadata
\State \Return $\Delta_s$
\end{algorithmic}
\end{algorithm}

\begin{algorithm}[H]
\caption{Online PSS clue retrieval and state-isolated answering}
\label{alg:pss_online}
\begin{algorithmic}
\Require Current query $q_s$; state $\Delta_s$; shared corpus $\mathcal{D}$;
frozen retriever $R_{\phi}$; frozen base model $p_{\theta}$; cutoff $k$;
optional guard $g$
\Ensure Retrieval clue $\tilde q_s$, evidence $E_s$, and response $y_s$
\State Temporarily load $\Delta_s$ into the base model
\State $\tilde q_s\gets G_{\theta+\Delta_s}(q_s)$ \Comment{ search clue}
\State $r_{\mathrm{PSS}}\gets R_{\phi}(\tilde q_s,\mathcal{D})$
\If{guard $g$ is enabled}
    \State Compute raw and card rankings and deployment-observable router features
    \State $a_s\gets g(q_s,\tilde q_s,\text{ranking features})$
    \State $E_s\gets\operatorname{TopK}_{k}$ of the raw, card, or raw--PSS RRF ranking selected by $a_s$
\Else
    \State $E_s\gets\operatorname{TopK}_{k}(r_{\mathrm{PSS}})$
\EndIf
\State Unload $\Delta_s$ and restore the frozen base model $\theta$
\State $y_s\sim p_{\theta}(y\mid q_s,E_s)$
\State \Return $(\tilde q_s,E_s,y_s)$
\end{algorithmic}
\end{algorithm}

\subsection{Why session-intent memory benefits from a different evaluation}
\label{app:why_pss320}

\paragraph{The identification gap in conventional QA.}
A standard long-context QA item gives the model an information-sufficient
question: the relevant entities, topic, relation, or time frame normally appear
in the current input. Such an item can measure document retrieval and reading,
but it does not test the failure mode in \Cref{fig:rag_failure}. A retriever may
succeed without recovering any session state because the query already says
what to search for. Conversely, an end-to-end answer score does not reveal
whether a failure arose from interpreting the session, retrieving evidence, or
decoding the answer. Merely prepending more history also changes the online
token budget and gives a long-context model direct access to the very
information that a persistent memory is supposed to recover.

\paragraph{Conditions for causal identification.}
The claim tested by \method{} is narrower and causal: given an underspecified
follow-up, does the state written from session $s$ restore the missing retrieval
intent of that same session? Our evaluation instantiates the following
conditions. First, the online query omits decisive project, entity, relation,
or version terms while remaining resolvable from the ordered history. Second,
the writer has no access to the answer, gold document ID, or evaluation label.
Third, test sessions retrieve from a shared, locked candidate corpus; choosing
a separate index for each session would reveal the active project before the
query is interpreted. Fourth, correct-state results are paired with
matched-wrong, shuffled, and zero-state controls, reducing the chance that a
generic rewrite or regularization effect is mistaken for session memory.
Finally, null, OOD, revision, and conflict probes expose
false activation and stale-scope failures, while session-disjoint splits and
session-level inference prevent probes from the same history from being
treated as independent evidence.

\begin{table}[H]
\centering
\small
\setlength{\tabcolsep}{4pt}
\begin{tabularx}{\linewidth}{@{}l>{\raggedright\arraybackslash}X>{\raggedright\arraybackslash}X>{\raggedright\arraybackslash}X@{}}
\toprule
Property & Conventional QA & Public sessionization & \pss{} \\
\midrule
Current query & Usually self-contained &  ambiguous follow-up & Controlled ambiguous and compositional probes \\
Candidate corpus & Task document(s) & Released public documents & Shared locked corpus across test sessions \\
Primary attribution & Retrieval plus reading & Correct vs. wrong/shuffled session & Correct vs. wrong/shuffled/zero state \\
Robustness probes & Rarely explicit & Null and OOD variants & Null/OOD, revision, and conflict variants \\
Main role & General task performance & External validity & Causal mechanism identification \\
\bottomrule
\end{tabularx}
\caption{Evaluation roles and identification conditions. All PSS writer inputs
exclude answers and target-document identifiers.}
\label{tab:why_pss320}
\end{table}

\paragraph{Why two complementary tracks are retained.}
 sessionizations of released benchmarks preserve real source
documents, established metrics, and cross-domain variation, but some instances
remain easy or reach single-document ceilings. \pss{} supplies the missing
experimental control: balanced ambiguity types, revision structure, evidence
groups, causal state interventions, and a shared cross-session retrieval pool.
It is therefore used to identify the session-intent mechanism rather than to
replace public benchmark evaluation. The public and controlled tracks answer
different questions---whether the method transfers to released data, and
whether its gains are specifically caused by recovering the corresponding
session's omitted intent. Construction, human-audit, and leakage-check details
appear in \Cref{app:data,app:statistics}.

\subsection{Benchmarks and sessionization}

\paragraph{Public sessionized ambiguity.}
\label{sec:public_protocol}
We derive 326  sessions from the QA portions of NarrativeQA,
Qasper, MultiFieldQA-en, HotpotQA, 2WikiMQA, MuSiQue, and InfiniteBench En.QA.
Each history contains up to four released queries and marks one as active; the
online input is a fixed underspecified continuation. Documents, answers, and
target IDs are hidden from the writer. Seed fixes 32 guard-fit, 32
confirmation, and 262 sealed-test sessions before core evaluation. Every
session has correct, same-task wrong-history, shuffled-history, null, and OOD
variants; the audited run writes all 652 required states with zero failures.
Retrieval uses BGE-M3 and target-document All-Recall@3. The \emph{parsed
active-request reference} deterministically extracts and persists the
complete request designated by the history's activity marker. It can be
executed when these explicit markers are available and serves as a text
reference with resolved request selection. The same request supplies
PSS's positive readout target. Both paths therefore receive explicit
request associations: this protocol evaluates retrieval access under
shared supervision, rather than intent identification in unmarked
natural conversations. This reference is distinct from the compact
structured card used on \pss{}.

\paragraph{UltraDomain.}
\label{sec:cross_model_design}
A separate 112-session  transformation tests transfer to finance,
legal, and miscellaneous documents. The original released QA task contains no
session history, so PSS is inapplicable; long-context and BGE-M3 results on the
released task are descriptive references reported in \Cref{app:cross_model}.

\paragraph{Controlled sessions (\pss{}).}
Public document QA is externally grounded but can become easy after
sessionization. We therefore use \pss{}, 320 AI-controlled synthetic
technical-project sessions and 1,920 probes. We reserve 64 sessions for
state/router development and report a source-group-disjoint test of 256
sessions (1,536 probes). Each test session contains canonical, paraphrase,
anaphora, composition, exact-fact, and null/conflict probes and specifies every
evidence group needed for success. The generator balances eight semantic
categories and multiple history lengths, topic counts, revisions, and relation
depths. Automated audits verify split disjointness, unique answers, and
evidence alignment. A completed stratified human comparison additionally
checks history coherence, intended scope, evidence sufficiency, null validity,
and exclusion of formal-answer supervision. Because these sessions remain
synthetic, \pss{} is a controlled mechanism benchmark; the construction contract is in \Cref{app:data}.

\subsection{Models, controls, and evaluation}

Before portability evaluation, we froze the writer objective, rank, topology,
precision, splits, comparisons, and multiplicity families. Qwen2.5-7B,
Mistral-7B, and Phi-3-mini each write architecture-specific states; no tensor
is shared across models. The primary study uses Qwen2.5-7B-Instruct in BF16
with greedy decoding, BM25, 192-token chunks, and $k=4$; interaction studies
add BGE-small, BGE-M3, summary-leaf, and entity-relation retrieval. Baselines
include raw query, recent context, structured card, summary, full-history
rewrite, Raw + PSS RRF, ConvSearch-R1, gold query/evidence, and capped full
context. Every state-bearing arm receives correct, matched-wrong,
parameter-shuffled, and zero states. The answer ablation holds the query,
retrieved chunk IDs, and chunk order fixed while varying only the decoding
state. \Cref{app:arms,app:answer} lists every arm and diagnostic.

Calibration selected a retrieval-aligned, rank-32, single-slot,
full-ordered-history BF16 state. The primary \pss{} result tables, together
with the state-fingerprint, functional-alignment, and Qwen query-bank diagnostics, aggregate
the independent H200 writer runs with seeds 91--93 under the same frozen
configuration and controls. The Mistral/Phi portability, 112-session
strong-backend confirmation, 320-session carrier confirmation, and public
transfer runs likewise report writer seeds 91--93 under their separate frozen
registrations; results are not pooled across protocols. We retain the session,
rather than probes or query variants, as the unit of inference. The primary endpoints are evidence \allrecall{} on
\pss{} and target-document All-Recall@3 on public data. Statistical reporting uses paired
category-stratified bootstrap intervals with 10,000 resamples, exact McNemar
or sign-flip tests as registered, and Holm correction. A dagger is generated
only when all registered positive comparisons against declared baselines have
Holm-adjusted $p<0.05$; complete gates and deviations appear in
\Cref{app:protocol,app:statistics}.

For visual reporting, tables use neutral backgrounds and script-computed bold
and underline for best and second-best performance. RQ1 figures use blue for
PSS clue/fusion arms, orange for the guarded system or router diagnostics, and
gray for baselines and controls. Color indicates experimental role rather than
statistical significance.

\IfFileExists{generated_state_geometry.tex}{%
\begin{figure}[H]
\centering
\includegraphics[width=\linewidth]{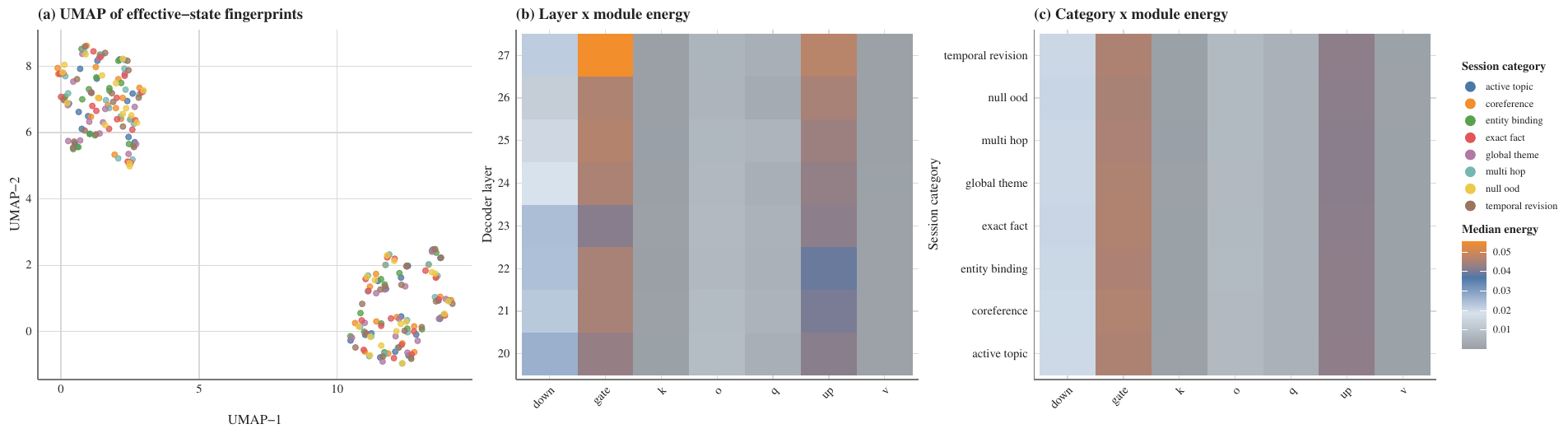}
\caption{Factorization-invariant geometry for 256 held-out sessions, aggregated
over independently written Qwen PSS updates from writer seeds 91--93:
descriptive UMAP and median effective-update energy heatmaps.}
\label{fig:state_geometry}
\end{figure}

\paragraph{State-geometry result.}
In the preregistered effective-update fingerprint space, the category cosine silhouette is -0.0090. The nearest neighbor shares its semantic category for 1.17\% of sessions, versus a 12.16\% finite-sample chance rate (enrichment -10.98 points; 10,000-label-permutation $p=1$). These statistics are computed before UMAP projection; \Cref{fig:state_geometry} is descriptive.

}{%
\paragraph{State-geometry artifact unavailable.}
The frozen result is omitted unless its generated fragment and audited figure
are present in the paper workspace.
}

\section{State Fingerprints and Functional-Output Alignment}
\label{app:rq2_details}

Raw LoRA $A/B$ entries are not identifiable because rescaling $A$ and
inversely rescaling $B$ leaves the effective update unchanged. We therefore
preregister a factorization-invariant analysis of the 256 correct Qwen states.
For each layer--module pair, we compute
$\lVert BA\rVert_F^2=\mathrm{tr}(B^\top BAA^\top)$ and deterministic
Rademacher projections $u^\top BAv$. The standardized projections form a state
fingerprint. Category silhouette and same-category nearest-neighbor enrichment
are evaluated before projection with a 10,000-label-permutation test;
two-dimensional UMAP is used only for visualization. Layer--module and
category--module heatmaps report median within-state energy shares.

The functional-alignment follow-up was frozen only after observing the
coarse-category clustering null. BGE-M3 embeds the raw query, the clue produced by the correct
state, matched-wrong and shuffled-state clues, and the gold retrieval-intent
rewrite. The gold rewrite is an evaluation target only; neither answer strings
nor answer predictions enter the analysis. Primary comparisons are
session-level averages over five answerable probes, tested with one-sided sign
flips and Holm correction. All effects are computed in the original
1,024-dimensional space rather than in UMAP. Since the state is trained with a
retrieval-clue objective, these tests validate output behavior and session
specificity; they are not an intrinsic probe of relational parameter geometry.

\IfFileExists{generated_relational_trajectory.tex}{%
\begin{figure}[H]
\centering
\includegraphics[width=\linewidth]{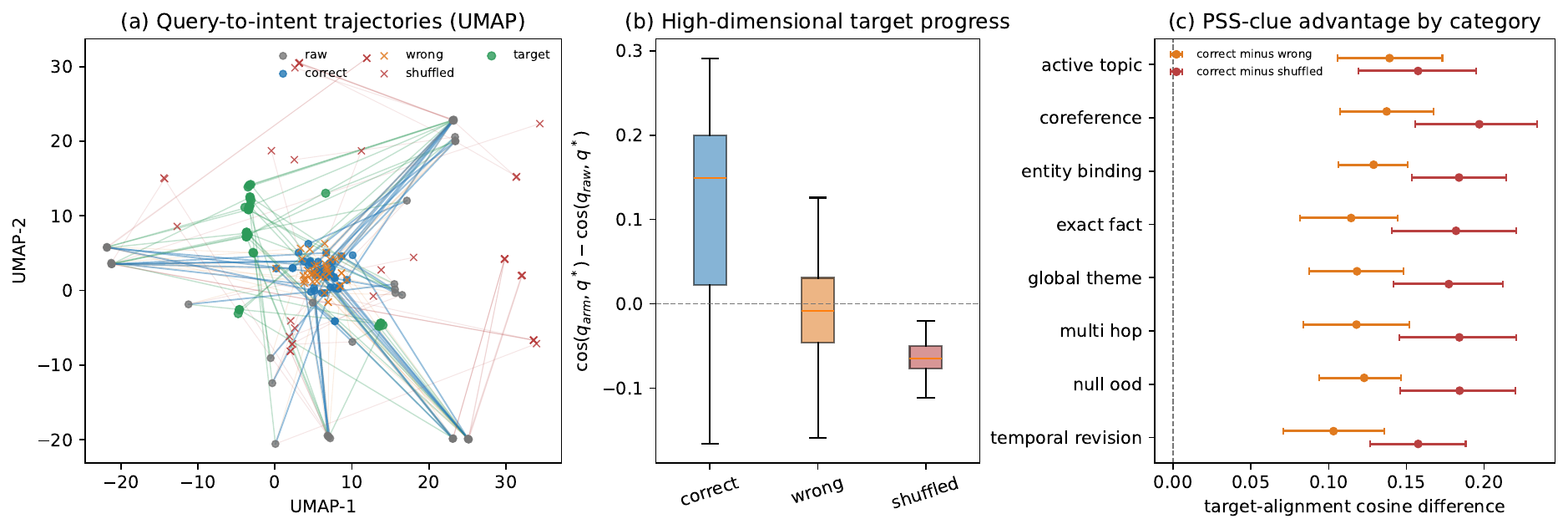}
\caption{Generated-clue trajectories and target-alignment effects. UMAP is descriptive; effects are computed in the original 1,024-dimensional BGE-M3 space.}
\label{fig:pss_relational_trajectory}
\end{figure}

\paragraph{Registered functional-output alignment follow-up.}
Motivated by the frozen coarse-category clustering null, we measured whether the clue generated by a state moves an ambiguous query toward its gold retrieval-intent rewrite. Across 256 sessions and writer seeds 91--93, PSS-clue progress over raw was +0.1136 (95\% CI [+0.1008, +0.1266]; Holm $p=0.00029997$). PSS-clue progress exceeded matched-wrong and shuffled controls by +0.1227 and +0.1779, respectively (Holm $p=0.00029997$ and $p=0.00029997$). All tests aggregate writer seeds before session-level inference; answers and answer predictions are not used. These effects validate session-specific output alignment, not relational geometry of the underlying parameters.

}{%
\paragraph{Functional-alignment artifact unavailable.}
The frozen result is omitted unless its generated fragment and audited figure
are present in the paper workspace.
}

\FloatBarrier
\section{Design Intervention Summary}
\label{app:design_interventions}

\begin{table}[h]
\caption{Registered design interventions on \pss{}. Effects are absolute
\allrecall{} differences except for the answer-F1 row.}
\label{tab:design_ablation}
\centering
\scriptsize
\setlength{\tabcolsep}{3.5pt}
\begin{tabularx}{\linewidth}{>{\raggedright\arraybackslash}p{0.20\linewidth}>{\raggedright\arraybackslash}p{0.28\linewidth}r>{\raggedright\arraybackslash}X}
\toprule
Design question & Treatment / control & Effect & Interpretation \\
\midrule
Use full ordered history? & Full-history / query-only writer & +0.1188 &
History contributes beyond the active request ($n=32$) $\checkmark$ \\
Session-specific state? & \ourstag{PSS clue} / matched wrong & +0.1750 &
Session identity matters $\checkmark$ \\
Learned parameter structure? & \ourstag{PSS clue} / shuffled & +0.1211 &
Not explained by equal-sized parameter loading $\checkmark$ \\
Guard clue use? & \finaltag{Guarded PSS} / \ourstag{PSS clue} & +0.0055 &
Best non-oracle score; \pss{} null/OOD activation is 0.78\% \\
Fuse unconditionally? & \ourstag{PSS clue} / \ourstag{Raw + PSS RRF} & +0.1297 &
PSS clue ranking is stronger than unconditional fusion \\
Keep state for answering? & \ourstag{Keep state} / \ourstag{unload state} & +0.0037 F1 &
Retrieval-first deployment uses the unload path \\
\bottomrule
\end{tabularx}
\end{table}

\subsection{Cross-backbone hyperparameter sensitivity}
\label{app:hparam_sensitivity}

\begin{figure}[H]
  \centering
  \includegraphics[width=\linewidth]{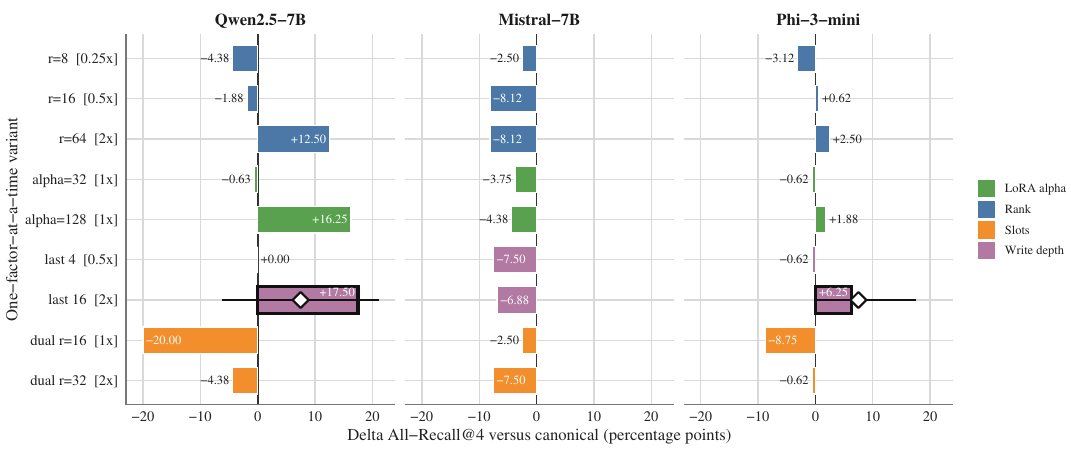}
  \caption{Three-backbone, one-factor-at-a-time sensitivity around the locked
  $r{=}32$, $\alpha{=}64$, last-8, single-slot configuration, with a disjoint
  confirmation set for the screen-selected cell.}
  \label{fig:pss_hparam_waterfalls}
\end{figure}

\paragraph{Writer and state design.}
Calibration first selected the retrieval-aligned objective, then preregistered
a 32-session one-factor-at-a-time screen over rank, LoRA scale, write depth,
and slot topology. All other factors remain fixed to full-history BF16, and the
canonical cell is $r{=}32$, $\alpha{=}64$, last 8 layers, and one slot.
\Cref{fig:pss_hparam_waterfalls} reports percentage-point changes in
\allrecall{} rather than cumulative bridge totals, because effects from
different cells cannot be added.

\paragraph{Backbone-specific parameter effects.}
Qwen is the most capacity-sensitive: relative to its 55.00\% canonical score,
$r{=}64$, $\alpha{=}128$, and last-16 writing gain 12.50, 16.25, and
17.50 points on the screen. The last-16 winner, however, yields only
$+7.50$ points on the disjoint 16-session confirmation set
(95\% CI $[-6.25,21.25]$, Holm $p{=}0.410$). Mistral behaves differently: its
79.38\% canonical cell wins, and every tested change is lower; $r{=}16$,
$r{=}64$, and last-16 each lose 6.88--8.12 points. Phi is less sensitive to
rank and scale ($+0.62$ for $r{=}16$, $+2.50$ for $r{=}64$, and $+1.88$ for
$\alpha{=}128$), while last-16 wins the screen by 6.25 points and confirms at
$+7.50$ points (95\% CI $[0.00,17.50]$, Holm $p{=}0.258$). These profiles show
that PSS supports backbone-specific tuning while the locked canonical cell
remains a robust shared configuration across all three models.

\paragraph{Capacity, placement, and topology.}
The cross-model pattern is non-monotone. Doubling rank helps Qwen and only
slightly helps Phi, but harms Mistral; increasing $\alpha$ is a zero-storage
change that helps Qwen strongly, helps Phi modestly, and harms Mistral.
Writing the last 16 layers doubles state size and is selected for Qwen and Phi,
but not for Mistral. A capacity-controlled second slot never improves the
canonical cell: dual-$r{=}16$ loses 20.00, 2.50, and 8.75 points for Qwen,
Mistral, and Phi, respectively. The result strongly supports the locked
single-slot topology, whose canonical state sizes are 44, 40, and 24 MiB per
session. Additional semantic-peak analysis appears in
\Cref{app:semantic_peak}.

\section{Supervision and Carrier Confirmation}
\label{app:carrier_confirmation}

This preregistered follow-up tests whether ordered history contributes beyond
the canonical target supplied to the writer and whether parameters offer
utility beyond persistent text produced offline from the same information. It
uses the newly generated PSS-Carrier-320 confirmation cohort (320 sessions; 40
per category), created after and disjoint from the original PSS-320 benchmark
and all development and smoke-test sessions. It includes all answerable probes, BGE-M3
\citep{multi2024m3} and BM25, and writer seeds 91--93. All histories fit
the registered 16K window; the audit passes every split, seed,
compute-matching, and coverage check.

The matched state receives $(H_s,z_s^+,z_s^-)$; the target-only state omits
$H_s$. The strongest text baseline uses the same history, supervision, offline
model family, 256-token budget, and matched per-session generation compute.
Neither method replays history online. Direct retrieval is primary; a guard
fitted outside the confirmation sessions supplies the secondary view.

\begin{table}[H]
\caption{Information-by-carrier confirmation (session-weighted \allrecall{},
percent). Retriever cells report direct/guarded retrieval. ``Null act.'' is
frozen-guard null/OOD activation.}
\label{tab:carrier_all_arms}
\centering
\scriptsize
\setlength{\tabcolsep}{3.2pt}
\renewcommand{\arraystretch}{0.92}
\begin{tabularx}{\linewidth}{@{}Xrrr@{}}
\toprule
Arm & BGE-M3 & BM25 & Null act. \\
\midrule
Raw query & 12.50/12.50 & 17.32/17.32 & 0.00 \\
Direct target & 17.50/16.43 & 41.96/34.11 & 79.46 \\
Direct target pair & 17.32/17.86 & 38.39/30.89 & 33.93 \\
Text from history, 32 tokens & 11.79/13.75 & 31.25/26.07 & 83.48 \\
Text from history, 128 tokens & 13.93/14.64 & 39.46/29.46 & 85.71 \\
Text from history, 256 tokens & 15.36/15.54 & 42.14/30.89 & 85.71 \\
Text from $(H,z)$, 32 tokens & 11.07/13.04 & 34.46/25.89 & 78.57 \\
Text from $(H,z)$, 128 tokens & 16.43/16.96 & 38.21/29.29 & 80.80 \\
Text from $(H,z)$, 256 tokens & 21.25/19.46 & 53.04/38.57 & 83.93 \\
Compute-matched text from $(H,z)$ & 21.07/20.00 & 53.39/38.39 & 82.59 \\
PSS from history only & 4.64/9.64 & 16.07/19.46 & 85.71 \\
PSS from target only & 21.79/21.61 & 32.14/27.32 & 42.41 \\
PSS from $(H,z)$, correct & \textbf{27.32}/21.07 & 42.32/31.07 & 17.86 \\
PSS from $(H,z)$, wrong & 0.36/11.25 & 3.21/14.82 & 17.41 \\
PSS from $(H,z)$, shuffled & 3.27/9.40 & 15.18/16.85 & 97.17 \\
\bottomrule
\end{tabularx}
\end{table}

\begin{table}[H]
\caption{Preregistered direct-retrieval effects in percentage points. Holm
correction is within the registered family; ``all seeds'' requires a common
direction across the three writer seeds.}
\label{tab:carrier_primary}
\centering
\scriptsize
\setlength{\tabcolsep}{3.0pt}
\renewcommand{\arraystretch}{0.92}
\begin{tabularx}{\linewidth}{@{}lXrrrc@{}}
\toprule
Retriever & Comparison & Effect & 95\% CI & Holm $p$ & All seeds \\
\midrule
BGE-M3 & history increment & +5.54 & [$+0.36,+10.71$] & 0.0428 & yes \\
BGE-M3 & PSS $-$ matched text & +6.25 & [$+1.25,+11.43$] & 0.0346 & yes \\
BGE-M3 & correct $-$ wrong & +26.96 & [$+22.32,+31.79$] & $<10^{-4}$ & yes \\
BGE-M3 & correct $-$ shuffled & +24.05 & [$+19.17,+28.81$] & $<10^{-4}$ & yes \\
BM25 & history increment & +10.18 & [$+3.93,+16.61$] & 0.0051 & yes \\
BM25 & PSS $-$ matched text & $-11.07$ & [$-18.04,-4.11$] & 0.9990 & no \\
BM25 & correct $-$ wrong & +39.11 & [$+33.39,+44.82$] & $<10^{-4}$ & yes \\
BM25 & correct $-$ shuffled & +27.14 & [$+21.55,+32.80$] & $<10^{-4}$ & yes \\
\bottomrule
\end{tabularx}
\end{table}

History increment is positive after correction for both retrievers and every
seed, showing that $H_s$ contributes beyond $(z_s^+,z_s^-)$. The carrier
effect is representation-dependent: PSS beats matched text with BGE-M3, while
BM25 favors lexically richer text. Large correct--wrong and
correct--shuffled margins rule out arbitrary-adapter and generic-rewrite
explanations. The frozen guard exceeds its 5\% null/OOD threshold (17.86\% for
correct PSS), so its transfer gate fails. We do not tune on the confirmation
set; direct retrieval remains the registered primary analysis.

\section{Limitations and Threats to Validity}
\label{app:limitations}

This section consolidates the scope qualifications for the performance claims
in the main paper. In particular, it records the distinction between retrieval
and answer decoding, router-transfer behavior, UltraDomain ceiling effects,
cost relative to textual state, backend implementation status, benchmark
construction, and privacy or continual-learning questions. These limitations
do not change the registered retrieval improvements; they define the settings
to which those improvements can currently be generalized.

\subsection{Experimental status}

The selected writer configuration, controls, and decision rules are fixed
across writer seeds 91--93. All reported \pss{} values and diagnostics
aggregate these seeds while retaining sessions as the independent units for
paired inference, including when seed runs were executed on separate machines.
The Mistral, Phi-3, UltraDomain, and 112-session strong-backend protocols were
frozen before their respective reported comparisons, providing separate
portability and backend confirmations. The carrier confirmation in
\Cref{app:carrier_confirmation} uses the newly generated, disjoint
PSS-Carrier-320 cohort and the same writer seeds 91--93 to test history increment and a
text baseline under the reported wall-clock budget and a separate preregistration.

\subsection{Synthetic benchmark and sampled human audit}

\pss{} is generated from controlled templates with model-assisted text
realization and programmatic evidence alignment. This supports causal
interventions and exact evidence groups, but also introduces repeated
linguistic patterns, artificial identifiers, and a narrow technical-project
domain. The completed 32-session stratified review compares sampled histories,
scope-resolving rewrites, supporting evidence, null probes, and the separation
between writer targets and formal answers. It provides a human check on the
benchmark's construction and labels. 

The public extension reduces dependence on that generator . Histories contain released source
questions plus an explicit request-ordinal control turn. En.QA and NarrativeQA sessions each index one shared
document and therefore saturate target-document recall for nearly every arm.

UltraDomain adds domain diversity but not natural dialogue. Its sessionized
raw arm already reaches 94.64--96.43\%, and none of the three models passes
the preregistered joint support threshold. Correct PSS states nevertheless
raise All-Recall@3 over raw by 1.79--4.46 points for every backbone; the ``no''
support flags record the threshold decision, not an absence of retrieval
benefit. The original released QA task is
useful for long-context and retrieval baselines but does not evaluate PSS because
it contains no session history.

\subsection{Router transfer across domains}

The 32-session public router chooses only raw or parsed-reference actions and never the
PSS-RRF action on the 262-session sealed test. Although it improves over raw
retrieval, it reduces recall relative to PSS clue retrieval
and activates on 11.83\% of null and 25.57\% of OOD probes. The corresponding
confirmation rates, 3.13\% and 34.38\%, also show unstable OOD behavior. The
0.78\% conditional activation reported on \pss{} therefore does not provide
a portable safety guarantee. The independently frozen guard in the carrier
confirmation likewise misses its registered 5\% transfer gate: correct-PSS
null/OOD activation is 17.86\%, and routing reduces both BGE-M3 and BM25
retrieval relative to the direct primary analysis. We do not retune either
threshold on its test set. A deployable guard may benefit from larger, more
diverse negative data and a separately frozen cross-domain evaluation.

\subsection{Separation of retrieval and answer generation}

The primary endpoint is evidence retrieval, not answer correctness. All
answer arms have zero Exact Match, and even gold-query and capped-full-context
controls reach only about 0.35 token F1. These controls isolate the demonstrated
contribution to evidence access; reader optimization is outside the retrieval
intervention evaluated here.

\subsection{State capacity and scaling}

The state has fixed shape for one session, but the experiment does not prove
that it preserves all information in the session. The writer is explicitly
optimized for session fields and retrieval clues, and retention is expected to degrade
when useful information exceeds its finite capacity. Write compute grows
linearly with history chunks. Storage grows linearly in the number of
sessions, and current query, retrieved text, and generated-token KV still
consume online memory. Hence ``zero history tokens'' is not ``zero tokens'',
``zero memory'', or an asymptotic constant-memory system over arbitrarily many
sessions.

\subsection{Cost relative to text state}

The primary Qwen rank-32 state occupies 44.04 MiB and takes 29.03 seconds to
write; audited logical sizes are 44, 40, and 24 MiB for Qwen, Mistral, and
Phi-3. The sealed parsed request averages 65.65 raw UTF-8 bytes. Text cards
therefore optimize storage, editability, and write latency, while PSS targets
retrieval utility through a reusable non-text carrier. The public parsed reference provides
the stronger retrieval/F1 trade-off, and Equation~\ref{eq:state_choice} makes
positive held-out utility, reuse.

 With identical history and target information, the same offline
model family, a 256-token text budget, and matched per-session generation
compute, PSS gains 6.25 points under BGE-M3, while BM25 gains 11.07 points from
lexically explicit text. 

\subsection{Retriever and domain interaction}

BM25 is unusually strong because \pss{} contains project IDs, version strings,
configuration names, and checkpoint identifiers. Dense BGE-small performs
poorly on the generated clues, whereas BGE-M3 improves substantially when
paired with the same state rewrite. The graph-compatible result also transfers, but
its performance depends on the entity-extraction, graph-construction,
community-summarization, and retrieval configurations, as well as their
indexing cost. Likewise, the 112-session HyDE- and RQ-RAG-compatible rows
freeze their query-transformation logic around the common retriever contract;
they are not official end-to-end reproductions of those systems. These are
model/data interactions, not a general ranking of lexical, dense, hierarchical,
and graph retrieval. Natural paraphrases, multilingual sessions, and
semantically equivalent descriptions could change the result.

\subsection{Capped context and efficiency scope}

The primary \pss{} full-context input is capped at 32,768 tokens. Nominal 65K
histories are therefore truncated and the arm is not a true full-history
oracle. Original UltraDomain uses different declared context regimes: Qwen is
YaRN-compatible, Mistral is native-32K truncated, and Phi-3 is native-128K;
their Full rows are consequently not a controlled backbone ranking. Efficiency
uses 16 sessions and one measured generation length; it does not report
run-to-run standard deviation, energy, validated FLOPs, or all length-specific
curves. The 67.64\% latency and 11.60\% peak-allocation reductions are
descriptive for the measured hardware/software state, not universal system
constants.

\subsection{Qualitative and capacity-analysis scope}

Only 42 of 240 semantic-peak combinations yield identifiable breakpoints. The
observed breakpoints are therefore reported descriptively and do not identify a
capacity law or a globally optimal rank.

\subsection{Privacy, deletion, and continual learning}

Per-session adapters can retain sensitive information. Correct/wrong/shuffled
controls do not test membership inference, source extraction, adapter theft,
unlearning, or deletion. The registry experiment does not demonstrate
continual learning, non-interference across sequential updates, or rollback.
Deployment considerations include access control, encryption, lifecycle
management, and privacy testing.

\subsection{Threats by validity class}

\paragraph{Internal validity.}
The active query and  clue are available to some writer objectives,
so query-only and target-only controls are important. The three-seed carrier
confirmation shows that adding history improves over the target-only state for
both registered retrievers, but its history-only state remains below raw. This
supports retrieval-supervised history integration. Matched-wrong states are balanced by category and
difficulty but do not match every lexical property. The
sequential calibration can create researcher degrees of freedom. Aggregating
the locked configuration over seeds 91--93, together with the separately
frozen cross-model and strong-backend studies, reduces but does not eliminate
this concern.

\paragraph{Construct validity.}
Rewrite F1 is not retrieval success; retrieval success is not reader success;
state bytes are not source compression ratio; and a fixed adapter shape is not
evidence of losslessness. Likewise, clue-to-target progress is close to the
retrieval-clue training objective: it validates functional output alignment but
does not reveal relational geometry in parameter space. We therefore prioritize
all-support evidence recall, report answer controls separately, and avoid
generic ``context compression'' or intrinsic-geometry claims.

\paragraph{External validity.}
Seven public QA tasks, UltraDomain, two additional retrievers, and three
backbones broaden the original synthetic-domain result. They establish
method-level portability across the tested adapter layouts, not direct reuse of
one state across models. One language, three tested writer seeds per
state-writing evaluation, three related
decoder-only model families.

\paragraph{Statistical conclusion validity.}
H1/H2 have paired intervals and corrected tests, but H3 control gaps lack a
complete per-control interval table in the current artifact summary. All
reported state-writing analyses aggregate seeds 91--93; category, semantic-peak, and
efficiency analyses remain exploratory where indicated. Public
and cross-retriever sign-flip tests use session-level pairs and Holm correction,
but their minimum reported Monte Carlo $p$-value is
$1/(100{,}000+1)$; it is not an exact value below that resolution.
Cross-model tables preserve the preregistered Holm pass/fail decisions; they do
not infer daggers from rounded aggregate means. Original UltraDomain has no
joint paired all-method family; sessionized UltraDomain yields positive gains
for every model without crossing the joint support threshold.

\section{Public Sessionized Extension}
\label{app:public}

\subsection{Per-task retrieval and answering}

\begin{table}[t]
\caption{Sealed public target-document All-Recall@3 (percent). ``PSS-RRF''
fuses raw and PSS-clue rankings. The final row is session-weighted.
``Parsed ref.'' denotes the protocol-resolved active-request reference:
the complete request designated by the history's activity marker and
used directly for retrieval.}
\label{tab:public_tasks}
\centering
\scriptsize
\resizebox{\textwidth}{!}{%
\begin{tabular}{lrrrrrrrrrr}
\toprule
Task & $n$ & Raw & Full & Zero & Wrong & Shuffled & PSS-RRF & PSS clue & Router diag. & Parsed ref. \\
\midrule
2WikiMQA & 40 & 30.00 & 62.50 & 52.50 & 42.50 & 22.50 & 95.00 & 100.00 & 72.50 & 100.00 \\
HotpotQA & 40 & 45.00 & 57.50 & 27.50 & 47.50 & 15.00 & 97.50 & 100.00 & 82.50 & 100.00 \\
$\infty$Bench En.QA & 56 & 100.00 & 100.00 & 100.00 & 100.00 & 100.00 & 100.00 & 100.00 & 100.00 & 100.00 \\
MultiFieldQA-en & 30 & 53.33 & 63.33 & 60.00 & 66.67 & 36.67 & 93.33 & 100.00 & 73.33 & 100.00 \\
MuSiQue & 40 & 45.00 & 80.00 & 50.00 & 45.00 & 20.00 & 97.50 & 100.00 & 97.50 & 100.00 \\
NarrativeQA & 16 & 100.00 & 100.00 & 100.00 & 100.00 & 100.00 & 100.00 & 100.00 & 100.00 & 100.00 \\
Qasper & 40 & 57.50 & 70.00 & 67.50 & 62.50 & 52.50 & 82.50 & 77.50 & 77.50 & 80.00 \\
\midrule
Session-weighted & 262 & 60.69 & 75.95 & 64.50 & 65.27 & 48.47 & 95.04 & 96.56 & 86.26 & 96.95 \\
\bottomrule
\end{tabular}}
\vspace{-2em}
\end{table}

\begin{table}[h]
\caption{Public answer token F1 (percent) after unloading the state.}
\label{tab:public_answer}
\centering
\scriptsize
\resizebox{\textwidth}{!}{%
\begin{tabular}{lrrrrrrrr}
\toprule
Method & 2Wiki & Hotpot & En.QA & MultiField & MuSiQue & Narrative & Qasper & Weighted \\
\midrule
Raw & 0.12 & 0.46 & 0.75 & 2.16 & 0.11 & 2.65 & 2.30 & 1.03 \\
Full-history rewrite & 5.85 & 2.12 & 1.12 & 10.13 & 1.44 & 5.91 & 2.39 & 3.56 \\
PSS clue retrieval & 31.74 & 24.53 & 8.58 & 43.17 & 10.37 & 11.92 & 15.22 & 20.00 \\
Router diagnostic & 30.02 & 21.25 & 2.36 & 19.76 & 10.61 & 8.65 & 15.00 & 15.03 \\
Parsed ref. & 37.33 & 25.98 & 6.73 & 42.48 & 10.06 & 12.55 & 22.96 & 21.78 \\
\bottomrule
\end{tabular}}
\end{table}

The answer results are auxiliary because the primary intervention targets
retrieval. Unlike the earlier \pss{} reader diagnostic, public Exact Match is
not universally zero; nevertheless, the generated aggregate currently reports
only F1 in the paper table. PSS-clue F1 is 20.00\%, close to the parsed
reference's 21.78\% and well above the ambiguous query's 1.03\%, under the strict
answer isolation boundary.

\subsection{Conversational-rewrite retrieval-to-QA extension}
\label{app:cqr_qa}

The follow-up registration was written before its formal run, and the completed
artifact passed the independent matrix audit. Its primary
track reuses the locked 262-session public sealed split; sessionized
UltraDomain provides a separate 112-session replication. Both tracks forbid
answers, documents, and target IDs from every rewriter. Raw, parsed-reference, and
PSS rewrites are joined with an
official Apache-2.0 ConvSearch-R1 Qwen2.5-3B checkpoint trained on QReCC
. We retain the
official history/rewrite prompt and temperature 0.7; this is explicitly
cross-domain inference, not retraining on our sessionization. One generated
rewrite is shared by all three readers. IterCQR has an official repository and
Google Drive model release, but the released QReCC is a 66,322,432-byte truncated PyTorch zip with no
central-directory record; the listed full-size TopiOCQA asset has not yet
passed integrity and reuse-authorization checks. CHIQ releases code but no
checkpoint or observed license. AdaRewriter's README displays an MIT badge,
but the inspected revision has no license file and no complete released
candidate-generator/reward checkpoint. Exact rows for these methods therefore
require complete author-released predictions or a complete checkpoint with an
explicit authorization record; unavailable methods receive no invented score.

For every method, BGE-M3 revision retrieves the same 512-token,
64-token-overlap evidence interface at top-3. History is kept hidden and PSS
is unloaded. For the primary QA
endpoint, the answer model receives one method-invariant explicit source
question plus method-retrieved chunks, while the method rewrite is hidden;
therefore answer differences isolate the downstream value of retrieved
evidence. We report rewrite-plus-evidence deployment QA secondarily. Qwen2.5-7B,
Mistral-7B, and Phi-3-mini decode greedily in BF16. The two primary endpoints
are target-document All-Recall@3 and answer F1 on both public sessionizations.
We first average within
\texttt{base\_session\_id}, then use paired two-sided sign-flip tests, paired
10,000-sample session bootstraps, and one Holm family per track across reader,
method, and endpoint, including every method--raw contrast and every
PSS--baseline contrast. We do not combine the two track-level $p$-values.
We declare a reliable retrieval-to-QA bridge against raw only when
both retrieval and F1 effects against raw are positive after correction. A
paper-table dagger is stricter and denotes positive endpoints against every
declared baseline in that track's family. Cross-dataset statements are reserved
for cases in which both tracks pass independently. The
retrieval-rescued subset is descriptive and is not called a mediation test.
The tables below retain positive mean changes that miss the adjusted threshold,
but label them as descriptive rather than significant; negative results are
reported without suppression.

\IfFileExists{cqr_appendix/generated_cqr_qa_results.tex}{%
  \paragraph{Outcome summary.}

On the primary public sessionized track, PSS passes the registered retrieval-to-QA criterion against the ambiguous raw query for all three readers: All-Recall@3 rises by 35.50--35.88 points and controlled answer F1 rises by 4.06--10.60 points (all $p_{\mathrm{Holm}}=3.0\times10^{-4}$). PSS also significantly exceeds ConvSearch-R1 in retrieval by 14.89--15.27 points for every backbone, while its F1 advantage is significant only for Qwen2.5-7B.

Sessionized UltraDomain exhibits a retrieval ceiling: PSS improves the mean All-Recall@3 contrast against raw by 1.79 points, but the adjusted retrieval test is nonsignificant. Controlled answer F1 nevertheless improves by 10.63--18.69 points for all readers ($p_{\mathrm{Holm}}=3.0\times10^{-4}$). We report these positive means and significant F1 effects, but do not relabel the track as passing the registered joint criterion.

\begin{figure*}[t]

\centering

\includegraphics[width=\textwidth]{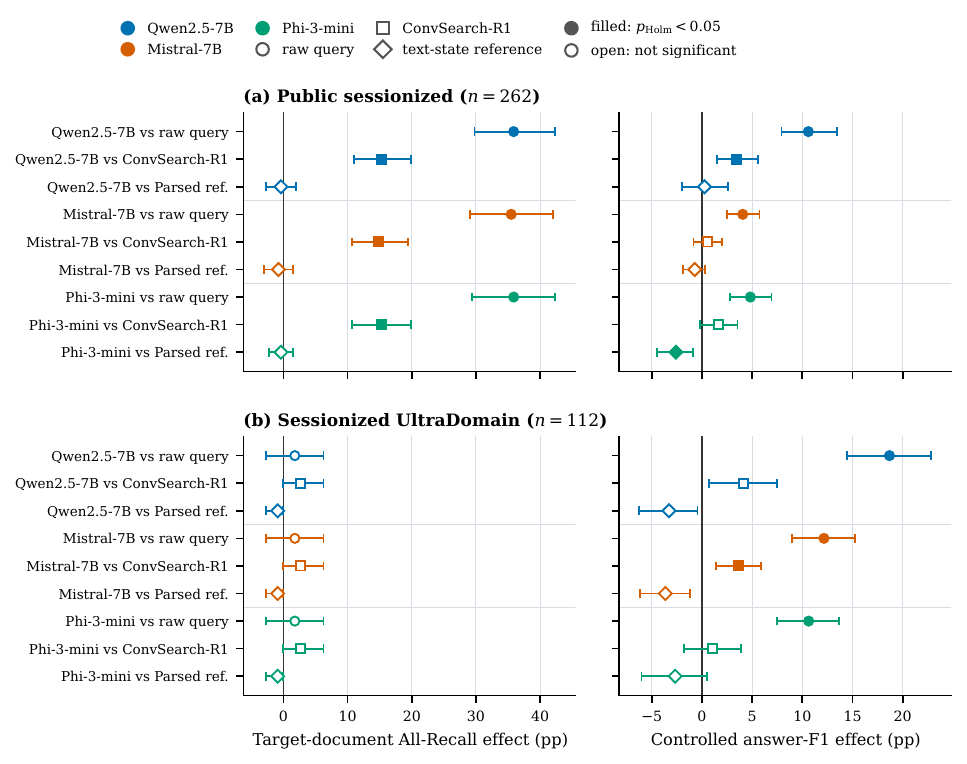}

\caption{Session-level paired retrieval and controlled-answer effects on the public and UltraDomain sessionizations; whiskers are paired 95\% bootstrap CIs.}

\label{fig:cqr_qa_effects}

\end{figure*}

\begin{table*}[t]
\centering
\scriptsize
\setlength{\tabcolsep}{4.2pt}
\begin{tabular}{llrrrrr}
\toprule
Reader & Method & R@3 & Ctrl F1 & Ctrl EM & Deploy F1 & Deploy EM \\
\midrule
Qwen2.5-7B & Ambiguous raw query & 60.69 & 4.62 & 0.38 & 1.31 & 0.00 \\
Qwen2.5-7B & Parsed ref. & 96.95 & 14.96 & 2.67 & 15.25 & 2.67 \\
Qwen2.5-7B & ConvSearch-R1 & 81.30 & 11.74 & 2.67 & 6.80 & 0.38 \\
Qwen2.5-7B & PSS clue retrieval & 96.56 & 15.22 & 3.05 & 15.00 & 2.29 \\
\midrule
Mistral-7B & Ambiguous raw query & 60.69 & 6.54 & 0.00 & 2.34 & 0.00 \\
Mistral-7B & Parsed ref. & 96.95 & 11.32 & 0.00 & 11.24 & 0.00 \\
Mistral-7B & ConvSearch-R1 & 81.30 & 10.02 & 0.00 & 6.35 & 0.00 \\
Mistral-7B & PSS clue retrieval & 96.18 & 10.60 & 0.00 & 11.07 & 0.00 \\
\midrule
Phi-3-mini & Ambiguous raw query & 60.69 & 8.61 & 0.00 & 2.38 & 0.00 \\
Phi-3-mini & Parsed ref. & 96.95 & 16.03 & 1.53 & 16.08 & 1.53 \\
Phi-3-mini & ConvSearch-R1 & 81.30 & 11.81 & 0.76 & 6.60 & 0.38 \\
Phi-3-mini & PSS clue retrieval & 96.56 & 13.44 & 1.15 & 12.98 & 0.76 \\
\bottomrule
\end{tabular}
\caption{Complete Public sessionized retrieval-to-QA results (percent). Ctrl uses the method-invariant source question and method-retrieved evidence; Deploy exposes each method rewrite. Grok-based recParsed ref. denotes the complete active-request text selected by the
protocol's explicit activity marker and used directly for retrieval.ord selection}
\label{tab:cqr-qa-public-sessionized}
\end{table*}

\begin{table*}[t]
\centering
\scriptsize
\setlength{\tabcolsep}{3.5pt}
\begin{tabular}{llrrrrc}
\toprule
Reader & PSS vs. & $\Delta$R@3 [95\% CI] & $p_{\mathrm{Holm}}$ & $\Delta$Ctrl F1 [95\% CI] & $p_{\mathrm{Holm}}$ & Joint$^+$ \\
\midrule
Qwen2.5-7B & raw query & +35.88 [+29.77, +42.37] & 3.00e-04 & +10.60 [+7.93, +13.42] & 3.00e-04 & yes \\
Qwen2.5-7B & ConvSearch-R1 & +15.27 [+11.07, +19.85] & 3.00e-04 & +3.48 [+1.51, +5.57] & 0.006 & yes \\
Qwen2.5-7B & Parsed ref. & -0.38 [-2.67, +1.91] & 1.000 & +0.26 [-1.98, +2.56] & 1.000 & no \\
\midrule
Mistral-7B & raw query & +35.50 [+29.01, +41.98] & 3.00e-04 & +4.06 [+2.47, +5.74] & 3.00e-04 & yes \\
Mistral-7B & ConvSearch-R1 & +14.89 [+10.69, +19.47] & 3.00e-04 & +0.58 [-0.83, +2.00] & 1.000 & no \\
Mistral-7B & Parsed ref. & -0.76 [-3.05, +1.53] & 1.000 & -0.71 [-1.85, +0.35] & 1.000 & no \\
\midrule
Phi-3-mini & raw query & +35.88 [+29.39, +42.37] & 3.00e-04 & +4.83 [+2.77, +6.93] & 3.00e-04 & yes \\
Phi-3-mini & ConvSearch-R1 & +15.27 [+10.69, +19.85] & 3.00e-04 & +1.62 [-0.17, +3.54] & 0.619 & no \\
Phi-3-mini & Parsed ref. & -0.38 [-2.29, +1.53] & 1.000 & -2.59 [-4.48, -0.89] & 0.029 & no \\
\bottomrule
\end{tabular}
\caption{Complete Public sessionized session-level PSS contrasts in percentage points, with paired 10,000-sample bootstrap CIs and Holm-adjusted tests. Joint$^+$ requires both effects to be positive with adjusted $p<0.05$.}
\label{tab:cqr-effects-public-sessionized}
\end{table*}

\begin{table*}[t]
\centering
\scriptsize
\setlength{\tabcolsep}{4.2pt}
\begin{tabular}{llrrrrr}
\toprule
Reader & Method & R@3 & Ctrl F1 & Ctrl EM & Deploy F1 & Deploy EM \\
\midrule
Qwen2.5-7B & Ambiguous raw query & 96.43 & 21.70 & 0.00 & 8.59 & 0.00 \\
Qwen2.5-7B & Parsed ref. & 99.11 & 43.66 & 0.00 & 44.55 & 0.00 \\
Qwen2.5-7B & ConvSearch-R1 & 95.54 & 36.27 & 0.00 & 32.46 & 0.00 \\
Qwen2.5-7B & PSS clue retrieval & 98.21 & 40.38 & 0.00 & 39.50 & 0.00 \\
\midrule
Mistral-7B & Ambiguous raw query & 96.43 & 27.20 & 0.00 & 13.74 & 0.00 \\
Mistral-7B & Parsed ref. & 99.11 & 43.00 & 0.00 & 42.72 & 0.00 \\
Mistral-7B & ConvSearch-R1 & 95.54 & 35.72 & 0.00 & 33.35 & 0.00 \\
Mistral-7B & PSS clue retrieval & 98.21 & 39.36 & 0.00 & 39.35 & 0.00 \\
\midrule
Phi-3-mini & Ambiguous raw query & 96.43 & 30.28 & 0.00 & 14.86 & 0.00 \\
Phi-3-mini & Parsed ref. & 99.11 & 43.58 & 0.00 & 43.94 & 0.00 \\
Phi-3-mini & ConvSearch-R1 & 95.54 & 39.84 & 0.00 & 32.68 & 0.00 \\
Phi-3-mini & PSS clue retrieval & 98.21 & 40.92 & 0.00 & 38.59 & 0.00 \\
\bottomrule
\end{tabular}
\caption{Complete Sessionized UltraDomain retrieval-to-QA results (percent). Ctrl uses the method-invariant source question and method-retrieved evidence; Deploy exposes each method rewrite. Grok-based recParsed ref. denotes the complete active-request text selected by the
protocol's explicit activity marker and used directly for retrieval.ord selection}
\label{tab:cqr-qa-ultradomain-sessionized}
\end{table*}

\begin{table*}[t]
\centering
\scriptsize
\setlength{\tabcolsep}{3.5pt}
\begin{tabular}{llrrrrc}
\toprule
Reader & PSS vs. & $\Delta$R@3 [95\% CI] & $p_{\mathrm{Holm}}$ & $\Delta$Ctrl F1 [95\% CI] & $p_{\mathrm{Holm}}$ & Joint$^+$ \\
\midrule
Qwen2.5-7B & raw query & +1.79 [-2.68, +6.25] & 1.000 & +18.68 [+14.41, +22.81] & 3.00e-04 & no \\
Qwen2.5-7B & ConvSearch-R1 & +2.68 [+0.00, +6.25] & 1.000 & +4.11 [+0.72, +7.45] & 0.335 & no \\
Qwen2.5-7B & Parsed ref. & -0.89 [-2.68, +0.00] & 1.000 & -3.28 [-6.24, -0.44] & 0.519 & no \\
\midrule
Mistral-7B & raw query & +1.79 [-2.68, +6.25] & 1.000 & +12.16 [+9.00, +15.26] & 3.00e-04 & no \\
Mistral-7B & ConvSearch-R1 & +2.68 [+0.00, +6.25] & 1.000 & +3.64 [+1.37, +5.90] & 0.044 & no \\
Mistral-7B & Parsed ref. & -0.89 [-2.68, +0.00] & 1.000 & -3.64 [-6.18, -1.19] & 0.095 & no \\
\midrule
Phi-3-mini & raw query & +1.79 [-2.68, +6.25] & 1.000 & +10.63 [+7.51, +13.65] & 3.00e-04 & no \\
Phi-3-mini & ConvSearch-R1 & +2.68 [+0.00, +6.25] & 1.000 & +1.08 [-1.76, +3.91] & 1.000 & no \\
Phi-3-mini & Parsed ref. & -0.89 [-2.68, +0.00] & 1.000 & -2.67 [-6.01, +0.51] & 1.000 & no \\
\bottomrule
\end{tabular}
\caption{Complete Sessionized UltraDomain session-level PSS contrasts in percentage points, with paired 10,000-sample bootstrap CIs and Holm-adjusted tests. Joint$^+$ requires both effects to be positive with adjusted $p<0.05$.}
\label{tab:cqr-effects-ultradomain-sessionized}
\end{table*}

\paragraph{Interpretation boundary.}

These results separate two claims. First, PSS reliably restores omitted session intent on the primary public sessionization. Second, positive mean changes below the registered threshold remain descriptive and are retained in the figure and complete tables; they are not promoted to significant retrieval-to-QA support. Accordingly, no dagger is assigned in this extension and no cross-track combined $p$-value is constructed.
}{}

\subsection{Paired effects, controls, and audit}

\begin{table}[h]
\caption{Public sealed-test paired All-Recall@3 effects. Tests are two-sided
session-level sign flips with a single Holm family of six comparisons.}
\label{tab:public_stats}
\centering
\small
\begin{tabular}{lrrr}
\toprule
Comparison & Difference & $p$ & $p_{\mathrm{Holm}}$ \\
\midrule
\ourstag{PSS clue} $-$ raw & +0.3588 & $9.9999{\times}10^{-6}$ & $5.99994{\times}10^{-5}$ \\
\ourstag{PSS clue} $-$ wrong & +0.3130 & $9.9999{\times}10^{-6}$ & $5.99994{\times}10^{-5}$ \\
\ourstag{PSS clue} $-$ shuffled & +0.4809 & $9.9999{\times}10^{-6}$ & $5.99994{\times}10^{-5}$ \\
\diagtag{Router diagnostic} $-$ raw & +0.2557 & $9.9999{\times}10^{-6}$ & $5.99994{\times}10^{-5}$ \\
\diagtag{Router diagnostic} $-$ Parsed ref. & -0.1069 & $1.99998{\times}10^{-5}$ & $5.99994{\times}10^{-5}$ \\
\diagtag{Router diagnostic} $-$ full rewrite & +0.1031 & 0.001610 & 0.001610 \\
\bottomrule
\end{tabular}
\end{table}

The reporting script assigns no dagger to the public method because its rule
is satisfied only when the public router diagnostic significantly surpasses
raw, the parsed reference, and full history; the reference comparison has a negative effect. For the PSS-320 retriever
interaction, the four state-versus-raw/card comparisons each use 256 paired
sessions and have $p_{\mathrm{Holm}}=3.99996\times10^{-5}$, so both state rows
in \Cref{tab:cross_retrievers} receive a dagger.

\begin{table}[h]
\caption{Public null/OOD false activation (percent). Unguarded output is
non-abstaining for every probe.}
\label{tab:public_negative}
\centering
\small
\begin{tabular}{llrr}
\toprule
Split & Probe & \ourscell{PSS clue} & \diagcell{Router diag.} \\
\midrule
Confirmation ($n=32$) & Null & \ourscell{100.00} & \diagcell{3.13} \\
Confirmation ($n=32$) & OOD & \ourscell{100.00} & \diagcell{34.38} \\
Sealed test ($n=262$) & Null & \ourscell{100.00} & \diagcell{11.83} \\
Sealed test ($n=262$) & OOD & \ourscell{100.00} & \diagcell{25.57} \\
\bottomrule
\end{tabular}
\end{table}

The final aggregate contains 4,192 rows for 262/262 expected sessions. The
state audit covers three write shards, 652 states, zero failures, and no shard
outside the registered failure tolerance. On sealed answerable sessions the
router chooses the parsed reference 158 times and raw 104 times, with zero PSS-RRF choices.
These counts explain why the guard improves abstention while sacrificing much
of the correct-state retrieval gain.

\section{Cross-Model and UltraDomain Details}
\label{app:cross_model}

\begin{table}[h]
\caption{Full cross-model control matrix (percent target-document
All-Recall@3). Public and UltraDomain are separate sessionized protocols.
``Support'' applies the preregistered positive, Holm-corrected comparisons of
the PSS clue produced by the correct state against raw, wrong, and shuffled
controls. The router diagnostic includes the parsed-reference action.}
\label{tab:cross_model_controls}
\centering
\scriptsize
\resizebox{\textwidth}{!}{%
\begin{tabular}{llrrrrrrrc}
\toprule
Model & Track & $n$ & Raw & Wrong & Shuffled & \ourscell{PSS clue} & \diagcell{Router diag.} & Parsed ref. & Support \\
\midrule
Qwen2.5-7B & Public & 262 & 60.69 & 65.27 & 48.47 & \ourscell{96.56} & \diagcell{86.26} & 96.95 & yes \\
Mistral-7B & Public & 262 & 61.83 & 62.21 & 47.71 & \ourscell{95.80} & \diagcell{83.21} & 98.09 & yes \\
Phi-3-mini & Public & 262 & 60.69 & 63.36 & 48.09 & \ourscell{95.80} & \diagcell{85.11} & 97.33 & yes \\
\midrule
Qwen2.5-7B & Ultra sessionized & 112 & 96.43 & 90.18 & 89.29 & \ourscell{98.21} & \diagcell{99.11} & 99.11 & no \\
Mistral-7B & Ultra sessionized & 112 & 94.64 & 91.96 & 87.50 & \ourscell{98.21} & \diagcell{98.21} & 99.11 & no \\
Phi-3-mini & Ultra sessionized & 112 & 94.64 & 91.07 & 88.39 & \ourscell{99.11} & \diagcell{98.21} & 99.11 & no \\
\bottomrule
\end{tabular}}
\end{table}

The public correct--wrong gaps are 31.30, 33.59, and 32.44 points, and the
correct--shuffled gaps are 48.09, 48.09, and 47.71 points for Qwen, Mistral,
and Phi-3. These matched controls rule out a generic benefit from loading any
adapter. UltraDomain's direction is also consistent, but the high raw and parsed-reference
ceilings prevent the complete preregistered decision from passing.

\begin{table}[h]
\caption{Original released UltraDomain QA (normalized token F1, percent).
PSS is N/A because no session history is provided. Qwen is YaRN-compatible,
Mistral is explicitly 32K-truncated, and Phi-3 is native 128K. Formatting is
script-generated; no dagger is emitted because there is no joint paired
all-method test artifact.}
\label{tab:original_ultradomain}
\centering
\scriptsize
\resizebox{\textwidth}{!}{%
\begin{tabular}{lllrrrr}
\toprule
Model & Family & Method & Fin & Legal & Misc (Mix) & Ave. \\
\midrule
Qwen2.5-7B & Long context & YaRN-compatible & \underline{46.86} & \best{50.16} & 47.66 & \underline{48.23} \\
Qwen2.5-7B & Retrieval & BGE-M3 & 44.96 & 45.35 & \underline{47.87} & 46.06 \\
Mistral-7B & Long context & Native-32K truncated & 42.26 & 38.97 & 44.54 & 41.92 \\
Mistral-7B & Retrieval & BGE-M3 & 45.87 & 44.04 & 46.23 & 45.38 \\
Phi-3-mini & Long context & Native-128K & \best{51.17} & \underline{46.52} & \best{48.09} & \best{48.59} \\
Phi-3-mini & Retrieval & BGE-M3 & 45.77 & 45.47 & 47.43 & 46.22 \\
\bottomrule
\end{tabular}}
\end{table}

Phi-3 native-128K obtains the best average (48.59), narrowly ahead of Qwen's
YaRN-compatible 48.23. BGE-M3 beats Mistral's truncated long-context path by
3.46 points, whereas long context beats BGE-M3 by 2.17 points for Qwen and
2.37 points for Phi-3. These are descriptive comparisons across unequal
context budgets; original UltraDomain evaluates long-context QA, not PSS.

\begin{table}[h]
\caption{Audited model-state contracts. Logical state size is measured from
the actual BF16 state manifest; every model has zero write failures. Context
labels describe the original UltraDomain long-context baseline, not the PSS
writer's 128-token chunks.}
\label{tab:model_state_contracts}
\centering
\scriptsize
\resizebox{\textwidth}{!}{%
\begin{tabular}{llllrr}
\toprule
Model & Revision & Long-context label & LoRA targets & Layers & MiB/session \\
\midrule
Qwen2.5-7B & \texttt{main} & YaRN-compatible 128K & q,k,v,o,gate,up,down & final 8 & 44 \\
Mistral-7B & \texttt{b3af163} & native 32K, truncated & q,k,v,o,gate,up,down & final 8 & 40 \\
Phi-3-mini & \texttt{f3c06aed} & native 128K & qkv,o,gate-up,down & final 8 & 24 \\
\bottomrule
\end{tabular}}
\end{table}

The smallest state, Phi-3's 24 MiB snapshot, obtains the strongest \pss{}
scores in both added retrievers. This is an efficiency observation, not a
causal architecture claim: hidden size, fused projections, backbone quality,
and state parameter count change together. The experiment establishes
method-level portability, not cross-model interchangeability of state files.
 \section{\pss{} Construction and Audit}
\label{app:data}

\subsection{Generation contract}

Each session describes one synthetic technical project. The generator first
samples a semantic category and a balanced combination of load factors. It
then instantiates:
\begin{enumerate}
    \item an active project, topic, component, configuration, and current
    decision;
    \item archived same-type alternatives and, where applicable, superseded
    decisions;
    \item one to eight source documents containing the controlled evidence;
    \item an ordered session history containing the evidence plus routine
    distractor turns;
    \item a 32-token structured card, a 128-token summary, and 
    writer fields;
    \item six evaluation probes, gold rewrites, answer strings, and one or
    more required evidence groups.
\end{enumerate}

The history and source documents use independent surface realizations of the
same controlled facts. Formal probe answers are serialized only in the
evaluation records and are prohibited from writer input. The source is
released under CC0-1.0.

\subsection{Splits and balance}

\begin{table}[h]
\caption{\pss{} split and calibration roles.}
\label{tab:pss_splits}
\centering
\small
\begin{tabular}{lrrl}
\toprule
Partition & Sessions & Probes & Role \\
\midrule
Calibration--selection & 32 & 192 & writer/rank/topology selection \\
Calibration--guard fit & 16 & 96 & feature scaling and router fit \\
Calibration--confirmation & 16 & 96 & threshold/risk confirmation \\
Held-out test & 256 & 1536 & aggregate over writer seeds 91--93 \\
\midrule
Total & 320 & 1920 & \\
\bottomrule
\end{tabular}
\end{table}

Every semantic category contains eight calibration and 32 test sessions. Each
of the four levels of every load axis appears in 16 calibration and 64 test
sessions. Source-group IDs are disjoint across calibration and test. 
Both are retained to distinguish the source records from the prepared
experiment representation.

\subsection{Probe definitions}

\begin{table}[h]
\caption{Six probes instantiated for every session.}
\label{tab:probe_types}
\centering
\small
\begin{tabularx}{\linewidth}{lX}
\toprule
Probe & Intended capability \\
\midrule
Canonical readout & Recover an explicitly recorded active request. \\
Unseen paraphrase & Resolve a new surface form of an existing intent. \\
Anaphoric query & Resolve references such as ``that configuration''. \\
New composition & Retrieve all groups in a relation chain not stated as one sentence. \\
Exact fact & Locate versions, dates, thresholds, and checkpoints. \\
Null/conflict & Reject out-of-scope or superseded requests. \\
\bottomrule
\end{tabularx}
\end{table}

The first three probes diagnose session-scope recovery. New composition and
exact fact stress evidence access rather than free-form parametric recall.
Null/conflict probes have empty evidence groups and a gold rewrite of
UNSUPPORTED.

\subsection{Evidence and leakage audit}

For answerable probes, each evidence group contains one or more text spans that
jointly support one component of the answer. \allrecall{} equals one only when
at least one chunk from every group is present in the top four. Automated
checks enforce: 320 unique sessions; 1,920 probes; balanced category/axis
counts; source-group split disjointness; unique formal answer strings; and no
formal answer use by the writer. All automated issues are zero.

The human audit samples one calibration and three test sessions from each
category. Reviewers compare each sampled record with its history, intended
scope, evidence groups, null status, and writer target to check history
coherence, scope resolution, evidence sufficiency, null validity, and
formal-answer exclusion. The completed 32-session comparison provides sampled
human validation of these construction criteria.
 \section{Writer Training and State Construction}
\label{app:training}

\subsection{Backbone and adapter placement}

The backbone is Qwen2.5-7B-Instruct in BF16. LoRA is inserted into
\texttt{q\_proj}, \texttt{k\_proj}, \texttt{v\_proj},
\texttt{o\_proj}, \texttt{gate\_proj}, \texttt{up\_proj}, and
\texttt{down\_proj} in the final eight transformer layers. The selected state
has rank 32 and $\alpha/r=2$. Only the state parameters are optimized; the
backbone remains frozen.

\subsection{Language-model phase}

The selected history coverage is \texttt{full\_ordered\_history}. Tokenized
history is divided into 128-token chunks. The number of LM updates is
$\max(16,N_{\mathrm{chunks}})$, ensuring every chunk is visited at least once.
Optimization uses AdamW, learning rate $2\times10^{-5}$, no weight decay, and
gradient clipping at 1.0. Activations are bounded by the chunk size, but total
write time increases with the number of chunks.

\subsection{Retrieval-aligned readout}

After the LM phase, the selected \texttt{retrieval\_aligned} implementation
uses 64 updates at learning rate $10^{-4}$: the sum of the configured 32
readout and 32 retrieval steps. Every update in this phase uses the loss in
\Cref{eq:retloss}, with a newly initialized phase optimizer. Its positive
target is a compact  clue containing
the project, current configuration/component, checkpoint, and route terms.
The hard negative uses a same-session archived configuration or unrestricted
answer path. The margin in \Cref{eq:retloss} is 0.2. Training excludes the
evaluation probe answer.

\subsection{Writer objective screen}

\begin{table}[h]
\caption{Writer objectives in the registered screen.}
\label{tab:writer_objectives}
\centering
\small
\begin{tabularx}{\linewidth}{lX}
\toprule
Objective & Training signal \\
\midrule
LM-only & Ordered-history next-token modeling. \\
Query-only & Canonical active-query readout; no history LM. \\
LM + query & History LM plus active-query readout. \\
Multi-field & Topic, entities, latest decision, and query readout. \\
Retrieval-aligned & History LM plus positive clue over hard-negative margin. \\
Multi-field + retrieval & Multi-field targets plus retrieval alignment. \\
\bottomrule
\end{tabularx}
\end{table}

The final lock chose \texttt{retrieval\_aligned}; this differs from the base
configuration file's original rank-4 \texttt{multi\_field\_plus\_retrieval}
default. The run artifact, not the unmodified default, is the source of truth
for the reported method.

\subsection{History and content ablations}

The history-coverage matrix contains active query only, last 128 tokens, first
25/50/75\%, full ordered history, full history with shuffled chunk order, and
structured card only. Content deletion separately removes active query, active
topic, entities, key evidence, latest decision, or superseded decision. These
interventions distinguish query caching, bag-of-facts retention, order
sensitivity, and version handling.

\subsection{Capacity, topology, and precision}

The registered ranks are 2, 4, 8, 16, and 32 with fixed $\alpha/r$.
Topologies are a single slot and a dual slot intended to separate stable
document state from writable session state. Precision arms are BF16 full
snapshot, BF16 base plus FP16 residual, INT8 residual, and INT4 residual.
The final lock is rank 32, single slot, BF16 full snapshot. The semantic-peak
stage does not establish rank 32 as globally optimal
(\Cref{app:semantic_peak}).

\subsection{Causal states}

For each test session we construct four states:
\begin{itemize}
    \item \textbf{correct:} written from that session;
    \item \textbf{matched wrong:} written from another session matched by
    category and difficulty;
    \item \textbf{shuffled:} the correct state's parameter values are
    permuted within the registered control transformation;
    \item \textbf{zero reset:} the initial no-write state.
\end{itemize}
All controls use the same backbone, prompts, retriever, and scoring code.
 \section{Router Features, Cross-Fitting, and Fallback}
\label{app:router}

\subsection{Feature set}

The first 11 features compare raw and PSS-clue paths:
\begin{enumerate}
    \item explicit state refusal;
    \item raw/rewrite Jaccard similarity;
    \item raw-query containment in the rewrite;
    \item rewrite novelty;
    \item raw/PSS top-4 chunk overlap;
    \item raw/PSS top-1 agreement;
    \item normalized PSS top-1 versus top-2 margin;
    \item normalized PSS versus raw top-score gain;
    \item raw-query overlap with raw evidence;
    \item raw-query overlap with state-retrieved evidence;
    \item PSS-clue overlap with PSS-retrieved evidence.
\end{enumerate}
The three-way router adds the analogous refusal, similarity, novelty, ranking,
score, and evidence-overlap features for the card path, plus state/card top-4
overlap and top-1 agreement. All text overlap is computed over normalized
lexical tokens. \texttt{UNSUPPORTED}, \texttt{UNANSWERABLE},
\texttt{NO ANSWER}, and \texttt{INSUFFICIENT EVIDENCE} are explicit refusals.

\subsection{Model fitting}

The router contains one balanced logistic scope model and one logistic reward
model per action (\texttt{raw}, \texttt{card}, and the implementation key
\texttt{state\_rrf}, reported as PSS-RRF).
Feature means and scales are learned on calibration data only. Four-fold
memory-level cross-fitting ensures that a session does not both fit a model and
produce its own out-of-fold threshold prediction. The scope threshold and
action margins are fixed before test evaluation. A low scope score selects the
raw action. Otherwise, the highest adjusted action score wins, with raw
preferred on exact ties.

\subsection{Information boundary}

Permitted features use only the current raw query, generated rewrites,
retrieved chunk IDs/scores, and text overlap with returned chunks. The router
is not given access to the session ID, probe ID, semantic category, state-control
label, gold evidence, formal answer, or test outcome. The state writer and
router calibration both complete before held-out evaluation.

\subsection{Why three actions are retained}

The raw path protects literal or out-of-scope questions. The card provides a
cheap and interpretable global-theme summary. PSS-RRF can recover fine-grained
project/configuration terms absent from both. The final test routes 518 probes
to raw, 472 to card, and 546 to PSS-RRF. These counts show non-degenerate use
of all three paths, but action-conditioned accuracy is selected and therefore
not a causal estimate of action quality.

\subsection{RRF implementation}

For each candidate chunk, the implementation sums reciprocal ranks from the
raw and PSS-clue lists with constant 60, sorts by decreasing fused score and
stable chunk ID, and keeps the first four. No learned reranker is used.
Raw + PSS RRF is an unconditional fusion ablation; the formal PSS-RRF action
is selected by the Guarded PSS router.
 \section{Registered Questions, Gates, and Deviations}
\label{app:protocol}

\begin{table}[h]
\caption{Registered hypotheses and observed decisions.}
\label{tab:hypotheses}
\centering
\small
\begin{tabularx}{\linewidth}{lXl}
\toprule
Hypothesis & Gate & Result \\
\midrule
H1 raw superiority & $\Delta\allrecall{}\geq0.10$ and CI lower $>0$ &
Pass: $+0.3328$ \\
H2 card non-inferiority & CI lower $\geq-0.02$ &
Pass; actually $+0.1117$ \\
H3 causal binding & correct--wrong and correct--shuffled $\geq0.10$ &
Pass: $+0.1750,+0.1211$ \\
H4 full-history semantics & full history--query only $\geq0.05$ &
Mechanism pass: $+0.1188$ \\
H5 answer combination & correct exceeds unload, wrong, shuffled by 0.05 &
Fail; unload gap $+0.0037$ \\
Null/OOD & conditional false activation $\leq0.05$ &
Pass: 0.0078 \\
Failure & maximum arm failure rate $\leq0.02$ &
Pass: 0 \\
\bottomrule
\end{tabularx}
\end{table}

\subsection{Registered research questions}

The protocol asks whether the state supports zero-history retrieval, whether
it contains more than an active-query cache, whether it improves complete
evidence retrieval, whether keeping it active improves answering, where
semantic capacity fails, and when its cost is preferable to a text card.
H1--H5 map these questions to explicit effect-size gates.

\subsection{Sequential calibration design}

The original full factorial would require six objectives, five ranks, two
topologies, and three seeds. The revised design first screens six objectives at
ranks 4/8 with a single slot, then screens rank and topology only for the
selected objective. This estimates the principal objective effect in the
pilot-supported rank range and a selected-objective capacity/topology curve;
it does not estimate every objective-by-rank-by-topology interaction.

\subsection{Evaluation status and claim scope}

The selected configuration is evaluated with writer seeds 91, 92, and 93 on
H200 hardware. The same frozen configuration, controls, and decision rules are
used for every seed, and seed-level evidence is summarized before reporting. Thus,
passing H1--H4 and the safety gates reflects
replication across initializations. Claim scope remains limited by synthetic
session construction, mechanical public sessionization, and the reader and
router limitations reported elsewhere.

The automated report contains an internal
\texttt{context\_compression\_claim\_allowed=true} flag because H4 passes its
mechanism threshold. The top-level report correctly leaves
\texttt{parameterized\_context\_compression\_supported=false}. We follow the
top-level interpretation in this paper.
 \section{Complete Arm Definitions and Aggregate Retrieval Results}
\label{app:arms}

\begin{table}[h]
\caption{Thirteen main arms. History tokens denote the configured online
history-derived text budget, including cards and summaries. Guarded PSS
includes its card branch. ``Unload'' means that the LoRA state is removed
before answer generation.}
\label{tab:arms}
\centering
\scriptsize
\begin{tabularx}{\linewidth}{l l X}
\toprule
Arm & History tokens & Retrieval and answer path \\
\midrule
Raw & 0 & Raw ambiguous query; base 7B + RAG. \\
Recent-128 & 128 & Recent-window rewrite; base 7B + RAG. \\
Card-32 & 32 & Structured-card rewrite; base 7B + RAG. \\
Summary-128 & 128 & Text-summary rewrite; base 7B + RAG. \\
Full session & full & Full-history rewrite; base 7B + RAG. \\
PSS clue, unload & 0 & PSS clue retrieval; unload; base 7B + RAG. \\
Guarded PSS, unload & 32 (card) & Raw/card/PSS-RRF routing; unload; base 7B + RAG. \\
PSS clue, keep & 0 & PSS clue retrieval; keep selected answer state + RAG. \\
Guarded PSS, keep & 32 (card) & Guarded query/evidence; keep answer state + RAG. \\
PSS clue + card & 32 & PSS and card hybrid rewrite; unload + RAG. \\
Gold query & 0 & Canonical query; base 7B + RAG. \\
Gold evidence & 0 & Gold evidence groups; base 7B answer. \\
Full context & full & Capped history, no retrieval; base 7B answer. \\
\bottomrule
\end{tabularx}
\end{table}

\begin{table}[h]
\caption{Aggregate retrieval diagnostics. History tokens denote the
configured online history-derived text budget; Guarded PSS includes its
card branch. Actual prompt-token and full-path cost measurements are
reported separately in \Cref{app:followup_cost}.}
\label{tab:retrieval_aggregate}
\centering
\small
\begin{tabular}{lrrr}
\toprule
Path & History tokens & \allrecall{} & Difference from raw \\
\midrule
Raw query & 0 & 0.1734 & -- \\
Recent-128 & 128 & 0.1359 & -0.0375 \\
Card-32 & 32 & 0.3945 & +0.2211 \\
Summary-128 & 128 & 0.4023 & +0.2289 \\
Full-session rewrite & full & 0.3578 & +0.1844 \\
PSS clue retrieval & 0 & 0.5008 & +0.3273 \\
Raw + PSS RRF & 0 & 0.3711 & +0.1977 \\
Guarded PSS & 32 (card) & 0.5063 & +0.3328 \\
Gold query & 0 & 0.8281 & +0.6547 \\
\bottomrule
\end{tabular}
\end{table}

\subsection{State controls}

For the guarded arm, correct, matched-wrong, and shuffled states obtain
0.5063, 0.3313, and 0.3852 \allrecall{}. The equal-sized four-control summary
implies an approximate zero-reset value of 0.3867; we treat this absolute
number as provisional pending regeneration from per-control rows. The
registered causal decisions depend on the directly reported correct--wrong and
correct--shuffled gaps, not the reconstructed zero value.
 \section{Semantic Category Analysis}
\label{app:categories}

\begin{table}[h]
\caption{BM25 retrieval diagnostic by semantic category. This table uses PSS
clue retrieval with the correct state to isolate clue quality; it is not the
Guarded PSS router's category-conditioned performance.}
\label{tab:category_results}
\centering
\scriptsize
\begin{tabular}{lrrrrr}
\toprule
Category & Raw & Card & \ourscell{PSS clue} & \ourscell{PSS$-$Raw} & \ourscell{PSS$-$Card} \\
\midrule
Active topic & 0.2188 & 0.3875 & \ourscell{0.4813} & \ourscell{+0.2625} & \ourscell{+0.0938} \\
Coreference & 0.1500 & 0.3625 & \ourscell{0.5375} & \ourscell{+0.3875} & \ourscell{+0.1750} \\
Entity binding & 0.1563 & 0.3625 & \ourscell{0.5000} & \ourscell{+0.3438} & \ourscell{+0.1375} \\
Exact fact & 0.1563 & 0.3625 & \ourscell{0.5250} & \ourscell{+0.3688} & \ourscell{+0.1625} \\
Global theme & 0.2125 & 0.5063 & \ourscell{0.4563} & \ourscell{+0.2438} & \ourscell{-0.0500} \\
Multi-hop & 0.1500 & 0.3438 & \ourscell{0.4875} & \ourscell{+0.3375} & \ourscell{+0.1438} \\
Null/OOD diagnostic & 0.1938 & 0.4625 & \ourscell{0.5313} & \ourscell{+0.3375} & \ourscell{+0.0688} \\
Temporal revision & 0.1500 & 0.3688 & \ourscell{0.4875} & \ourscell{+0.3375} & \ourscell{+0.1188} \\
\bottomrule
\end{tabular}
\end{table}

PSS clue retrieval's strongest card-relative gains occur in coreference, exact fact, multi-hop,
and entity binding. These categories benefit from recovering a project name,
component, configuration, version, or relation-chain endpoint before
retrieval. The global-theme card is better by five points, consistent with a
structured text summary being sufficient for broad topic recovery.

The null/OOD row in \Cref{tab:category_results} is a retrieval diagnostic, not
a safety result. A null query has no gold evidence groups, and an unguarded
rewrite can still retrieve lexically related chunks. Safety is instead measured
by the router's conditional false-activation rate.

\begin{table}[h]
\caption{Guarded PSS action counts by category over 192 probes per
category. Counts describe behavior, not action-specific causal quality.}
\label{tab:router_category}
\centering
\small
\begin{tabular}{lrrr}
\toprule
Category & Raw & Card & \finalcell{PSS-RRF} \\
\midrule
Active topic & 70 & 56 & \finalcell{66} \\
Coreference & 58 & 59 & \finalcell{75} \\
Entity binding & 63 & 54 & \finalcell{75} \\
Exact fact & 78 & 53 & \finalcell{61} \\
Global theme & 58 & 73 & \finalcell{61} \\
Multi-hop & 66 & 59 & \finalcell{67} \\
Null/OOD & 67 & 64 & \finalcell{61} \\
Temporal revision & 58 & 54 & \finalcell{80} \\
\midrule
Total & 518 & 472 & \finalcell{546} \\
\bottomrule
\end{tabular}
\end{table}

The highest card count occurs for global theme, matching the card's retrieval
strength. PSS-RRF is selected most often for temporal revision and is also
frequent for coreference and entity binding. However, the router was trained
to allocate actions based on observable confidence. Comparing outcome means
between selected action groups would be confounded by sample difficulty.
 \section{Retriever Interaction}
\label{app:retrievers}

\begin{table}[h]
\caption{\allrecall{} under four retrievers and five query paths. Each LoRA
value uses the same PSS clue generated by the correct state.}
\label{tab:retrievers}
\centering
\small
\begin{tabular}{lrrrrr}
\toprule
Retriever & Raw & Card & \ourscell{PSS clue} & \ourscell{Raw + PSS RRF} & Gold query \\
\midrule
BM25 flat & 0.1734 & 0.3945 & \ourscell{0.5008} & \ourscell{0.3711} & 0.8281 \\
Dense BGE-small & 0.0891 & 0.0539 & \ourscell{0.2320} & \ourscell{0.2164} & 0.5438 \\
Hierarchical summary-leaf & 0.1516 & 0.2445 & \ourscell{0.4938} & \ourscell{0.3648} & 0.7648 \\
Entity-relation graph & 0.1594 & 0.3281 & \ourscell{0.4797} & \ourscell{0.3570} & 0.8297 \\
\bottomrule
\end{tabular}
\end{table}

The entity-relation graph has the strongest gold-query score, while BM25 has
the strongest PSS-clue score. This inversion suggests that the graph can
exploit a fully specified relation query, whereas the learned clue most
reliably emits exact lexical identifiers. The hierarchical retriever nearly
matches BM25 under PSS clue retrieval and is less dependent on a single flat chunk level.

Dense BGE-small underperforms on every path. The clue and benchmark contain
synthetic IDs such as configuration names and checkpoints that are
discriminative lexically but not necessarily close under that frozen embedding
model. The BGE-M3 extension in \Cref{tab:cross_retrievers} reverses the strong
version of this diagnosis: state All-Recall@4 reaches 0.3391 versus 0.1242 for
raw retrieval. The result is encoder-specific evidence, not a general ranking
of dense retrieval or a tuned upper bound.

Raw + PSS RRF is below PSS clue retrieval for all four retrievers. With $k=4$,
fusing weak raw-query hits can displace one required evidence group. This
motivates the calibrated action selector rather than unconditional hybrid
retrieval.

\paragraph{Strong-backend public confirmation.}
The second panel of \Cref{tab:cross_retrievers} uses a preregistered,
task-stratified 112-session public subset that excludes all seven sessions used
for implementation smoke testing.  Selection uses task and session IDs only;
answers, target documents, and query variants are not consulted.  Each frozen
backend receives the same raw, full-history, parsed-reference, correct-state,
matched-wrong, shuffled-state, RRF, and gold query sources.  This design holds
the index and retriever fixed, so the paired effect measures the value of the
session-memory query interface rather than differences between document
stores.

Correct PSS clues improve All-Recall@3 for every completed backend.  BGE-M3,
Contriever, Qwen3-Embedding-4B, and compatible HyDE/RQ-RAG pipelines pass the
joint Holm-corrected rule against raw, wrong, and shuffled states.  Their
individual gains over raw are 35.71, 30.36, 28.57, 35.71, and 35.71 points,
respectively.  BM25 gains 18.75 points but its correct--wrong comparison does
not survive correction; RAPTOR gains 10.71 points but likewise fails the joint
control rule.  Across all seven rows, PSS differs from the parsed reference by only
$-0.89$ to $+1.79$ points, with no significant comparison.  Thus the
confirmation supports portable recovery of a structured retrieval intent
through parameter states, with the parsed reference providing a text
comparison whose request selection is resolved by the protocol's
explicit activity markers.

The earlier single-session wrong-state reversal disappears at confirmation
scale: correct PSS exceeds wrong state by 30.36--47.32 points for all five
supported backends.  GraphRAG is reported in the cross-model evaluation in
Panel A; HippoRAG-v2 and Cognee produced no audited scores and remain absent
from the 112-session confirmation rather than being assigned zeros.  The
HyDE-compatible and RQ-RAG-compatible rows isolate their query-transformation
ideas over the same BGE-M3 backend and are labeled compatible rather than
official end-to-end reproductions.
 \section{Answer-Stage Diagnostics}
\label{app:answer}

\subsection{Reader ceilings}

\begin{table}[h]
\caption{Aggregate answer token F1. Exact Match is 0 for every listed arm.}
\label{tab:answer_paths}
\centering
\small
\begin{tabular}{lr}
\toprule
Answer path & Token F1 \\
\midrule
Raw query + RAG & 0.1570 \\
Structured card + RAG & 0.0560 \\
Summary-128 + RAG & 0.0975 \\
Full-session rewrite + RAG & 0.0905 \\
Gold canonical query + RAG & 0.3547 \\
Gold evidence + base 7B & 0.2870 \\
Capped full context + base 7B & 0.3531 \\
\bottomrule
\end{tabular}
\end{table}

Gold evidence underperforming gold-query RAG is unexpected. Possible causes
include answer/citation formatting, different evidence rendering, excessive
or insufficient evidence, and reader prompt mismatch. Because all Exact Match
values are zero, retrieval-to-answer transition metrics based on Exact Match
have no positive baseline cases.

\subsection{Same-query, same-evidence state ablation}

\begin{table}[h]
\caption{Answer token-F1 differences after fixing the query and retrieved
chunk sequence.}
\label{tab:answer_state}
\centering
\small
\begin{tabular}{lrrl}
\toprule
Comparison & Difference & 95\% CI & H5 status \\
\midrule
\ourstag{Keep correct PSS state} $-$ unload/base & +0.0037 & [-0.0040, 0.0115] & fail \\
\ourstag{Correct PSS state} $-$ matched wrong & +0.0442 & [0.0369, 0.0514] & $<0.05$ \\
\ourstag{Correct PSS state} $-$ shuffled & +0.0042 & [-0.0033, 0.0121] & fail \\
\bottomrule
\end{tabular}
\end{table}

\begin{table}[h]
\caption{Correct-state answer F1 gain over unloading the state, by category.}
\label{tab:answer_category}
\centering
\small
\begin{tabular}{lrlr}
\toprule
Category & \ourscell{PSS gain} & Category & \ourscell{PSS gain} \\
\midrule
Active topic & \ourscell{-0.0437} & Coreference & \ourscell{+0.0427} \\
Entity binding & \ourscell{+0.0250} & Exact fact & \ourscell{+0.0249} \\
Global theme & \ourscell{-0.0019} & Multi-hop & \ourscell{+0.0114} \\
Null/OOD & \ourscell{-0.0197} & Temporal revision & \ourscell{-0.0091} \\
\bottomrule
\end{tabular}
\end{table}

No category simultaneously exceeds unloaded, wrong, and shuffled controls by
0.05. Correct-versus-wrong is positive because the wrong adapter can corrupt
the decoder, but correct-versus-unloaded and correct-versus-shuffled are near
zero. The state is therefore not a reliable decoder-side memory.

\subsection{Vacuous transition statistic}

The generated report contains \texttt{resilience\_rate=1.0}; however,
\texttt{resilience=0} and there are zero Exact-Match-correct baseline cases.
The value results from the implementation's empty-denominator convention, so
we exclude it from evidence of success. All 1,536 probe pairs are in the
``no recovery'' cell under Exact Match.

\subsection{Reader-analysis improvements}

Before using answer results to assess the broader architecture, a future run
could strengthen the analysis by:
\begin{enumerate}
    \item separate answer text from citations before normalization;
    \item support multiple reference answers and standard token-F1;
    \item report answerable and null probes separately;
    \item inspect gold-evidence rendering and order;
    \item use a prompt that requests the same output schema as the references;
    \item manually audit a stratified answer sample;
    \item retain the same-query, same-evidence ablation after repair.
\end{enumerate}
 \section{Semantic-Peak Capacity Diagnostic}
\label{app:semantic_peak}

The semantic-peak experiment asks where a fixed state ceases to retain at
least 90\% of its low-load retrieval behavior. It crosses eight semantic
categories, five ranks, and six independent load axes:
\begin{equation*}
\begin{split}
&|H| \in \{1\mathrm{K},4\mathrm{K},16\mathrm{K},65\mathrm{K}\},\\
&\text{topics}\in\{1,2,4,8\},\quad
\text{same-type entities}\in\{2,4,8,16\},\\
&\text{hops}\in\{1,2,3,4\},\quad
\text{revisions}\in\{0,1,2,4\},\\
&\text{evidence sections}\in\{1,2,4,8\}.
\end{split}
\end{equation*}
The balanced constrained design yields 240 category--rank--axis
combinations. Only 42 combinations (17.5\%) have an identifiable breakpoint.
The remainder are flat, noisy, or never cross the registered 90\% threshold
within the observed levels.

This result is insufficient to fit a stable capacity law. In particular, it
does not show that increasing rank monotonically shifts every breakpoint or
that rank 32 is saturated. The diagnostic is retained because it prevents a
misleading single aggregate from being presented as evidence of unbounded
retention. A future capacity study needs more sessions per cell, multiple
writer seeds, and a monotone model specified before observing outcomes.
 \section{RQ4: Inference, Storage, and Break-Even Accounting}
\label{app:efficiency}

We distinguish the original 16-token profile below from the complete
64-token clue/retrieval/router profile in \Cref{app:followup_cost}.
The latter includes both LLM calls, all three retrieval calls, and card
input for Guarded PSS, and supplies the eight-session workload block in
\Cref{fig:main_retrieval}(b). The two profiles retain their own measured
write times and amortization estimates.

Relative to the available mixed-length full-history profile, the Qwen PSS clue
path uses 98.97\% fewer online prompt tokens, 67.64\% lower total latency, and
11.60\% lower peak allocation. Its 29.03-second offline write amortizes after
an estimated 43 repeated reads against average full-history prefill, but never
against the 32-token card. Audited logical states occupy 44, 40, and 24 MiB per
session for Qwen, Mistral, and Phi, respectively. These measurements describe
the tested hardware and workload rather than a universal efficiency law.

\subsection{Online profile}

\begin{table}[h]
\caption{Measured online profile on 16 sessions with 16 generated tokens.
``Swap/setup'' is the measured adapter or state-transition overhead recorded
by the harness.}
\label{tab:efficiency_full}
\label{tab:answer_efficiency}
\centering
\small
\begin{tabular}{lrrrr}
\toprule
Method & Prompt & Generation & Swap/setup & Peak alloc. \\
\midrule
Full-history prefill & 5522.00 & 1.0144 s & -- & 17.368 GB \\
PSS clue path & 56.75 & 0.3054 s & 0.0228 s & 15.353 GB \\
Structured card-32 & 80.75 & 0.3079 s & 0.0155 s & 15.362 GB \\
Recent-128 & 96.75 & 0.3329 s & 0.0171 s & 15.368 GB \\
\bottomrule
\end{tabular}
\end{table}

The average full-history row mixes available 1K, 4K, and 16K profiles.
Configured 65K and 128K rows are unavailable because they exceed the model
input cap or source length. The reported state total of 0.3282 seconds is
generation plus swap. The 56.75 prompt tokens include the current query,
instruction, and output template, so the online request is not token-free.

\subsection{State bytes and residency}

\begin{table}[h]
\caption{State construction and multi-session residency.}
\label{tab:state_storage}
\centering
\small
\begin{tabular}{lr}
\toprule
Quantity & Value \\
\midrule
Offline write time per session & 29.033 s \\
Logical state bytes & 46,137,344 \\
Physical state bytes & 46,175,478 (44.04 MiB) \\
Physical bytes for 256 test states & 11,820,922,368 (11.01 GiB) \\
Estimated states at 80\% free-GPU budget & 579 \\
Four-state LRU hit rate & 0.8125 \\
Mean host-to-device latency on miss & 0.01658 s \\
Mean adapter swap latency in LRU profile & 0.00205 s \\
\bottomrule
\end{tabular}
\end{table}

The early rank-8 pilot used roughly 11 MiB per state, but the final rank-32
method uses 44.04 MiB. Only the latter is relevant to the main result.
The 579-state estimate is a capacity calculation under the observed free-memory
snapshot, not a sustained serving benchmark.

\subsection{Amortization}

For method $m$ and $N$ reads, the accounting is
\begin{equation}
    C_m(N) = C^{\mathrm{write}}_m +
             N(C^{\mathrm{retrieve}}_m+C^{\mathrm{read}}_m).
\end{equation}
Under the measured latency components, LoRA crosses average full-history
prefill at 43 reads. It never crosses the text-card path because the card has
no model-optimization write and similar online generation cost. Storage,
index construction, and task value are not collapsed into this latency-only
break-even; we report them separately.

The parsed sealed-test active request averages only 65.65 UTF-8 bytes. The
logical Qwen, Mistral, and Phi states are therefore approximately
$702{,}831\times$, $638{,}937\times$, and $383{,}362\times$ larger than that
raw text, excluding container overhead. On the public extension, PSS-minus-reference
differences are negative for both retrieval ($-0.38/-2.29/-1.53$ points) and
answer F1 ($-1.77/-1.49/-3.96$) for Qwen/Mistral/Phi. The public parsed-reference
and compact-card settings therefore represent different information budgets.
The \pss{} compact-card comparison measures session-scope recovery, while
\Cref{app:followup_qa} separately isolates the value of retrieved evidence
for a fixed reader question.

\subsection{What is and is not bounded}

For one active session, online historical state is bounded by one fixed adapter
plus fixed retrieval/query/generation budgets. Offline write activations are
bounded by the 128-token chunk, but write steps grow with history length.
Across sessions, adapter storage is linear. During generation, KV still grows
with the current prompt and output. These distinctions prevent the efficiency
profile from being described as strict constant total memory.
 \section{Statistical Analysis Details}
\label{app:statistics}

\subsection{Unit of analysis}

Each session contributes six correlated probes. Probe rows are first aggregated
within session; they are not treated as 1,536 independent observations.
Bootstrap resampling is paired across methods and stratified by the eight
semantic categories. The final intervals use 10,000 resamples and a fixed
experiment seed.

\subsection{Binary paired tests}

For H1, 134 of 256 sessions are correct only under the guarded state method,
zero are correct only under raw retrieval, and 134 are discordant. The exact
McNemar $p$-value is $9.18\times10^{-41}$ and the Holm-adjusted value is
$1.84\times10^{-40}$.

For H2, 56 sessions are correct only under guarded state retrieval and seven
only under the card, yielding 63 discordant pairs. The exact and Holm-adjusted
$p$-value is $1.36\times10^{-10}$. H2 is a non-inferiority question, so its
primary decision uses the paired difference interval
$[0.0898,0.1344]$ relative to the $-0.02$ margin. The small $p$-value is
supporting paired binary evidence, not the non-inferiority test itself.

\subsection{Bootstrap results}

\begin{table}[h]
\caption{Session-clustered paired bootstrap intervals.}
\label{tab:bootstrap}
\centering
\small
\begin{tabular}{lrrr}
\toprule
Comparison & $n$ sessions & Difference & 95\% CI \\
\midrule
Guarded $-$ raw \allrecall{} & 256 & +0.3328 & [0.3063, 0.3586] \\
Guarded $-$ card \allrecall{} & 256 & +0.1117 & [0.0898, 0.1344] \\
Full history $-$ query-only & 32 & +0.1188 & [0.0375, 0.2188] \\
Keep correct $-$ unload F1 & 256 & +0.0037 & [-0.0040, 0.0115] \\
Correct $-$ shuffled F1 & 256 & +0.0042 & [-0.0033, 0.0121] \\
Correct $-$ wrong F1 & 256 & +0.0442 & [0.0369, 0.0514] \\
\bottomrule
\end{tabular}
\end{table}

\subsection{Multiplicity and exploratory analyses}

H1 and H2 form the corrected primary family. Category, retriever, semantic
peak, efficiency, and answer-category analyses are exploratory and are
interpreted through effect sizes rather than uncorrected significance claims.
H3 uses registered 0.10 causal-gap thresholds; the current summary does not
include bootstrap intervals for each retrieval control gap, which is a
reporting item for a future comprehensive report.
The public extension uses one six-comparison Holm family
(\Cref{tab:public_stats}); its dagger marks guarded candidates that beat raw,
parsed reference, and full-history rewriting, and no public row meets this rule. The
PSS-320 BGE-M3/GraphRAG extension uses one
four-comparison Holm family and assigns daggers only to the two state rows that
beat both raw and card.

\subsection{Failures and denominators}

All stage audits report zero issues and the maximum arm failure rate is zero.
The main artifact contains 43,008 rows across arms and controls. A
camera-ready result table would benefit from per-metric and per-arm counts for
total, valid, missing, empty-output, OOM, NaN, and timeout cases; aggregate stage
success alone does not prove identical denominators for every auxiliary
metric. The public aggregate contains 4,192 rows with complete 262-session
coverage, and its write audit reports 652/652 states and zero failures.
 \section{Reproducibility Checklist}
\label{app:reproducibility}

\subsection{Completed artifacts}

\pss{} source JSONL, generator, manifest, and automated audit;
source-group-disjoint calibration/test split and sealed-test hash;
deterministic 32/16/16 calibration subsplits;
formal-answer exclusion contract for writer training;
correct, matched-wrong, shuffled, and zero-reset controls;
writer objective, rank, topology, history, content, and precision
    configurations;
four retriever implementations and the three-way router;
pinned LongBench/InfiniteBench adapters,  public
    sessionization, deterministic 32/32/262 split, and null/OOD controls;
BGE-M3 and GraphRAG PSS-320 runs;
 frozen Qwen/Mistral/Phi-3 portability contracts with architecture-
    native LoRA targets and audited 44/40/24 MiB logical states;
pinned original and  sessionized UltraDomain tracks, with
    exact-document-hash grouping and explicit PSS non-applicability on the
    original QA task;
652-state write manifest with zero failures and complete
    three-shard coverage;
 session-clustered bootstrap and exact McNemar implementation;
  session-level sign-flip tests with Holm correction and generated
    best/second-best/dagger formatting;
 a preregistered, independently generated PSS-Carrier-320
    confirmation with 320 sessions and three writer seeds, using target-only,
    history-only, matched persistent-text, wrong-state, and
    shuffled-state controls under BGE-M3 and BM25;
stage, environment, checksum, and run auditors, with no formal
    blocker in the cross-model run;
unit and smoke tests covering adapters, leakage, merge, statistics,
    formatting, resume behavior, and model-specific chat templates.

\subsection{Key run configuration}

\begin{table}[H]
\caption{Configuration used to reproduce the reported state.}
\label{tab:repro_config}
\centering
\small
\begin{tabular}{ll}
\toprule
Field & Value \\
\midrule
Primary backbone & Qwen2.5-7B-Instruct \\
Portability backbones & Mistral-7B-Instruct-v0.2; Phi-3-mini-128k \\
Precision & BF16 \\
Writer objective & \texttt{retrieval\_aligned} \\
Primary PSS-320 writer seeds & 91, 92, 93 \\
Carrier-confirmation writer seeds & 91, 92, 93 \\
Carrier-confirmation cohort & PSS-Carrier-320; 320 independent sessions \\
LoRA rank / alpha & 32 / 64 \\
LoRA layers & final 8 \\
Topology & single slot \\
History coverage & full ordered history \\
LM chunk / minimum steps & 128 / 16 \\
LM / readout learning rate & $2{\times}10^{-5}$ / $10^{-4}$ \\
Readout / retrieval steps & 32 / 32 \\
Retrieval margin & 0.2 \\
RAG chunk / overlap / top-$k$ & 192 / 48 / 4 \\
Public RAG chunk / overlap / top-$k$ & 512 / 64 / 3 \\
Public split seed / counts & 20260812 / 32, 32, 262 \\
Public dense retriever & BGE-M3, pinned revision \\
Portability split/test seed & session-disjoint / 20260812 \\
UltraDomain released counts & Fin 345; Legal 438; Mix 130 \\
UltraDomain sessionized sealed test & 112 sessions \\
RRF constant & 60 \\
Rewrite / answer budget & 64 / 64 new tokens \\
Full-context input cap & 32,768 tokens \\
\bottomrule
\end{tabular}
\end{table}

\subsection{Hardware and scheduling}

The seed-specific runs were distributed across separate machines. The primary
Qwen aggregation used H200 hardware; the state-fingerprint and
functional-alignment diagnostics, public extension, and cross-model runs used
three GPUs reported by their runtime as NVIDIA GeForce RTX 4090 with 49,140 MiB
each. Every reported state-writing analysis aggregates writer seeds 91--93.
Calibration and write stages scheduled up to two short-sequence processes per
GPU; full main generation scheduled one process per GPU. Capacity checks
required 23,000 MiB per calibration/write job and 32,000 MiB per main job.
Completed markers allowed interruption-safe resume. The public extension used
one process on each GPU for state writing and each evaluation split. The
cross-model run used the same three devices; its audit
verified pinned safetensors for Mistral, Phi-3, and BGE-M3, dataset checksums,
architecture-native target coverage, sufficient disk, and zero formal
blockers. The audited environment reports PyTorch 2.5.1+cu124,
Transformers 4.57.6, Datasets 5.0.1, Sentence-Transformers 3.4.1,
\texttt{rouge} 1.0.1, and \texttt{rouge-score} 0.1.2.

\subsection{Frozen RQ2 state-geometry run}

The RQ2 script reads the already-audited states and does not train or alter a
model. Writer outputs are aggregated over seeds 91--93; the separate
preregistered analysis seed is 20260814, with eight Rademacher projections per
LoRA pair, 10,000 label permutations, 15 UMAP neighbors, cosine distance, and
minimum distance 0.1. On the server, after installing
\texttt{umap-learn==0.5.7} from the configured package mirror.
The command refuses to overwrite an existing output directory, embedding/energy CSV files, PDF/PNG figures, and a LaTeX
fragment whose numbers are generated from the JSON result. The fragment enters
the paper workspace after the cross-machine audit confirms all 768 state
instances (256 held-out sessions for each of writer seeds 91--93), complete
category coverage, and non-empty figures.

 \section{Qualitative and Error-Analysis Protocol}
\label{app:qualitative}

The registered protocol specifies a stratified, double-annotated sample of 100
failed sessions with adjudication. Planned labels cover:
\begin{itemize}
    \item session scope, project identification, or entity/configuration binding;
    \item temporal revision or superseded-version conflict;
    \item incomplete multi-hop evidence or retriever miss despite a usable rewrite;
    \item router misroute, evidence neglect, or unsupported/invalid generation.
\end{itemize}
Planned materials include the complete session, raw/card/state rewrites and
rankings, router scores, gold evidence groups, and answer outputs. Reporting
includes raw agreement, Cohen's $\kappa$, and third-annotator adjudication.

No completed manual error labels are present, so we report neither fabricated
counts nor selected traces. Quantitative patterns motivate three audit
hypotheses: cards may favor broad themes; LoRA may help when identifiers are
missing; and reader failures may persist after correct retrieval. These remain
unvalidated qualitative hypotheses.

This failure-taxonomy exercise is separate from the completed 32-session human
audit of \pss{} construction and label quality described in \Cref{app:data}.

The four planned case quadrants are: LoRA correct/card wrong, card
correct/LoRA wrong, both correct, and both wrong. Published examples would
include retrieved chunk IDs and exact source spans to support verification of
the claimed mechanism.

\section{Session-State Readout and Complete Serving Costs}
\label{app:readout_cost_followup}

PSS compiles session information into a reusable parameter state from which
retrieval text can be generated online or materialized for subsequent reuse.
This appendix studies that compilation-and-readout path through
information-matched text controls, readout interfaces, and complete online
cost accounting. Protocols S and R analyze 32 existing sessions, three writer
seeds (91--93), and a shared 256-session candidate corpus. A separate
question-conditioned extension evaluates distinct historical-record banks
on sixteen composite memory groups
(\Cref{app:question_conditioned_records}). These experiments complement the
original benchmark and the independent carrier confirmation.

\subsection{Information-matched text controls}
\label{app:followup_text}

\paragraph{Text-state roles.}
A compact structured card, a parsed active-request reference, and a supplied
retrieval clue represent different text states. The original PSS-320
compact-card comparison tests recovery of session scope under a short
online context: PSS clue retrieval reaches 50.08\% versus 39.45\% for the
card, a 10.63-point gain. Guarded PSS reaches 50.63\%, an 11.18-point gain
over that card.

In the public sessionization protocol, the parsed active-request reference
preserves the complete request designated by the history's explicit
activity marker. This request is the protocol-defined target for
active-request reconstruction and also supplies the positive readout
target during session-specific PSS writing. Direct retrieval with this
resolved request provides a reference for evaluating how much of its
retrieval utility PSS recovers through parameter readout. This reference
characterizes retrieval when request selection is already determined.

\paragraph{Protocol S: text-memory controls.}
The compiler and generated-text baselines receive the same ordered history
and answer-free positive/negative clues. Text memories are generated
independently at 128-, 256-, and 512-token budgets; a separate 256-token
structured format records current identifiers, relations, and superseded
alternatives. Each short budget is generated directly rather than extracted
as a prefix of a longer memory. Direct caches retain the supplied positive
clue, the current query plus that clue, or a labeled positive/negative pair.
The fixed session selection yields 4,800 retrieval rows across BGE-M3 and
BM25. Formal answers and gold rewrites are excluded from memory construction.

\begin{table}[H]
\centering
\footnotesize
\setlength{\tabcolsep}{4pt}
\caption{Session-level evidence All-Recall@4 (percent), with complete method coverage in each protocol. S and R reuse the same 32 sessions and 96 states but execute separate readout runs; their estimates are reported separately. A dash denotes a method outside that protocol, not a zero score.}
\label{tab:followup_recall}
\begin{tabular}{lrrrr}
\toprule
& \multicolumn{2}{c}{Text-memory controls (S)} & \multicolumn{2}{c}{Readout controls (R)} \\
\cmidrule(lr){2-3}\cmidrule(lr){4-5}
Method & BGE-M3 & BM25 & BGE-M3 & BM25 \\
\midrule
Raw query & 11.88 & 16.25 & 11.88 & 16.25 \\
Supplied clue & 61.25 & 85.00 & 61.25 & 85.00 \\
Query + supplied clue & 60.62 & 87.50 & 60.62 & 87.50 \\
Query + labeled clue pair & 57.50 & 86.88 & -- & -- \\
Compact card-32 rewrite & 6.88 & 38.75 & -- & -- \\
Text-128 rewrite & 17.50 & 47.50 & -- & -- \\
Text-256 rewrite & 15.00 & 43.75 & 15.00 & 43.12 \\
Text-512 rewrite & 20.00 & 46.88 & 20.62 & 48.12 \\
Structured-256 rewrite & 16.88 & 48.12 & 17.50 & 47.50 \\
Base-model rewrite & -- & -- & 2.50 & 13.75 \\
Supplied clue-pair rewrite & -- & -- & 15.00 & 46.25 \\
PSS online readout & 23.96 & 42.92 & 23.54 & 42.71 \\
Matched-wrong PSS & 0.21 & 2.08 & 0.21 & 2.08 \\
Query + fixed PSS readout & -- & -- & 64.38 & 80.83 \\
Fixed PSS readout + base rewrite & -- & -- & 7.92 & 30.00 \\
\bottomrule
\end{tabular}
\end{table}

\paragraph{Shared-state readout protocol.}
Protocol R reuses the same 96 serialized PSS states and selected sessions;
it adds readout controls without another training stage. Each state is read
once with the fixed training instruction, ``Return only a compact search
clue for the active session.'' This produces a query-independent text
$c_s$ before any future probe is supplied. The two protocols execute separate
generation passes; Table~\ref{tab:followup_recall} preserves both sets of
estimates. The analysis is a targeted follow-up on previously evaluated
sessions, distinct from the independently generated PSS-Carrier-320 cohort.

\subsection{Online parameter readout and PSS-derived text reuse}
\label{app:followup_readout}

Protocol R evaluates two reuse interfaces originating from the same compiled
state $\Delta_s$. Online parameter readout generates
$G_{\theta+\Delta_s}(q)$ for each request. Fixed readout first produces
$c_s=G_{\theta+\Delta_s}(c_0)$ from the training instruction $c_0$, before
any future probe is supplied, and then reuses $c_s$ as a text cache.
The text interfaces include direct retrieval from $q+c_s$ and base-model
rewriting from $(q,c_s)$, alongside supplied-clue caches. Base-only and
matched-wrong controls test parameter conditioning and session correspondence.
Text-256, Text-512, and Structured-256 remain additional text-memory controls.
The protocol contains 3,200 query rows and 6,400 retrieval rows; each
PSS-derived cache retains the seed of its source state.

\paragraph{State-conditioned rewriting.}
Relative to the base-only rewrite, online PSS improves BGE-M3 and BM25
All-Recall@4 by 21.04 and 28.96 points, respectively, with positive
Holm-corrected effects. On BGE-M3, it also improves over rewriting from the
fixed PSS text readout by 15.62 points ($p_H=0.00035$). These comparisons
identify a useful parameter-conditioning mechanism within the tested
rewriting interface. The corresponding BM25 effect over cached-readout
rewriting is 12.71 points ($p_H=0.0826$).

\paragraph{Materializing a PSS-derived retrieval cache.}
The fixed-readout path reaches 64.38\% All-Recall@4 on BGE-M3 and 80.83\%
on BM25 using $q+c_s$. The corresponding supplied-clue cache $q+z_s^+$
reaches 60.62\% and 87.50\%. Here $c_s$ is generated from the optimized
session state rather than copied directly from the supplied clue. These
results show that a compiled parameter state can produce useful retrieval
text for subsequent reuse. Online parameter readout and materialized text
reuse are two operational interfaces to this compilation path: one conditions
each request on $\Delta_s$, and the other persists its fixed readout $c_s$.

The reported retrieval scores characterize the complete interfaces. Fixed
readout uses $q+c_s$, whereas the online PSS arm retrieves with its generated
string without an additional raw-query concatenation. Table~\ref{tab:followup_effects_R}
reports the paired rewrite comparisons; the fixed-cache versus supplied-cache
means above are descriptive. A materialized PSS cache incurs the original
state compilation and one fixed readout before text-only serving.

\paragraph{Statistical unit and generation budget.}
We average probes within a seed and seeds within a session, then perform
two-sided paired session-level sign-flip tests with bootstrap intervals.
Protocol S has 10 primary and four secondary comparisons; R has 12 primary
and 10 secondary comparisons. Holm correction is applied separately to
each complete family across both retrievers. The intervals in the tables
are unadjusted intervals, and a positive interval endpoint pattern does not
replace the corrected test. All selected answerable probes and all methods
are retained; descriptive category/probe strata do not define additional
confirmatory claims.

Online generation uses the same greedy raw-prompt protocol and 64-token cap
in every generated arm. S's generated memories and online rewrites all use
their full generation budgets. Its estimates therefore characterize this
fixed-budget interface; short natural termination is not assumed. R keeps
that decoding policy and reuses S's independently generated text memories.
Finite-logit checks retain finite empty responses, and empty retrieval
strings retrieve no evidence.

\begin{table}[H]
\centering
\footnotesize
\setlength{\tabcolsep}{2.8pt}
\caption{Protocol S: PSS online readout minus each reference (percentage points). Two-sided paired tests use session means; $p_H$ is Holm-adjusted within each listed family across both retrievers. Intervals are unadjusted paired bootstrap intervals. Missing S-secondary intervals were not included in the source summary.}
\label{tab:followup_effects_S}
\begin{tabular}{lrrrrrr}
\toprule
Reference & \multicolumn{3}{c}{BGE-M3} & \multicolumn{3}{c}{BM25} \\
\cmidrule(lr){2-4}\cmidrule(lr){5-7}
& $\Delta$ & 95\% CI & $p_H$ & $\Delta$ & 95\% CI & $p_H$ \\
\midrule
\multicolumn{7}{l}{\textit{Primary family}} \\
Supplied clue & -37.29 & [-49.38, -25.00] & $<0.001$ & -42.08 & [-53.12, -31.25] & $<0.001$ \\
Query + supplied clue & -36.67 & [-50.00, -23.96] & $<0.001$ & -44.58 & [-55.00, -34.17] & $<0.001$ \\
Text-256 rewrite & +8.96 & [+1.25, +17.08] & 0.240 & -0.83 & [-11.88, +10.42] & 1.000 \\
Text-512 rewrite & +3.96 & [-6.67, +14.79] & 1.000 & -3.96 & [-15.62, +7.92] & 1.000 \\
Structured-256 rewrite & +7.08 & [-0.63, +14.79] & 0.514 & -5.21 & [-19.38, +8.33] & 1.000 \\
\midrule
\multicolumn{7}{l}{\textit{Secondary family}} \\
Raw query & +12.08 & -- & 0.004 & +26.67 & -- & $<0.001$ \\
Matched-wrong PSS & +23.75 & -- & $<0.001$ & +40.83 & -- & $<0.001$ \\
\bottomrule
\end{tabular}
\end{table}

\begin{table}[H]
\centering
\footnotesize
\setlength{\tabcolsep}{2.8pt}
\caption{Protocol R: PSS online readout minus each reference (percentage points). Two-sided paired tests use session means; $p_H$ is Holm-adjusted within each listed family across both retrievers. Intervals are unadjusted paired bootstrap intervals.}
\label{tab:followup_effects_R}
\begin{tabular}{lrrrrrr}
\toprule
Reference & \multicolumn{3}{c}{BGE-M3} & \multicolumn{3}{c}{BM25} \\
\cmidrule(lr){2-4}\cmidrule(lr){5-7}
& $\Delta$ & 95\% CI & $p_H$ & $\Delta$ & 95\% CI & $p_H$ \\
\midrule
\multicolumn{7}{l}{\textit{Primary family}} \\
Supplied clue & -37.71 & [-50.21, -25.42] & $<0.001$ & -42.29 & [-53.12, -31.46] & $<0.001$ \\
Query + supplied clue & -37.08 & [-50.62, -24.38] & $<0.001$ & -44.79 & [-55.62, -33.75] & $<0.001$ \\
Base-model rewrite & +21.04 & [+12.92, +29.38] & $<0.001$ & +28.96 & [+19.38, +38.54] & $<0.001$ \\
Supplied clue-pair rewrite & +8.54 & [-0.83, +18.12] & 0.194 & -3.54 & [-16.46, +9.17] & 0.623 \\
Query + fixed PSS readout & -40.83 & [-53.54, -28.33] & $<0.001$ & -38.12 & [-49.58, -26.88] & $<0.001$ \\
Fixed PSS readout + base rewrite & +15.62 & [+8.96, +22.71] & $<0.001$ & +12.71 & [+2.29, +22.92] & 0.083 \\
\midrule
\multicolumn{7}{l}{\textit{Secondary family}} \\
Raw query & +11.67 & [+4.38, +18.96] & 0.038 & +26.46 & [+17.29, +35.83] & $<0.001$ \\
Text-256 rewrite & +8.54 & [+1.04, +16.67] & 0.297 & -0.42 & [-11.25, +10.42] & 1.000 \\
Text-512 rewrite & +2.92 & [-7.71, +13.75] & 1.000 & -5.42 & [-16.88, +6.46] & 1.000 \\
Structured-256 rewrite & +6.04 & [-1.67, +13.96] & 0.811 & -4.79 & [-19.17, +8.75] & 1.000 \\
Matched-wrong PSS & +23.33 & [+16.04, +30.83] & $<0.001$ & +40.62 & [+31.67, +49.79] & $<0.001$ \\
\bottomrule
\end{tabular}
\end{table}

\subsection{Question-conditioned readout from distinct historical records}
\label{app:question_conditioned_records}

PSS supports selective access to multiple historical records through one
compiled parameter state. Given an incoming question, the state generates
retrieval text without supplying the persistent text bank to the readout
model. We evaluate this capability with distinct-record memory banks,
using mechanism controls to assess state-dependent access and text
selectors to compare alternative access interfaces.

\paragraph{Protocol and information boundary.}
We construct eight development and sixteen evaluation memory groups from
96 source sessions, with four sessions per group and no overlap across
groups or splits. Each source contributes an action, a configuration,
a rationale, and a complete relation path. The $K=4$ bank contains one
record from each source, covering the four record types; $K=16$ adds the
remaining twelve records. Each evaluation group has five alias-conditioned
questions, including a second expression for an action record. Both bank
sizes use the same 80 questions and the complete 256-session test corpus;
development uses the 64-session calibration corpus. Evidence targets are
the historical record spans relevant to each question.

Parameter and text methods receive the same historical records and alias
associations. The records contain factual content, potentially including
answers, making this an extension distinct from the primary answer-free
protocols. Bank construction excludes evaluation questions, answers, and
gold rewrites. Training instructions and evaluation expressions are
separate; evaluation questions retain explicit aliases and record-type cues.

\paragraph{Question-conditioned compilation.}
A bank-only writer trains on four question--record templates per record,
using the positive/negative sequence objective without a preceding
history-LM phase. Four visits per training pair yield 64 updates for $K=4$
and 256 for $K=16$, holding per-record exposure constant. Writer seed 91
produces 16 development and 32 evaluation states. Local PSS and base-model
generation use the native chat instruction, greedy decoding, EOS handling,
and a 256-token output cap. Generated or selected text is concatenated
with the original question before evidence retrieval.

\paragraph{Access interfaces.}
All-record concatenation forms one query from the entire bank; record-wise
RRF retrieves separately for each record and fuses the rankings. BM25 and
BGE selectors choose one or two records from the text bank. Grok-based
record selection makes one API call per question to select a record from
the complete bank and its metadata. LLM bank rewriting uses the base model
to generate a retrieval expression from the question and the bank. PSS
accesses the compiled parameter state without bank-text input. Wrong-group
PSS uses another group's state with the same alias vocabulary. The
separately reported target-record reference uses the evaluator's target
record ID and is excluded from text-policy selection.

\begin{table}[H]
\centering
\footnotesize
\setlength{\tabcolsep}{4pt}
\caption{Mechanism controls for question-conditioned parameter readout:
group-averaged evidence All-Recall@4 (percent). Both bank sizes use the
same 16 evaluation groups and 80 questions. Text-selection and rewriting
comparisons are reported in Table~\ref{tab:distinct_record_text_interfaces}.}
\label{tab:distinct_record_readout}
\begin{tabular}{lrrrr}
\toprule
& \multicolumn{2}{c}{$K=4$} & \multicolumn{2}{c}{$K=16$} \\
\cmidrule(lr){2-3}\cmidrule(lr){4-5}
Method & BGE-M3 $\uparrow$ & BM25 $\uparrow$ & BGE-M3 $\uparrow$ & BM25 $\uparrow$ \\
\midrule
Raw query & 0.00 & 0.00 & 0.00 & 0.00 \\
Base-model rewriting & 0.00 & 1.25 & 0.00 & 1.25 \\
All-record concatenation & 25.00 & 47.50 & 18.75 & 35.00 \\
Record-wise RRF & 53.75 & 57.50 & 16.25 & 40.00 \\
\midrule
\textbf{PSS parameter readout} & 76.25 & 83.75 & 83.75 & 95.00 \\
Wrong-group PSS & 1.25 & 0.00 & 1.25 & 0.00 \\
\bottomrule
\end{tabular}
\end{table}

\paragraph{Selective access through parameter state.}
At $K=16$, PSS reaches 83.75\% All-Recall@4 with BGE-M3 and 95.00\%
with BM25. Compared with all-record concatenation, the gains are 65.00
and 60.00 percentage points. Compared with record-wise RRF, PSS gains
67.50 and 55.00 points while reducing the number of evidence queries
from sixteen to one. Wrong-group PSS reaches 1.25\% and 0.00\%,
supporting the role of group-specific information in parameter readout.
At $K=4$, PSS reaches 76.25\% and 83.75\%, exceeding concatenation by
51.25 and 36.25 points. The two bank sizes jointly vary memory content
and total write updates while keeping exposure per record constant.

\begin{table}[H]
\centering
\footnotesize
\setlength{\tabcolsep}{4pt}
\caption{Text-selection and rewriting interfaces under the same distinct-record
protocol. Values are group-averaged evidence All-Recall@4 (percent);
the PSS row is repeated for comparison. The target-record reference uses
evaluation information and is excluded from text-policy selection.
Selector top-1/top-2 denotes the number of memory records included in
the query; the final evidence cutoff is four throughout.
Table~\ref{tab:distinct_record_selected_text} reports descriptive differences
from development-selected strategies.}
\label{tab:distinct_record_text_interfaces}
\begin{tabular}{lrrrr}
\toprule
& \multicolumn{2}{c}{$K=4$} & \multicolumn{2}{c}{$K=16$} \\
\cmidrule(lr){2-3}\cmidrule(lr){4-5}
Method & BGE-M3 $\uparrow$ & BM25 $\uparrow$ & BGE-M3 $\uparrow$ & BM25 $\uparrow$ \\
\midrule
BM25 selector, top-1 & 72.50 & 97.50 & 77.50 & 88.75 \\
BM25 selector, top-2 & 67.50 & 95.00 & 71.25 & 91.25 \\
BGE selector, top-1 & 72.50 & 97.50 & 70.00 & 91.25 \\
BGE selector, top-2 & 70.00 & 90.00 & 68.75 & 92.50 \\
Grok-based record selection & 77.50 & 97.50 & 77.50 & 97.50 \\
LLM bank rewriting & 56.25 & 83.75 & 56.25 & 92.50 \\
\midrule
\textbf{PSS parameter readout} & 76.25 & 83.75 & 83.75 & 95.00 \\
Target-record reference & 77.50 & 97.50 & 83.75 & 97.50 \\
\bottomrule
\end{tabular}
\end{table}

\paragraph{Comparison with text access.}
At $K=16$, PSS achieves the highest BGE-M3 mean among the evaluated
non-reference methods, exceeding the strongest reported text score of
77.50\% by 6.25 points. With BM25, PSS reaches 95.00\%, exceeding all
four retrieval-based selector configurations and bank rewriting; Grok
selection reaches 97.50\%. These results demonstrate question-conditioned
retrieval through parameter state across both backends, with a BGE-M3
advantage in the sixteen-record setting.
Table~\ref{tab:distinct_record_selected_text} summarizes descriptive
mean differences from the development-selected text strategies.

\begin{table}[H]
\centering
\footnotesize
\setlength{\tabcolsep}{4pt}
\caption{Descriptive PSS-minus-text differences against the strategy selected
on development for each bank size and retriever. Values are differences
in group-averaged All-Recall@4, in percentage points, over the same
16 evaluation groups.}
\label{tab:distinct_record_selected_text}
\begin{tabularx}{\linewidth}{@{}rlXr@{}}
\toprule
$K$ & Retriever & Development-selected text strategy & Difference (pp) \\
\midrule
4 & BGE-M3 & BM25 selector, top-1 & $+3.75$ \\
4 & BM25 & BM25 selector, top-1 & $-13.75$ \\
16 & BGE-M3 & Grok-based record selection & $+6.25$ \\
16 & BM25 & LLM bank rewriting & $+2.50$ \\
\bottomrule
\end{tabularx}
\end{table}

\paragraph{Online access workload.}
PSS uses one parameter-conditioned generation and one evidence query per
question, without supplying memory text to the LLM. Grok selection uses
one external model call followed by one evidence query; bank rewriting
also uses one generation and one evidence query. Concatenation uses one
evidence query, whereas record-wise RRF uses $K$. This comparison separates
model-based memory access from evidence-retrieval branching.

The archived local component profile reports mean bank-rewriting memory
inputs of 485.06 and 1367.69 tokens at $K=4$ and $K=16$, respectively.
At $K=16$, mean generated lengths are 63.60 tokens for PSS and 16.89
for bank rewriting. In \Cref{fig:main_retrieval}(b), NR denotes unreported
Grok memory-input token usage. Call counts, input length, output length,
and state loading are separate workload dimensions; the serving-cost
profile is reported in \Cref{app:followup_cost}.

\subsection{Evidence-controlled answering}
\label{app:followup_qa}

The answer-stage analysis uses a fixed 16-session subset of S. For the
controlled interface, every method answers the same explicit gold rewrite
with the same frozen, unloaded reader; only the retrieved evidence changes.
PSS evidence improves F1 over raw-query evidence by 6.03 points (95\% CI
$[2.64,9.23]$, $p_H=0.0214$), with seed-specific gains of 4.53, 7.87, and
5.69 points. This isolates a contribution of evidence access to answering
under a common question. The explicit question is an evaluation control
and is never provided to the memory writer or retriever.

A second interface uses each method's retrieval string as the reader's
question. Table~\ref{tab:followup_qa} reports both interfaces under one
eight-comparison Holm family. For this second interface, PSS-minus-raw is
$-6.89$ F1 points. Together, the two interfaces distinguish an evidence
benefit from preservation of the user's answer request: a retrieval clue
and a reader question perform different roles. The public 1.03\% to
20.00\% F1 comparison describes its full query-and-evidence interface;
the controlled result here establishes evidence attribution on its own
16-session evaluation scope.

\begin{table}[H]
\centering
\footnotesize
\setlength{\tabcolsep}{4pt}
\caption{Answer-interface analysis on 16 sessions from S. Effects are PSS minus reference; Holm correction covers all eight comparisons. Both interfaces use the same retrieved evidence for a given method and probe.}
\label{tab:followup_qa}
\begin{tabular}{llrrr}
\toprule
Reader input & Reference & $\Delta$ F1 (pp) & 95\% CI (pp) & $p_H$ \\
\midrule
Fixed question & Raw query & +6.03 & [+2.64, +9.23] & 0.021 \\
Fixed question & Query + supplied clue & -6.95 & [-14.27, +0.24] & 0.252 \\
Fixed question & Text-256 rewrite & +2.40 & [-3.16, +7.53] & 0.803 \\
Fixed question & Structured-256 rewrite & +1.18 & [-2.67, +4.68] & 0.803 \\
Retrieval query as question & Raw query & -6.89 & [-9.49, -4.48] & $<0.001$ \\
Retrieval query as question & Query + supplied clue & -7.92 & [-10.98, -4.90] & 0.002 \\
Retrieval query as question & Text-256 rewrite & -5.28 & [-8.84, -2.15] & 0.021 \\
Retrieval query as question & Structured-256 rewrite & -3.49 & [-6.79, -0.43] & 0.214 \\
\bottomrule
\end{tabular}
\end{table}

\begin{table}[H]
\centering
\footnotesize
\setlength{\tabcolsep}{4pt}
\caption{Complete online clue/retrieval/router costs for S: eight sessions, three writer seeds, three timed repeats (72 requests per method), BM25, and a 64-token readout budget. Prompt and Hist. count total and history-derived tokens in LLM prompts; a directly searched cached clue is not an LLM input. LM/Ret. count model and retrieval calls. Final answering and offline indexing are excluded.}
\label{tab:followup_cost}
\begin{tabular}{lrrrrrrr}
\toprule
Method & Mean (s) & p50 (s) & p95 (s) & Prompt & Hist. & LM & Ret. \\
\midrule
Raw query & 0.0035 & 0.0034 & 0.0040 & 0.00 & 0.00 & 0 & 1 \\
Query + supplied clue & 0.0051 & 0.0051 & 0.0055 & 0.00 & 0.00 & 0 & 1 \\
Compact card rewrite & 1.2843 & 1.2830 & 1.2913 & 81.38 & 21.62 & 1 & 1 \\
PSS clue & 1.3415 & 1.3401 & 1.3492 & 56.75 & 0.00 & 1 & 1 \\
Full-history rewrite & 2.6186 & 1.8230 & 5.3449 & 8495.00 & 8435.25 & 1 & 1 \\
Complete Guarded PSS & 2.6296 & 2.6291 & 2.6420 & 138.12 & 21.62 & 2 & 3 \\
\bottomrule
\end{tabular}
\end{table}

\begin{figure}[H]
\centering
\includegraphics[width=\linewidth]{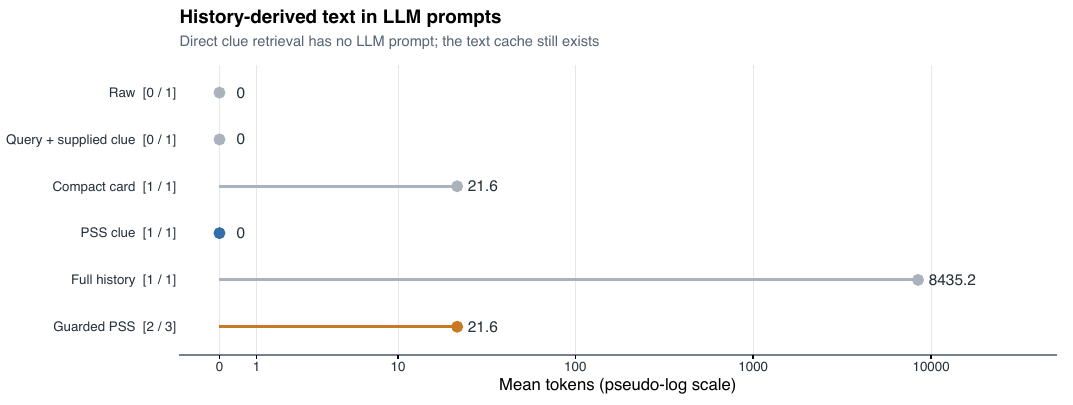}
\caption{History-derived tokens supplied to LLM prompts in the complete
cost profile. Brackets give LLM/retrieval call counts. The pseudo-log axis
includes a true zero. Direct clue retrieval uses persistent text as a
retrieval query without an LLM prompt; its zero plotted value does not
denote an absent text cache.}
\label{fig:followup_history_inputs}
\end{figure}

\subsection{Complete online accounting}
\label{app:followup_cost}

The cost profile measures eight sessions, three seeds, and three timed
repeats (72 requests per method) on the reported RTX 4090 with a reported
47.37 GiB device capacity. It uses BM25, warm indexes, CPU-resident states,
and 64-token readout generation. Timings include the complete
clue/retrieval/router path and state unloading; common final answering,
offline model/index construction, and disk state-cache misses are outside
this online boundary.

Pure PSS requires one clue generation and one retrieval, with no
history-derived text in its LLM prompt. It takes 1.341 seconds on average
versus 2.619 seconds for full-history rewriting, a 48.8\% reduction under
this workload. Complete Guarded PSS executes two generations and three
retrievals, including the card branch, and takes 2.630 seconds. It reads
21.625 card tokens in its LLM prompts; its total prompt count is 138.125.
These counts identify precisely where history replay is removed and where
a guarded text branch remains active.

The numerical-stability implementation uses CPU FP64 Adam master parameters
and moments, with GPU FP32 LoRA factors and BF16 serialized states. Its
mean write cost is 128.876 seconds over all 32 sessions and 113.058 seconds
over the matched cost-profile subset. Relative to the measured full-history
path, the latter amortizes after approximately 89 reads, assuming zero
one-time writer cost for the full-history reference. This 64-token profile
is separate from the earlier 16-token profile and its 43-read estimate.
The 44 MiB logical state, the mean 91.3-byte supplied clue, and the online
latencies represent separate resource dimensions. A fixed PSS-readout cache
also incurs the original state write and one readout before text-only use.

\end{document}